\documentclass[10pt,twocolumn,letterpaper]{article}

\usepackage{fontawesome5}
\usepackage[export]{adjustbox}
\usepackage{graphicx}
\usepackage[normalem]{ulem}
\usepackage{soul}
\usepackage{contour}

\usepackage{booktabs}
\usepackage{makecell}
\usepackage{multirow}

\usepackage{amssymb}
\usepackage{colortbl}

\usepackage[pagenumbers]{wacv} 

\definecolor{wacvblue}{rgb}{0.21,0.49,0.74}
\usepackage[pagebackref,breaklinks,colorlinks,allcolors=wacvblue]{hyperref}

\usepackage{tikz}
\usetikzlibrary{positioning, shadows.blur, shapes.geometric, arrows.meta, backgrounds, fit, calc}

\definecolor{Blue}{HTML}{1A5F7A}       
\definecolor{LightBlue}{HTML}{E7F6F2}  
\definecolor{Dark}{HTML}{2C3333}       
\definecolor{Gray}{HTML}{F0F0F0}       
\definecolor{Accent}{HTML}{FF8B13}   

\contourlength{0.8pt}
\newcommand{\myul}[1]{%
  \uline{\phantom{#1}}%
  \llap{\contour{white}{#1}}
}

\newcommand{\cmark}{\checkmark}

\newcolumntype{R}{>{\raggedleft\arraybackslash}X}

\newcommand\copyrighttext{%
\footnotesize \textcopyright 2026 IEEE. Personal use of this material is permitted.
Permission from IEEE must be obtained for all other uses, in any current or future
media, including reprinting/republishing this material for advertising or promotional
purposes, creating new collective works, for resale or redistribution to servers or
lists, or reuse of any copyrighted component of this work in other works.
}
\newcommand\copyrightnotice{%
\begin{tikzpicture}[remember picture,overlay]
\node[anchor=south,yshift=15pt] at (current page.south) {\fbox{\parbox{\dimexpr\textwidth-\fboxsep-\fboxrule\relax}{\copyrighttext}}};
\end{tikzpicture}%
}

\makeatletter
\def\thanks#1{\protected@xdef\@thanks{\@thanks
  \protect\footnotetext{#1}}}
\makeatother

\def\wacvPaperID{2}
\def\confName{WACV}
\def\confYear{2026}

\title{UEOF: A Benchmark Dataset for Underwater Event-Based Optical Flow}

\author{
Nick Truong$^{*}$, Pritam P. Karmokar$^{*}$, and William J. Beksi\\
The University of Texas at Arlington\\
Arlington, TX, USA\\
{\tt\small \{nxt0706,pritam.karmokar\}@mavs.uta.edu, william.beksi@uta.edu}%
\thanks{$^{*}$ Indicates equal contribution.}%
}

\begin{document}
\maketitle
\thispagestyle{empty}
\copyrightnotice

\begin{abstract}
Underwater imaging is fundamentally challenging due to wavelength-dependent
light attenuation, strong scattering from suspended particles,
turbidity-induced blur, and non-uniform illumination. These effects impair
standard cameras and make ground-truth motion nearly impossible to obtain. On
the other hand, event cameras offer microsecond resolution and high dynamic
range. Nonetheless, progress on investigating event cameras for underwater
environments has been limited due to the lack of datasets that pair realistic
underwater optics with accurate optical flow. To address this problem, we
introduce the first synthetic underwater benchmark dataset for event-based
optical flow derived from physically-based ray-traced RGBD sequences. Using a
modern video-to-event pipeline applied to rendered underwater videos, we
produce realistic event data streams with dense ground-truth flow, depth, and
camera motion. Moreover, we benchmark state-of-the-art learning-based and
model-based optical flow prediction methods to understand how underwater light
transport affects event formation and motion estimation accuracy. Our dataset
establishes a new baseline for future development and evaluation of underwater
event-based perception algorithms. The source code and dataset for this project
are publicly available at
\href{https://robotic-vision-lab.github.io/ueof}{https://robotic-vision-lab.github.io/ueof}.
\end{abstract}

\section{Introduction}
\label{sec:introduction}
Unmanned underwater vehicles (UUVs) require reliable perception systems for
tasks such as localization~\cite{fan2024underwater},
mapping~\cite{islam2024computer,wang2025eum,peng2025aquaticvision},
navigation~\cite{alvarez2023mimiruw,amer2023unavsim,novo2024neuromorphic,sheikder2025marine},
docking~\cite{cowen1997underwater,liu2018detection,yazdani2020survey},
examination~\cite{negahdaripour2007rov,betancourt2020integrated},
inspection~\cite{liljeback2017eelume,teigland2020operator}, object pose
estimation~\cite{mohammed20216d,nielsen2019evaluation}, and
manipulation~\cite{zhang2022dave,saksvik2023oslomet}. Nevertheless, underwater
vision-based perception remains extremely challenging due to the complex
light-medium interactions that cause rapid attenuation, wavelength-dependent
color shifts, low contrast, and strong backscatter from suspended
particles~\cite{steiner2013understanding,li2024breakthrough,barbosa2025physically}.
These effects degrade the performance of conventional cameras, especially
during fast motion or in low-light, turbid, or high-dynamic range (HDR)
conditions.

Compared to a frame-based camera, an event camera provides complementary
advantages in challenging underwater scenarios. The asynchronous, high temporal
resolution output of an event camera offers robustness to motion blur and HDR
lighting, making it well-suited for visual odometry, optical flow estimation,
and simultaneous localization and mapping (SLAM) in these environments. Yet,
advancements in underwater event-based perception have been significantly
constrained by a critical lack of data. Current underwater event-based datasets
rarely include accurate ground truth for motion and optical flow since light
detection and ranging and motion-capture systems do not function reliably
underwater. As a result, algorithm development and benchmarking have relied
predominantly on terrestrial datasets or synthetic data.

\begin{figure}
\includegraphics[width=\columnwidth,cfbox=black 1pt -1pt]{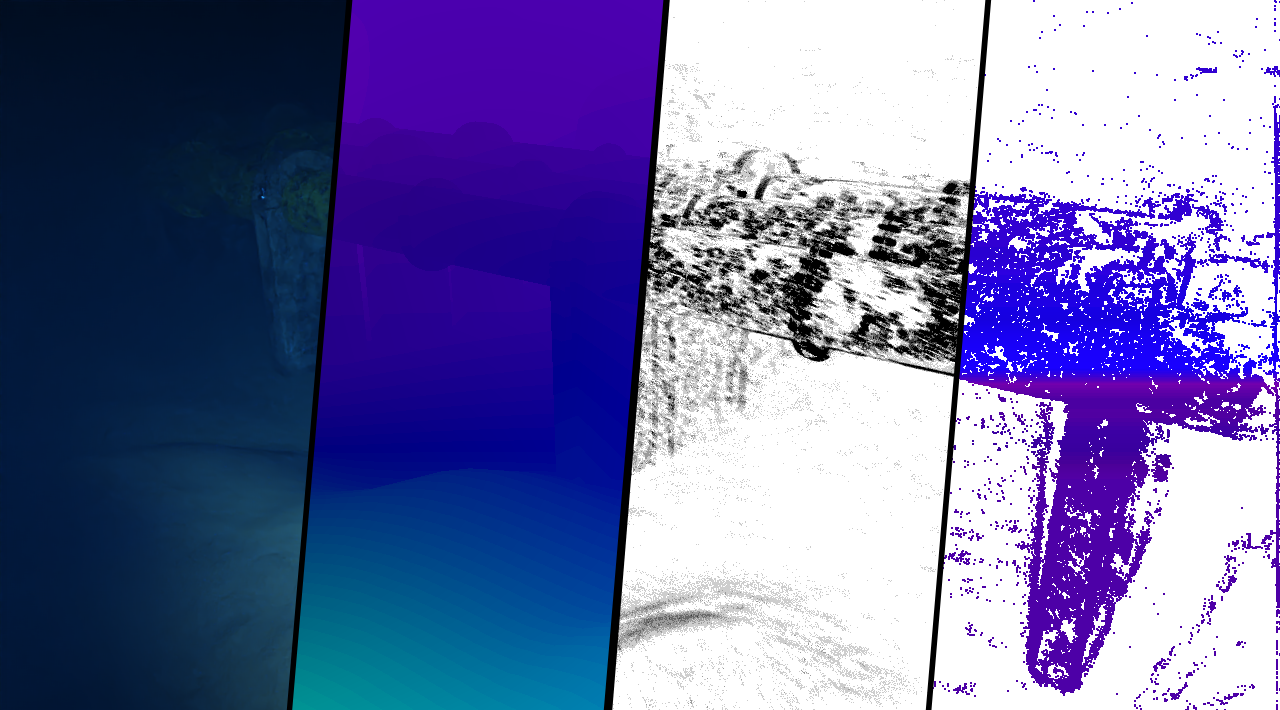}
\caption{An illustration of data and ground-truth modalities from the UEOF
dataset. Our dataset assembles physically-realistic underwater RGB imagery,
ground-truth optical flow, camera ego-velocities, and temporally dense event
streams, enabling the benchmarking of multimodal event-based optical flow
estimation algorithms.}
\label{fig:overview}
\end{figure}

To bridge this gap, we introduce a new underwater event-based optical flow
(UEOF) dataset generated by applying high-quality video-to-event conversion to
physically-based ray-traced (PBRT) RGBD sequences, Fig.~\ref{fig:overview}. In
particular, we obtain pseudo-event streams that capture the true temporal
dynamics of the scene, while preserving accurate ground-truth optical flow and
depth. This approach circumvents the limitations of both real-world data
collection and rasterization-based simulators. It provides the first benchmark
that combines physically accurate underwater rendering with temporally dense
event data for motion estimation. We summarize our contributions as follows.
\begin{enumerate}
  \item We present a synthetic underwater dataset that couples PBRT rendering
  with high temporal resolution pseudo-events through video-to-event
  conversion, enabling event-based evaluation under realistic underwater
  physics.

  \item We provide dense ground-truth optical flow, depth, and pose from PBRT
  simulation, making this the first underwater dataset suitable for
  benchmarking event-based optical flow algorithms.

  \item We benchmark a wide range of event-based optical flow techniques,
  including state-of-the-art event-driven and multimodal contrast maximization
  (CM) approaches, revealing the failure modes and domain-gap challenges unique
  to underwater scenes.

  \item We analyze how underwater optical phenomena (\eg, attenuation,
  backscatter, caustics, turbidity) affect event generation and event
  representations, offering insight into the suitability of event cameras for
  underwater perception.
\end{enumerate}
Our dataset establishes a foundation to study event-based motion estimation in
underwater environments and paves the way to realize robust multimodal
perception systems for UUV autonomy.

The remainder of this paper is organized as follows. In
\cref{sec:related_work}, we review the landscape of existing underwater
datasets and event simulators, identifying the critical gaps in ground truth
availability that motivates our work. \cref{sec:method} details the proposed
data generation pipeline, describing the integration of high-fidelity ray
tracing with event simulation to produce the UEOF dataset. We present a
comprehensive benchmarking of state-of-the-art supervised, unsupervised, and
model-based algorithms along with a critical analysis of how underwater optical
phenomena impacts estimation performance in \cref{sec:evaluation}. Finally, we
conclude in \cref{sec:conclusion_and_future_work} and discuss future directions
for underwater event-based perception.

\section{Related Work}
\label{sec:related_work}
\subsection{Real-World Underwater Event Datasets}
\label{subsec:real-world_underwater_event_datasets}
Several datasets contain real-world underwater event data, yet none capture
ground-truth optical flow due to the difficulty of obtaining accurate motion
supervision. For example, DAVIS-NUIUIED~\cite{bi2022non} furnishes frame and
event data recorded under non-uniform illumination to support underwater image
enhancement research, without motion or pose estimations.
FLSea~\cite{randall2023flsea} provides stereo RGB images and inertial
measurement unit (IMU) data for underwater SLAM evaluation, sans optical flow
or event annotations. The Aqua-Eye~\cite{luo2023transcodnet} dataset contains
annotated DAVIS346 images, events, and fused event-frame representations for
detecting transparent marine organisms. UTNet~\cite{guo2025utnet} extends
Aqua-Eye by integrating ResNet50~\cite{he2016deep} with submanifold sparse
convolutions for improved segmentation of transparent underwater objects.
Although both datasets include events, they do not provide depth or optical
flow ground truth.

The AquaticVision~\cite{peng2025aquaticvision} dataset provides synchronized
stereo DAVIS346 events, grayscale frames, and IMU data at high frequency, along
with 6-DoF motion-capture ground truth. The sequences cover clear water, HDR
conditions, and varying turbidity. An analysis shows that event representations
such as time surfaces perform well in clear water, but degrade significantly in
turbid scenes, while event packet and voxel-grid representations offer improved
robustness. The dataset is designed for visual SLAM and benchmarking, yet it
lacks optical flow annotations. EvtSlowTV~\cite{macaulay2025evtslowtv}
generates synthetic events from YouTube videos using video-to-event conversion,
including underwater exploration footage, however it is not designed for
optical flow or SLAM evaluations.

\begin{figure*}
  \centering
  \begin{adjustbox}{max width=\textwidth}
  \begin{tabular}{@{}c@{\thinspace}c@{\thinspace}c@{\thinspace}c@{}}
    \includegraphics[width=0.25\textwidth,cfbox=gray 0.1pt 0pt]{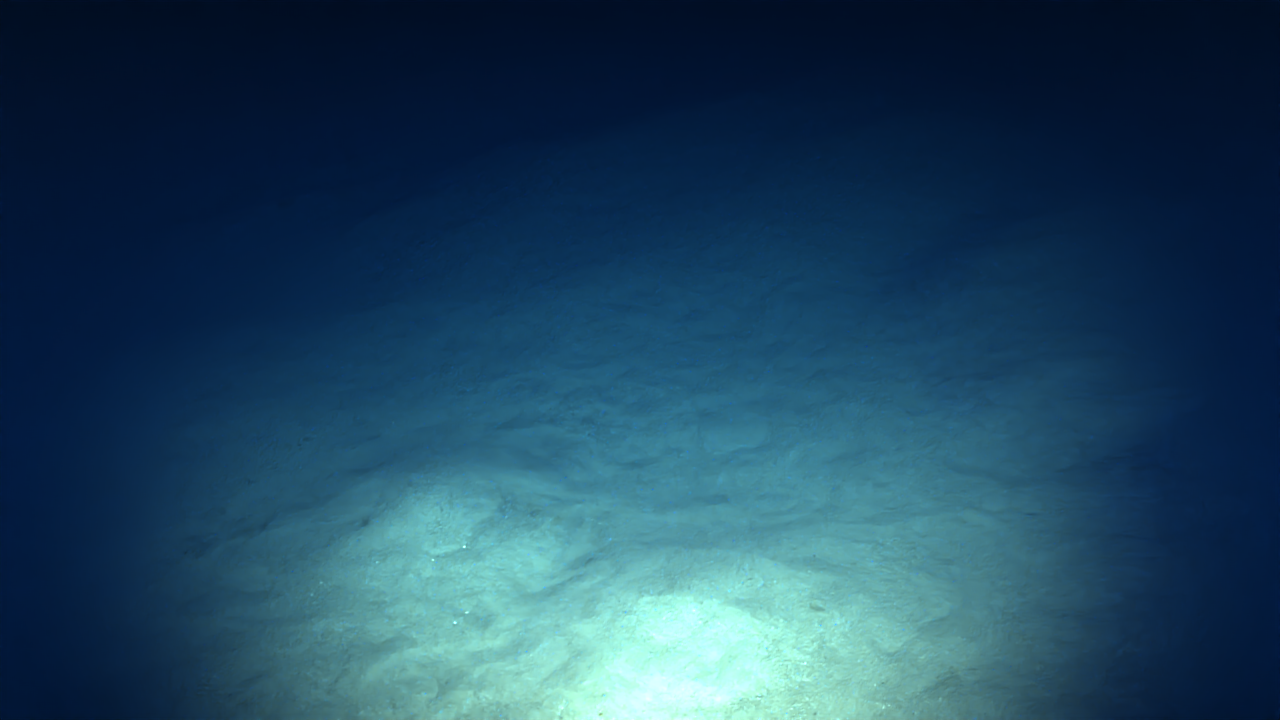}%
    & \includegraphics[width=0.25\textwidth,cfbox=gray 0.1pt 0pt]{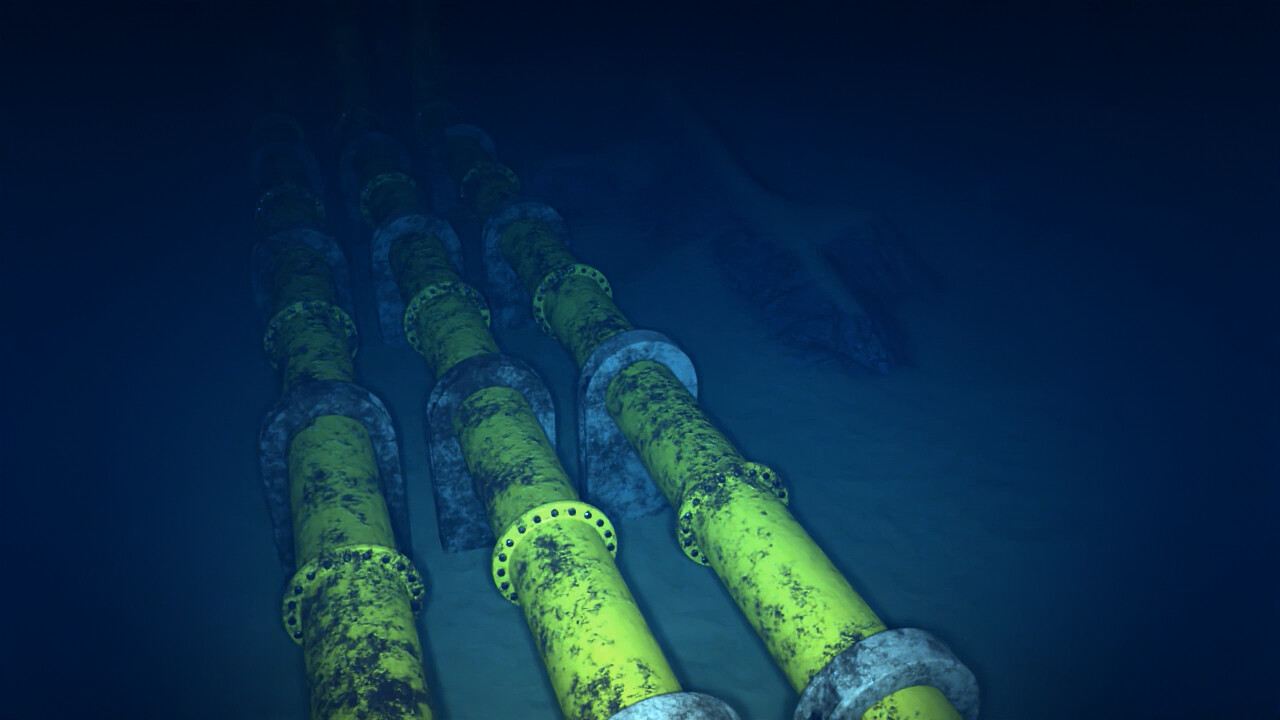}%
    & \includegraphics[width=0.25\textwidth,cfbox=gray 0.1pt 0pt]{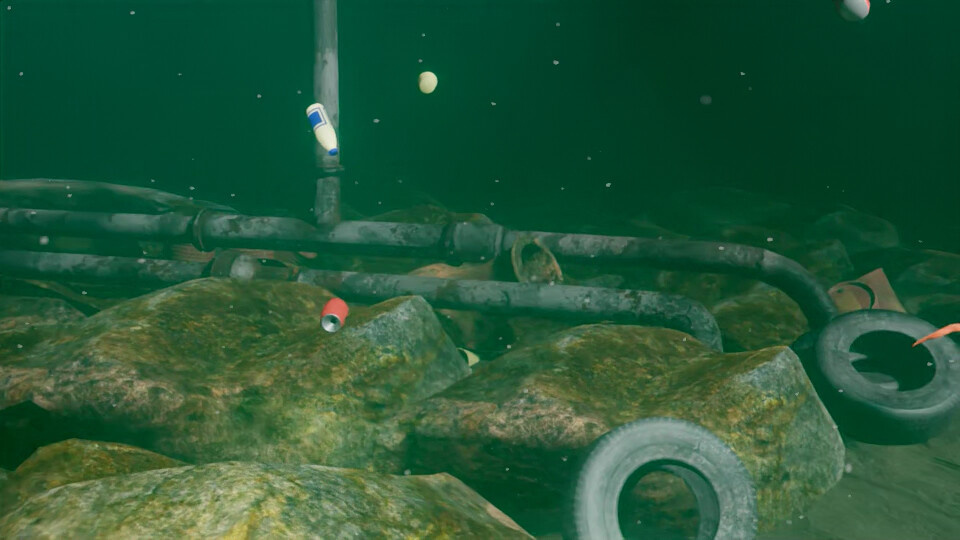}%
    & \includegraphics[width=0.25\textwidth,cfbox=gray 0.1pt 0pt]{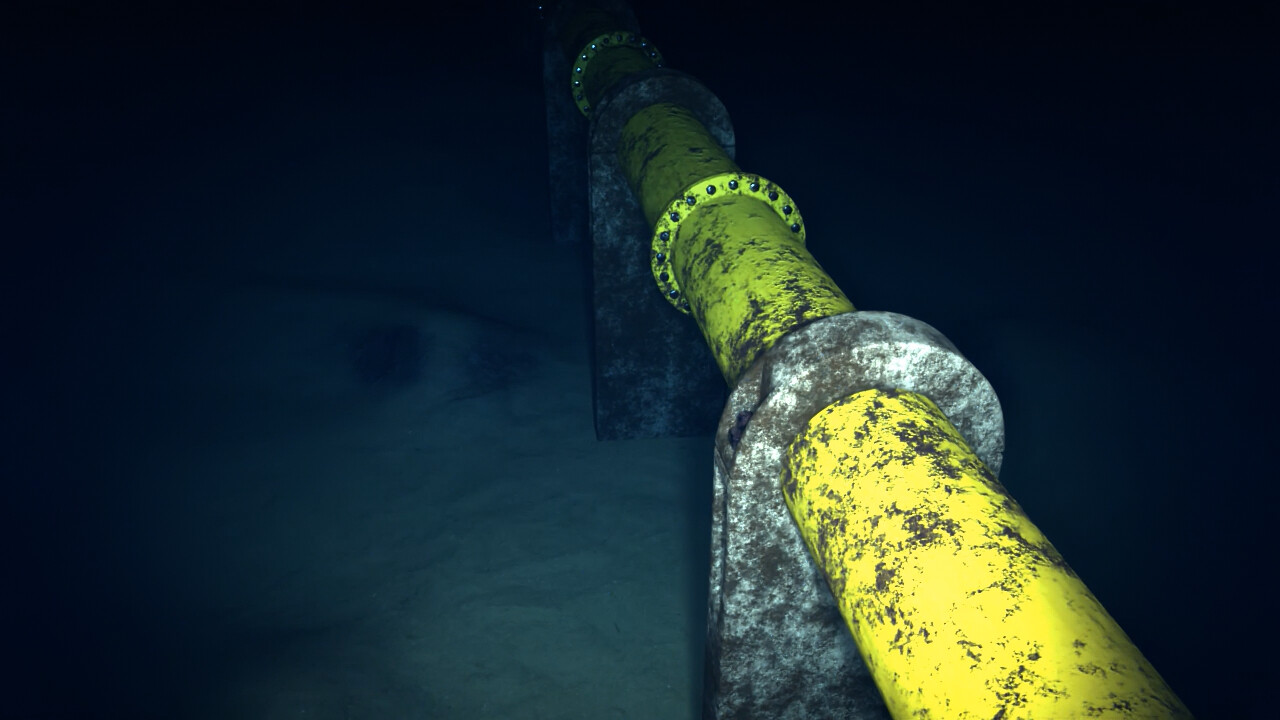}\\[-2pt]
    \includegraphics[width=0.25\textwidth,cfbox=gray 0.1pt 0pt]{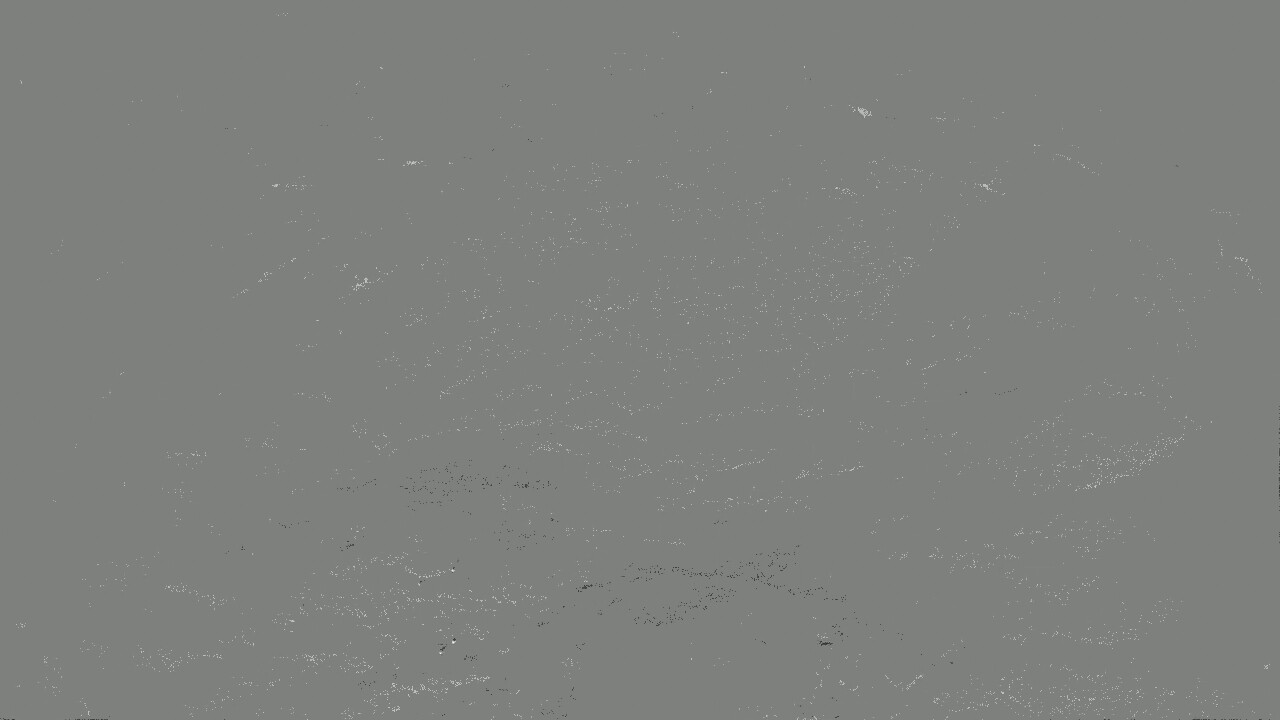}%
    & \includegraphics[width=0.25\textwidth,cfbox=gray 0.1pt 0pt]{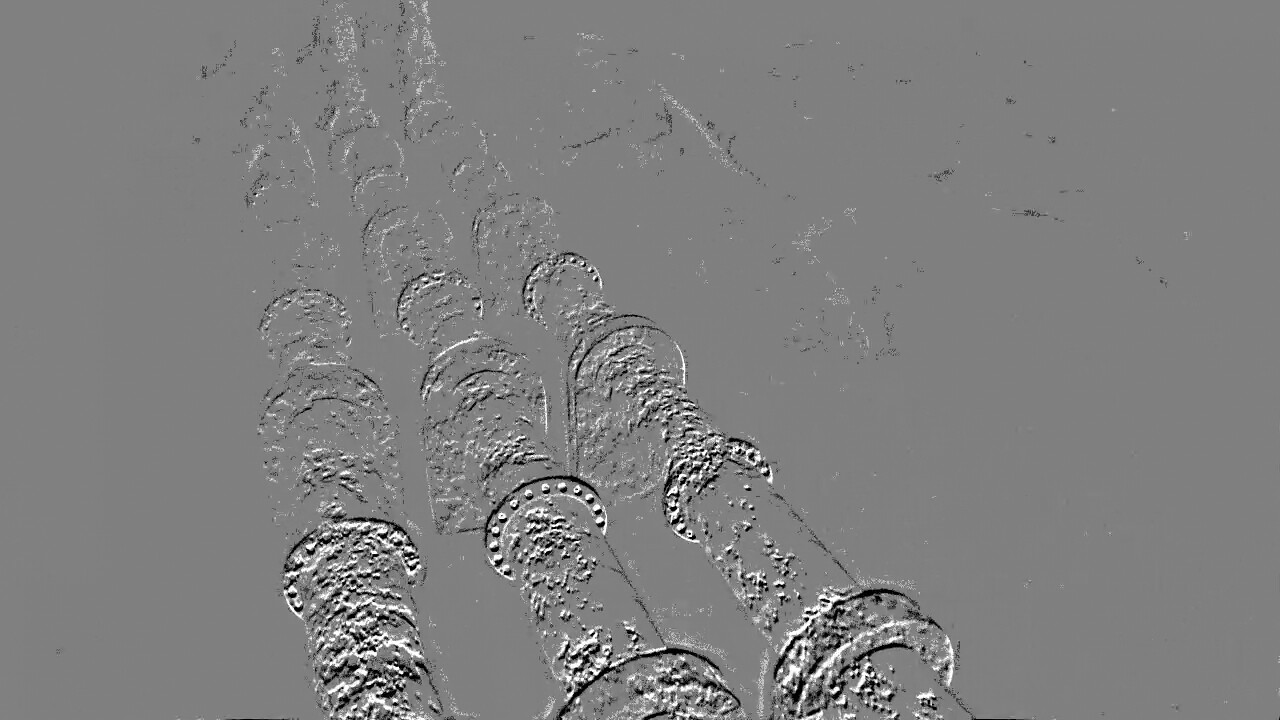}%
    & \includegraphics[width=0.25\textwidth,cfbox=gray 0.1pt 0pt]{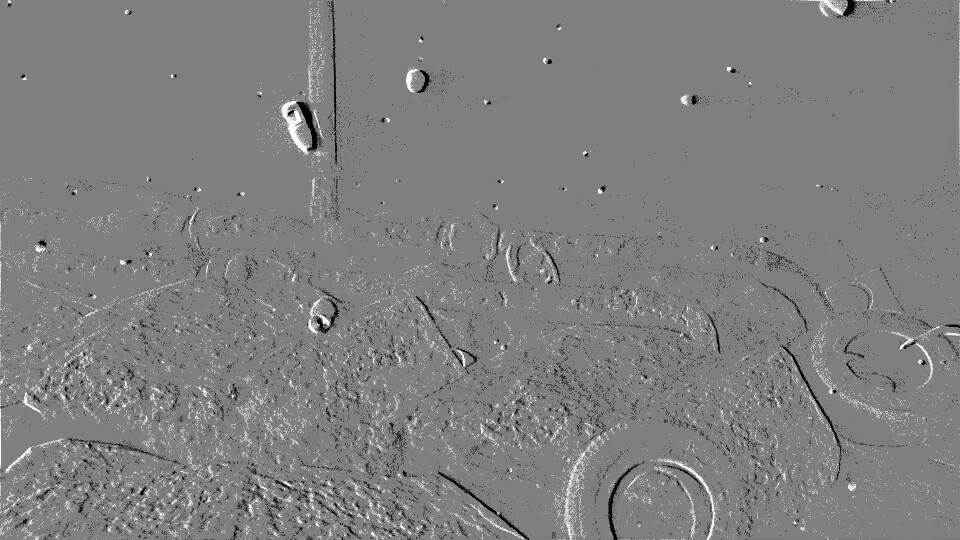}%
    & \includegraphics[width=0.25\textwidth,cfbox=gray 0.1pt 0pt]{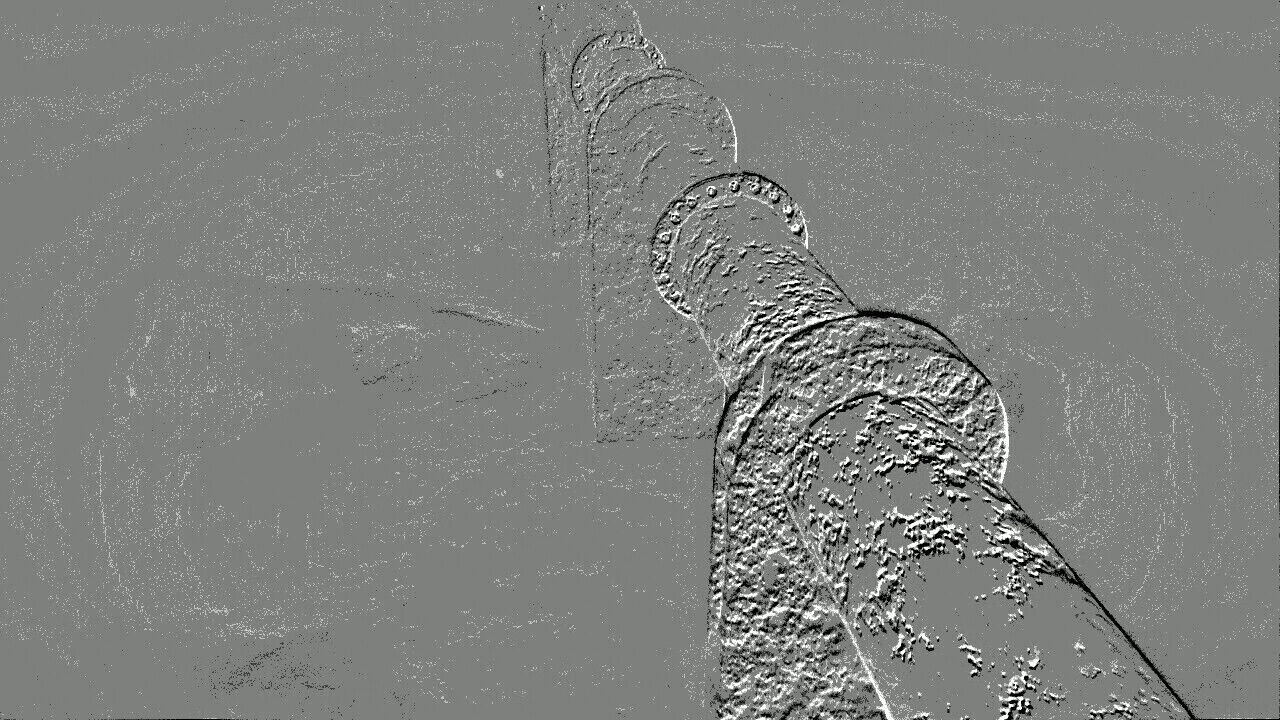}\\
    \small(a) Low contrast
    & \small(b) Turbidity
    & \small(c) Caustics, marine snow
    & \small(d) Haloing\\
  \end{tabular}
  \end{adjustbox}
  \caption{An illustration of key underwater visual degradation modes and their
  impact on event formation. RGB frames (top row) and the corresponding events
  (bottom row) are shown for a variety of challenging underwater scenarios.}
  \label{fig:underwater_imaging_challenges}
\end{figure*}

\subsection{Synthetic Underwater Scenes}
\label{subsec:synthetic_underwater_scenes}
Rasterization-based simulators and PBRT engines differ drastically in their
ability to model underwater light transport. For instance,
Gazebo~\cite{koenig2004gazebo}, UUV Simulator~\cite{manhaes2016uuv}, and
Stonefish~\cite{cieslak2019stonefish,grimaldi2025stonefish} rely on
shader-based approximations. They cannot simulate volumetric scattering,
spectral attenuation, or photon transport, leading to unrealistic underwater
imagery that is unsuitable for photometrically sensitive
tasks~\cite{liao2023gpu,yue2025monte}. MIMIR-UW~\cite{alvarez2023mimiruw} is
built atop Unreal Engine~\cite{unrealengine57} and
AirSim~\cite{shah2017airsim}.  It provides synthetic images, segmentation
masks, and 6-DoF ground truth for SLAM and object inspection, but it does not
include optical flow. Similarly, OysterSim~\cite{lin2022oystersim} simulates
oyster reef monitoring with underwater scattering and includes IMU, sonar, and
camera pose data, yet it does not have event-based optical flow data.

Ray-traced renderers such as Blender Cycles~\cite{blender50} and Unreal Engine
simulate photon transport, spectral absorption, and volumetric scattering.
VAROS~\cite{zwilgmeyer2021varos} yields ray-traced underwater RGB imagery with
depth and surface normals. LOFUE~\cite{ferone2023synthetic} extends this by
adding \emph{ground-truth optical flow}, making it one of the most physically
realistic underwater datasets with optical-flow labels.
PHISWID~\cite{kaneko2024phiswid} simulates underwater color degradation using
absorption, scattering, and marine snow applied to terrestrial RGBD sequences.
Although physically-based renderers produce highly realistic underwater
imagery, they do not natively generate event data. This motivates the use of
video-to-event conversion software such as the
v2e~\cite{gehrig2019video,hu2021v2e} toolbox.

\subsection{Event Simulation}
\label{subsec:event_simulation}
Existing simulators such as ESIM~\cite{rebecq2018esim},
MDR~\cite{luo2023learning}, and BlinkSim~\cite{li2023blinkflow} rely on
rasterized images and cannot model underwater light physics.
eCARLA-scenes~\cite{mansour2024ecarla} adapts CARLA~\cite{dosovitskiy2017carla}
for event data with a focus on autonomous driving.
eStonefish-scenes~\cite{mansour2025estonefish} employs the Stonefish simulator
to generate synthetic optical flow and events, but it inherits the limitations
of rasterization (\ie, no volumetric scattering and simplified light
attenuation). Blender Cycles is utilized by EREBUS~\cite{kyatham2025erebus} for
high-quality rendering. However, its event data is generated via the v2e
toolbox and does not target optical flow. No existing event simulator models
underwater light transport, no underwater event dataset provides ground-truth
optical flow, and no ray-traced underwater dataset includes events, which
beckons the need for new data generation pipelines. \textit{To the best of our
knowledge, UEOF is the first dataset to leverage PBRT RGBD data to generate
temporally dense pseudo-events for evaluating event-based optical flow
algorithms.}

\section{Underwater Event-Based Optical Flow Dataset}
\label{sec:method}
\subsection{Underwater Simulation}
\begin{figure*}
\centering
\includegraphics[width=0.7\textwidth, cfbox=gray 0pt 0pt]{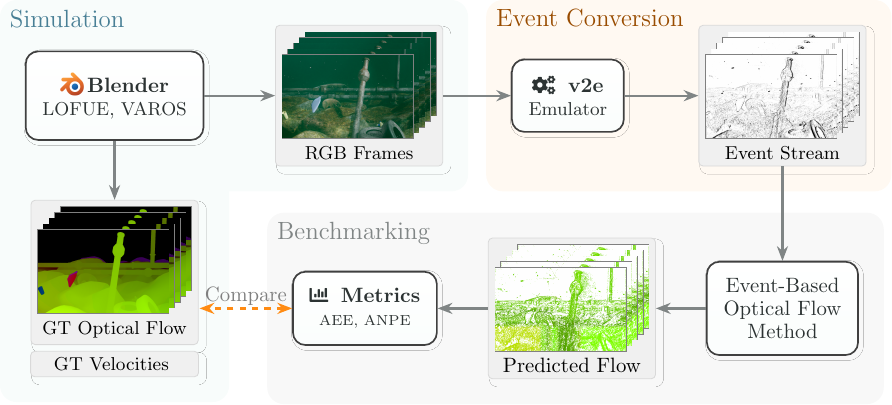}
\caption{The UEOF data generation pipeline. First, Blender is utilized to render
high-fidelity RGB frames and extract ground-truth data from the LOFUE and VAROS
datasets. Next, the RGB frames undergo event conversion via the v2e toolbox.
Finally, the resulting event streams are processed by event-based optical flow
methods and the predicted flow is evaluated against the ground-truth data using
standard endpoint error metrics.}
\label{fig:data_generation_pipeline}
\end{figure*}

Realistic underwater rendering requires modeling complex interactions between
light and a heterogeneous optical medium. Water contains dissolved organic
matter, temperature gradients, salinity variations, air bubbles, suspended
particles, and marine snow, all of which alter absorption and scattering in
space and time. Consequently, underwater image formation is nonlinear,
wavelength-dependent, and strongly range-dependent, making it difficult to
faithfully simulate. Underwater light transport involves wavelength-dependent
attenuation, and both forward and back scattering governed by complex phase
functions~\cite{barbosa2025physically}. Accurate simulation requires solving
the radiative transfer equation, typically via Monte Carlo photon simulation,
which is computationally expensive and sensitive to optical
parameters~\cite{sedlazeck2011simulating}.

As highlighted in \cref{fig:underwater_imaging_challenges}, optical properties
vary with depth, water type (coastal vs.~oceanic), turbidity, and environmental
conditions~\cite{schontag2025optical}. Bubbles, turbulence, flickering
caustics, and marine snow introduce temporal inconsistencies that are difficult
to capture using static rendering pipelines~\cite{yang2025seasplat}. Natural
light attenuates rapidly with depth, producing low-light blue-green-shifted
imagery~\cite{makam2024oceanlens}. Artificial illumination introduces strong
backscatter, halos, and veiling glare whose appearance depends on the geometry
of the light sources and the particle field~\cite{kaneko2024phiswid}.
Wavelength-dependent attenuation causes color shifts and contrast loss that
cannot be captured by simple haze-style degradation
models~\cite{li2024breakthrough}. For many underwater scenes, the true
reflectance or ``water-free'' reference appearance is unknown, complicating the
validation of synthetic renderings~\cite{zwilgmeyer2021varos}.

Real-world underwater event datasets with dense ground-truth optical flow are
nonexistent. Moreover, they rarely include accurate optical measurements,
geometry, or lighting conditions. This makes it difficult to parameterize
simulators and leads to significant domain gaps between synthetic and real
imagery~\cite{li2025realistic}. Furthermore, existing synthetic event
simulators are built on rasterization pipelines that do not model volumetric
light transport or scattering, limiting their applicability in the underwater
domain.  These fundamental challenges limit the fidelity of underwater
rendering engines, an issue that becomes even more restrictive when simulating
event data or generating training labels such as optical flow. We address these
issues via a high-fidelity pseudo-event dataset. Concretely, the UEOF dataset
provides the following: (i) photometrically realistic RGB imagery rendered
using Blender Cycles, (ii) asynchronous event streams emulated with the v2e
toolbox, (iii) dense per-pixel optical flow ground truth, and (iv) camera
velocities.

\subsection{Simulation Sources}
\label{subsec:simulation_sources}
UEOF was built from two underwater ray-traced datasets that complement one
another in scene complexity, motion patterns, and rendering realism. The first
dataset, VAROS, simulates UUV operations across infrastructure inspection and
seafloor exploration scenarios. We divide the VAROS data into five distinct
scenes where each one contains the following: (i) lossless RGB images, (ii)
surface normal vector images, (iii) metric depth maps, (iv) accurate 6-DoF
ground-truth poses, (v) synchronized IMU data, and (vi) depth-gauge data. The
scenes comprise pipes, man-made structures, and the natural seabed with
physically correct illumination and scattering. Although optical flow is not
provided, all dynamic motion arises purely from camera ego-motion (\ie, no
independently moving objects), enabling perfect geometric reconstruction.

The second dataset, LOFUE, features complex underwater scenes with dynamic
objects (\eg, fish, vegetation, floating debris, \etc), varied water turbidity,
and both static and moving cameras. Specifically, \texttt{scene1} and
\texttt{scene5} feature a static camera, while \texttt{scene2},
\texttt{scene3}, and \texttt{scene4} present a moving camera simulating
realistic vehicle motion.  All imagery is rendered using Blender Cycles with
physically-based light propagation, including volumetric scattering and
spectral attenuation. LOFUE provides RGB frames and dense optical flow ground
truth using the add-on VisionBlender~\cite{cartucho2020visionblender}. Together
VAROS and LOFUE offer complementary strengths. LOFUE includes dynamic-object
flow, while VAROS yields precise geometric and inertial measurements. Both
datasets supply physically-grounded underwater imagery.

\subsection{Event Data Generation}
\label{subsec:event_data_generation}
We generated asynchronous event streams from the rendered RGB sequences using
the v2e toolbox, which models the behavior of a dynamic vision sensor (DVS).
Standard v2e parameters are designed for terrestrial scenes and require
adaptation for underwater imagery, where contrast is heavily degraded due to
scattering and spectral attenuation. For each source sequence, the raw frames
were first converted to video at 30\,FPS and 10\,FPS from the LOFUE and VAROS
datasets, respectively. Then, the v2e software was used to perform slow-motion
interpolation such that no pixel translates more than 1\,px between adjacent
frames, ensuring realistic event triggering under fast camera motion or
low-light visibility.

To increase DVS sensitivity in low-texture or heavily attenuated regions (\eg,
sandy seafloors), we lowered the positive/negative thresholds from the default
values. This improves the event density in regions that would otherwise fail to
produce measurable illumination changes. Nonetheless, lower thresholds,
$\theta_{\text{nominal}}$, risk false triggering in darker areas. We countered
this by reducing the threshold variation, $\sigma_{\theta}$, effectively
suppressing fixed-pattern noise without blurring genuine edges. Events are
exported via the HDF5 file structure as a sequence of $(t, x, y, p)$ tuples,
where $t$ is the timestamp, $(x, y)$ is the pixel location, and $p$ is the
polarity of the fired event. An overview of the UEOF data generation pipeline
is shown in \cref{fig:data_generation_pipeline}.

\subsection{Ground-Truth Data Generation}
\label{subsec:ground-truth_data_generation}
\begin{table*}
  \centering
  \begin{adjustbox}{max width=0.75\linewidth}
  \begin{tabular}{l l c c c c c c}
    \toprule
    & & & & \multicolumn{2}{c}{\textbf{Modalities}} & \multicolumn{2}{c}{\textbf{Ground Truth}} \\
        
    \cmidrule(lr){5-6} \cmidrule(lr){7-8}
    
   \textbf{Dataset} & 
   \textbf{Sensor / Sim.} & 
   \textbf{Res ($W \times H$)} & 
   \textbf{Duration} & 
   \textbf{RGB} & \textbf{Events} & \textbf{Flow} & \textbf{Vel} \\
   \midrule

   \multicolumn{8}{l}{\textit{Real-World Datasets}} \\
   \midrule
        
   AquaticVision \cite{peng2025aquaticvision} & 
   DAVIS346 & $346 \times 260$ &
   13m06s & - & \cmark & - & \cmark \\
        
   OsloMet \cite{saksvik2023oslomet} & 
   DAVIS346 & $346 \times 260$ &
   -- & - & \cmark & - & \cmark \\

   Aqua-eye \cite{luo2023transcodnet} & 
   DAVIS346 & $346 \times 260$ &
   -- & \cmark & \cmark & - & - \\
        
   DAVIS-NUIUIED \cite{bi2022non} & 
   DAVIS346 & $346 \times 260$ &
   -- & \cmark & \cmark & - & - \\

   \midrule
   \multicolumn{8}{l}{\textit{Synthetic and Converted Datasets}} \\
   \midrule
        
   EvtSlowTV \cite{macaulay2025evtslowtv} & 
   YouTube$\to$ESIM & Var. & 
   9000m & \cmark & \cmark & - & - \\
        
   eStonefish \cite{mansour2025estonefish} & 
   Stonefish & $1280 \times 720$ & 
   6m & - & \cmark & \cmark & - \\
        
   EREBUS \cite{kyatham2025erebus} & 
   Blender$\to$V2E & $1920 \times 1080$ & 
    -- & \cmark & \cmark & - & - \\
    
   \midrule
   \rowcolor{gray!10} 
   \textbf{UEOF (Ours)} & 
   Blender$\to$V2E & \textbf{1280 $\times$ 720}$^{\ast}$ & 
   \textbf{12m51s} & \cmark & \cmark & \cmark & \cmark \\
        
   \bottomrule
   \end{tabular}
   \end{adjustbox}
   \caption{A comparison of underwater event-based datasets. While several
   datasets exist, UEOF is the first to provide ground-truth optical flow and
   camera ego-velocity derived from PBRT simulation. \emph{Res}: resolution.
   \emph{Flow}: optical flow. \emph{Vel}: camera ego-velocity. `--': duration
   not reported or available. $^{\ast}$Includes subsets at $960 \times 540$
   resolution.}
   \label{tab:dataset_comparison}
\end{table*}

The UEOF dataset contains dense per-pixel optical flow maps and 6-DoF camera
velocity ground truth for all sequences. The ground-truth optical flow is
stored in the standard Middlebury format (\texttt{.flo}), aligned with each RGB
frame.  For the LOFUE sequences, we utilized the precomputed optical flow
ground truth provided by the authors. The VAROS sequences do not provide
optical flow data.  Therefore, we reconstructed dense optical flow via
geometric back-projection.  Specifically, valid pixels $\mathbf{u}_t$ were
back-projected into 3D points $\mathbf{P}_t = \pi^{-1}(\mathbf{u}_t,
D_t(\mathbf{u}_t))$ using the provided metric depth map $ D_t $ and intrinsic
parameters. These points were then transformed to the subsequent frame via the
rigid-body motion $\mathbf{T}_{t \rightarrow t+1} \in SE(3) $ yielding the
induced flow $ \Delta \mathbf{u} = \pi(\mathbf{T}_{t \rightarrow t+1}
\mathbf{P}_t) - \mathbf{u}_t$. We also applied a geometric occlusion check by
comparing transformed points projected onto the target image plane, and their
computed depth $z_{t+1}$ against the target depth map $D_{t+1}$.
Correspondences were considered valid only if they fell within image bounds,
satisfied the near-plane constraint ($z_{t+1} > 0.1\,\text{m}$), and passed the
visibility test ($z_{t+1} \leq D_{t+1} + \epsilon$). We set the tolerance
$\epsilon = 0.01\,\text{m}$. This filters out occluded regions where the
projected point lies behind the visible surface.

The 6-DoF linear ($\mathbf{v}$) and angular ($\boldsymbol{\omega}$)
ground-truth camera ego-velocities are provided for all the sequences in the
UEOF dataset.  These measurements are stored and compressed as \texttt{.npz}
archives containing three primary keys: \texttt{lin\_vel}, \texttt{ang\_vel},
and \texttt{timestamps} in microseconds. For the LOFUE sequences, we utilized
the Blender Python API to directly extract the camera's instantaneous
rigid-body velocity vectors in the camera coordinate frame at each render step
(30\,Hz).  For the VAROS sequences, we derived the velocities from the provided
high-frequency (200\,Hz) ground-truth camera poses. Then, we computed the
body-frame velocities using central finite differences on the position and
quaternion parameters by converting quaternion derivatives to angular velocity
via the standard kinematic relationship $\boldsymbol{\omega} = 2
\hat{\mathbf{q}}^{-1} \otimes \dot{\hat{\mathbf{q}}}$, where $\otimes$ denotes
the Hamilton product, $\hat{\mathbf{q}}$ is the unit quaternion representing
camera orientation, and $\dot{\hat{\mathbf{q}}}$ is its time derivative.

\subsection{Dataset Statistics}
\label{subsec:dataset_statistics}
\begin{figure}
\centering
\begin{subfigure}[b]{\linewidth}
  \includegraphics[width=\linewidth]{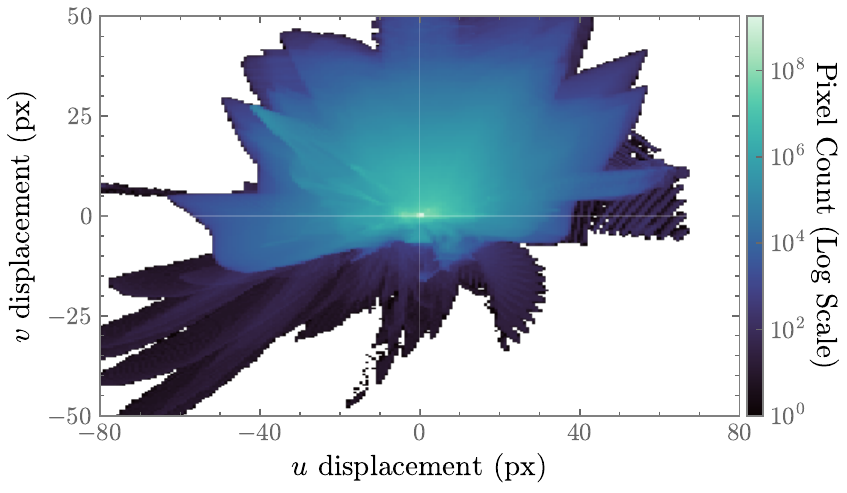}
  \caption{Pixel displacement}
  \label{subfig:optical_flow_displacement}
\end{subfigure}
\begin{subfigure}[a]{\linewidth}
  \includegraphics[width=\linewidth]{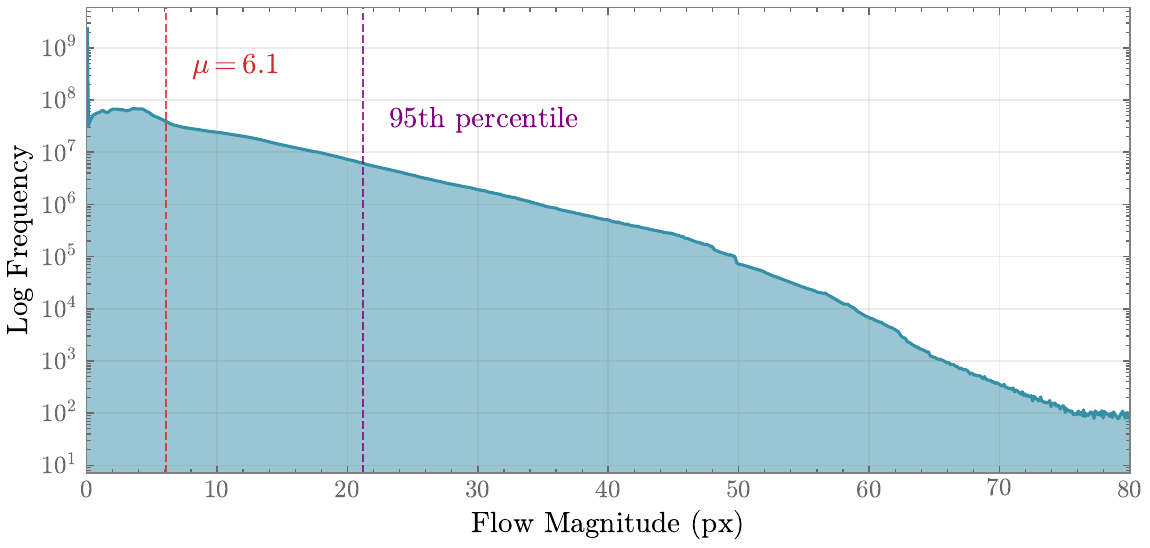}
  \caption{Magnitude distribution}
  \label{subfig:optical_flow_magnitude}
\end{subfigure}
\caption{Optical flow statistics for the UEOF dataset: (a) the joint
distribution of $(u,v)$ pixel displacements highlight dense coverage of motion
directions; (b) the semi-log histogram shows a heavy-tailed distribution with a
mean of 6.1\,px.} 
\label{fig:optical_flow_statistics}
\end{figure}

Using the proposed simulation pipeline, we generated 12 minutes and 51 seconds
of data across 13,714 RGB frames. This results in a total of 4.94 billion
events across all scenes. In summary, the UEOF dataset exhibits the following
key characteristics.
\begin{itemize}
  \item \myul{High Resolution}: As indicated in \cref{tab:dataset_comparison},
  the dataset provides event streams in two spatial resolutions: $960 \times
  540$ and $1280 \times 720$. Notably, the $1280 \times 720$ resolution
  coincides with the sensor specifications of the Prophesee EVK4 HD with the
  Sony IMX636ES HD sensor, a widely used event camera. This alignment minimizes
  the simulation-to-reality gap and establishes a realistic benchmark for
  evaluating event-based optical flow estimation methods at standard sensor
  resolutions. 
  
  \item \myul{Temporally Dense}: The LOFUE scenes operate at 30\,FPS for RGB
  frames, with synchronized ground-truth velocities and optical flow evaluation
  provided at 30\,Hz. The VAROS scenes run at 10\,FPS for RGB frames
  (evaluations at 10\,Hz). However, the camera ego-velocities are derived from
  high-precision logs recorded at 200\,Hz. In total, the dataset provides
  13,704 evaluation intervals.

  \item \myul{Comprehensive Motion}: As displayed in
  \cref{fig:optical_flow_statistics}, the dataset exhibits a high dynamic range
  of motion, with a mean flow magnitude of $6.1$\,px and a median of $3.6$\,px.
  The motion distribution is heavy-tailed. While a majority of pixels undergo
  moderate displacement (with a 95th percentile of $21.20$\,px), the dataset
  also includes significant high-magnitude flow upwards of $80$\,px.
\end{itemize}

\begin{table}[ht]
    \centering
    
    \begin{adjustbox}{max width=\columnwidth}
    \begin{tabular}{@{}c@{\thickspace}l c c c c c@{}}
        \toprule
        \multirow{2}{*}{} 
        & \multirow{2}{*}{Methods} 
        & \texttt{scene1} & \texttt{scene2} 
        & \texttt{scene3} & \texttt{scene4} 
        & \texttt{scene5}\\

        \cmidrule(lr){3-3} \cmidrule(lr){4-4} \cmidrule(lr){5-5} \cmidrule(lr){6-6} \cmidrule(lr){7-7}
        & & AEE ($\downarrow$) & AEE ($\downarrow$) & AEE ($\downarrow$) & AEE ($\downarrow$) & AEE ($\downarrow$) \\
        
        \midrule
        \multirow{2}{*}{\rotatebox[origin=c]{90}{LB}}
        & E-RAFT~\cite{gehrig2021eraft}
        & 8.67  & 9.05  & 10.52 & 8.64  & 16.32 \\
        & MotionPriorCMax~\cite{hamann2024motion}
        & \textbf{2.61} & \underline{1.54} & \textbf{1.92} & 1.49  & \underline{2.36} \\
        \midrule
        \multirow{3}{*}{\rotatebox[origin=c]{90}{MB}}
        & EINCM~\cite{karmokar2025secrets}
        & \underline{4.52} & 2.36 & \underline{2.06} & \textbf{1.01} & 4.12\\
        & MultiCM~\cite{shiba2022secrets}
        & 5.10 & \textbf{1.37} & 2.42 & \underline{1.37} & \textbf{2.27} \\
        & OPCM~\cite{karmokar2025inertia}
        & --     & 1.62     & 2.14     & 1.93     & -- \\
        \bottomrule
    \end{tabular}
    \end{adjustbox}
    \caption{The average endpoint error (AEE) in pixels on the UEOF dataset
    (shallow-water environment) across five scenes ($dt=1$). \textbf{Bold}
    indicates best, \underline{underline} indicates second-best.}
    \label{tab:lofue_aee}
\end{table}

\begin{table}
  \centering
  \begin{adjustbox}{max width=\linewidth}
  \begin{tabular}{@{}c@{\thickspace}l c c c c c@{}}
    \toprule
    \multirow{2}{*}{\rotatebox{90}{}}
    & \multirow{2}{*}{Methods}
    & \texttt{scene1} & \texttt{scene2} & \texttt{scene3} & \texttt{scene4} & \texttt{scene5} \\
    \cmidrule(lr){3-3} \cmidrule(lr){4-4} \cmidrule(lr){5-5} \cmidrule(lr){6-6} \cmidrule(lr){7-7}
    & & AEE ($\downarrow$) & AEE ($\downarrow$) & AEE ($\downarrow$) & AEE ($\downarrow$) & AEE ($\downarrow$) \\

    \midrule
        
    \multirow{2}{*}{\rotatebox[origin=c]{90}{LB}}
    & E-RAFT~\cite{gehrig2021eraft}
    & 5.82 & 10.11 & 4.88 & 7.64 & 3.61 \\
        
    & MotionPriorCMax~\cite{hamann2024motion}
    & 6.93 & 5.61 & 6.24 & 7.39 & 5.25 \\
    \midrule %
        
    \multirow{2}{*}{\rotatebox[origin=c]{90}{MB}}
    & EINCM~\cite{karmokar2025secrets}
    & \textbf{3.94} & \underline{2.58} & \underline{3.37} & \underline{2.36} & \underline{2.56} \\ 
        
    & MultiCM~\cite{shiba2022secrets}
    & \underline{4.44} & \textbf{2.39} & \textbf{3.32} & \textbf{2.05} & \textbf{2.35} \\
        
        
    \bottomrule
  \end{tabular}
  \end{adjustbox}
  \caption{The average end-point error (AEE) in pixels on the UEOF dataset
  (deep-water environment) across five scenes at $dt=1$. \textbf{Bold} indicates
  best, \underline{underline} indicates second-best.}
  \label{tab:varos_aee}
\end{table}

\begin{table*}
  \centering
  \begin{adjustbox}{max width=\textwidth}
  \begin{tabular}{%
    @{}c@{\thickspace}l 
       c c c c c c
       c c c c c c
       c c c@{}%
    }
    \toprule
    {}
    & \multirow{2}{*}{Methods}
    & \multicolumn{6}{c}{\texttt{scene1}}
    & \multicolumn{6}{c}{\texttt{scene2}}
    & \multicolumn{3}{c}{\texttt{scene3}}\\
        
    \cmidrule(lr){3-8} \cmidrule(lr){9-14} \cmidrule(lr){15-17} 
        
    {} & {}
    & A1PE ($\downarrow$)
    & A2PE ($\downarrow$)
    & A3PE ($\downarrow$)
    & A5PE ($\downarrow$)
    & A10PE ($\downarrow$)
    & A20PE ($\downarrow$)
    & A1PE ($\downarrow$)
    & A2PE ($\downarrow$)
    & A3PE ($\downarrow$)
    & A5PE ($\downarrow$)
    & A10PE ($\downarrow$)
    & A20PE ($\downarrow$)
    & A1PE ($\downarrow$)
    & A2PE ($\downarrow$)
    & A3PE ($\downarrow$)\\
        
    \midrule
        
    \multirow{2}{*}{\rotatebox[origin=c]{90}{LB}}
    & E-RAFT~\cite{gehrig2021eraft}
    & \textbf{59.24} & \underline{47.72} & \underline{42.30} & \underline{36.10} & 26.52 & 9.43
    & 97.9            & 92.70              & 85.10              & 66.80              & 23.50  & 4.00
    & 90.64           & 72.78              & 57.19 \\
        
    & MotionPriorCMax~\cite{hamann2024motion}
    & 83.13 & \textbf{30.53} & \textbf{21.38}   & \textbf{12.62} & \textbf{3.61}    & 1.20
    & 56.16 & \underline{18.28}           & \underline{8.10} & \textbf{3.25}  & \underline{1.06} & \underline{0.17}
    & 72.10  & \underline{24.20}            & \textbf{11.53} \\
        
    \midrule 
        
    \multirow{3}{*}{\rotatebox[origin=c]{90}{MB}}
    & EINCM~\cite{karmokar2025secrets}
    & \underline{69.50}  & 63.22              & 59.93   & 50.28   & 5.57  & \underline{0.01}
    & \underline{44.34}              & 26.67              & 20.40 & 13.66 & 8.42 & 7.01
    & \textbf{27.41} & \textbf{18.74} & \underline{14.88} \\
        
    & MultiCM~\cite{shiba2022secrets}
    & 75.11              & 73.59           & 71.19          & 61.17 & \underline{5.54} &  \textbf{$<$ 0.01}
    & \textbf{41.96} & \textbf{12.66} & \textbf{6.91} & 3.97  & 1.33              & 0.31
    & \underline{59.56}              & 35.00           & 23.31 \\

    & OPCM~\cite{karmokar2025inertia}
    & {--}       & {--}       & {--}       & {--}                  & {--}               & {--}
    & 58.47 & 25.14 & 11.17 & \underline{3.33} & \textbf{0.60} & \textbf{$<$ 0.01}
    & 72.80 & 33.45 & 19.66 \\
        

    \toprule
    {}
    & {}
    & \multicolumn{6}{c}{\texttt{scene4}}
    & \multicolumn{6}{c}{\texttt{scene5}}
    & \multicolumn{3}{c}{\texttt{scene3} (contd.)}\\
        
    \cmidrule(lr){3-8} \cmidrule(lr){9-14} \cmidrule(lr){15-17}
       
    {} & {}
    & A1PE ($\downarrow$)
    & A2PE ($\downarrow$)
    & A3PE ($\downarrow$)
    & A5PE ($\downarrow$)
    & A10PE ($\downarrow$)
    & A20PE ($\downarrow$)
    & A1PE ($\downarrow$)
    & A2PE ($\downarrow$)
    & A3PE ($\downarrow$)
    & A5PE ($\downarrow$)
    & A10PE ($\downarrow$)
    & A20PE ($\downarrow$)
    & A5PE ($\downarrow$)
    & A10PE ($\downarrow$)
    & A20PE ($\downarrow$)\\
        
    \midrule
        
    \multirow{2}{*}{\rotatebox[origin=c]{90}{LB}}
    & E-RAFT~\cite{gehrig2021eraft}
    & 96.26           & 91.34              & 85.62              & 68.21              & 19.78 & 4.16
    & \textbf{43.85} & \underline{36.71} & 33.70              & 29.71              & 23.52 & 20.07 
    & 42.11           & 20.20              & 10.23 \\
        
        
    & MotionPriorCMax~\cite{hamann2024motion}
    & 65.52           & \underline{15.67}            & \textbf{5.13}    & \textbf{2.25}  & \underline{0.37} & \textbf{0.02}
    & 83.09           & \textbf{20.60}  & \textbf{13.76}    & \textbf{0.98} & \textbf{0.35}    & \underline{0.24}
    & \textbf{5.19}  & \underline{1.47} & \underline{0.25} \\
        
    \midrule 
        
    \multirow{3}{*}{\rotatebox[origin=c]{90}{MB}}
    & EINCM~\cite{karmokar2025secrets}
    & \textbf{18.88} & \textbf{8.92} & \underline{6.22}  & 3.79  & 1.10  & 0.30
    & 75.61              & 66.22            & 58.84 & 35.92 & 4.11 & 0.88 
    & 10.60              & 6.20              & 1.60 \\
        
    & MultiCM~\cite{shiba2022secrets}
    & \underline{41.39}              & 18.27 & 9.23               & 3.80               & 0.58              & 0.20
    & \underline{69.33} & 40.26 & \underline{23.54} & \underline{10.00} & \underline{1.50} & \textbf{0.08}
    & 12.52              & 3.42  & 0.54 \\
        
        
    & OPCM~\cite{karmokar2025inertia}
    & 80.46            & 37.45          & 12.75 & \underline{2.41} & \textbf{0.25} & \underline{0.05}
    &      --            &        --        &    --   &           --       &     --           & --
    & \underline{8.29} & \textbf{1.17} & \textbf{0.05} \\
        
        
        
    \bottomrule
  \end{tabular}
  \end{adjustbox}
  \caption{The N-pixel error rates on the UEOF dataset (shallow-water
  environment) across five scenes at $dt=1$. \textbf{Bold} indicates best,
  \underline{underline} indicates second-best.}
  \label{tab:lofue_anpe}
\end{table*}

\begin{table*}
  \centering
  \begin{adjustbox}{max width=\textwidth}
  \begin{tabular}{%
      @{}c@{\thickspace}l 
      c c c c c c
      c c c c c c
      c c c@{}%
    }
    \toprule
    {}
    & \multirow{2}{*}{Methods}
    & \multicolumn{6}{c}{\texttt{scene1}}
    & \multicolumn{6}{c}{\texttt{scene2}}
    & \multicolumn{3}{c}{\texttt{scene3}}\\
        
    \cmidrule(lr){3-8} \cmidrule(lr){9-14} \cmidrule(lr){15-17} 
        
    {} & {}
    & A1PE ($\downarrow$)
    & A2PE ($\downarrow$)
    & A3PE ($\downarrow$)
    & A5PE ($\downarrow$)
    & A10PE ($\downarrow$)
    & A20PE ($\downarrow$)
    & A1PE ($\downarrow$)
    & A2PE ($\downarrow$)
    & A3PE ($\downarrow$)
    & A5PE ($\downarrow$)
    & A10PE ($\downarrow$)
    & A20PE ($\downarrow$)
    & A1PE ($\downarrow$)
    & A2PE ($\downarrow$)
    & A3PE ($\downarrow$)\\
        
    \midrule
        
    \multirow{2}{*}{\rotatebox[origin=c]{90}{LB}}
    & E-RAFT~\cite{gehrig2021eraft}
    & 72.43 & 51.13 & 38.96 & \textbf{23.72} & \textbf{8.63} & \underline{4.13}
    & 98.37 & 94.41 & 89.66 & 77.83 & 42.58 & 5.92
    & 59.61 & 41.97 & 31.88 \\
        
    & MotionPriorCMax~\cite{hamann2024motion}
    & 92.70 & 81.71 & 69.04 & 47.35 & 21.45 & 5.65
    & 84.83 & 70.28 & 59.20 & 42.94 & 16.66 & 2.20
    & 85.97 & 70.72 & 59.64 \\
        
    \midrule 
        
    \multirow{2}{*}{\rotatebox[origin=c]{90}{MB}}
    & EINCM~\cite{karmokar2025secrets}
    & \underline{60.92} & \underline{47.67} & \underline{38.76} & \underline{25.58} & \underline{10.44} & \textbf{2.95}
    & \textbf{41.39} & \textbf{30.39} & \textbf{24.40} & \underline{16.72} & \underline{6.75} & \underline{1.07}
    & \underline{50.15} & \underline{37.40} & \underline{30.55} \\
        
    & MultiCM~\cite{shiba2022secrets}
    & \textbf{60.38} & \textbf{44.94} & \textbf{36.86} & 27.62 & 13.81 & 4.34
    & \underline{47.33} & \underline{33.10} & \underline{24.67} & \textbf{14.48} & \textbf{4.52} & \textbf{0.60}
    & \textbf{48.66} & \textbf{36.03} & \textbf{29.19} \\
        

    \toprule
    {}
    & {}
    & \multicolumn{6}{c}{\texttt{scene4}}
    & \multicolumn{6}{c}{\texttt{scene5}}
    & \multicolumn{3}{c}{\texttt{scene3} (contd.)}\\
        
    \cmidrule(lr){3-8} \cmidrule(lr){9-14} \cmidrule(lr){15-17}
        
    {} & {}
    & A1PE ($\downarrow$)
    & A2PE ($\downarrow$)
    & A3PE ($\downarrow$)
    & A5PE ($\downarrow$)
    & A10PE ($\downarrow$)
    & A20PE ($\downarrow$)
    & A1PE ($\downarrow$)
    & A2PE ($\downarrow$)
    & A3PE ($\downarrow$)
    & A5PE ($\downarrow$)
    & A10PE ($\downarrow$)
    & A20PE ($\downarrow$)
    & A5PE ($\downarrow$)
    & A10PE ($\downarrow$)
    & A20PE ($\downarrow$)\\
        
    \midrule
       
    \multirow{2}{*}{\rotatebox[origin=c]{90}{LB}}
    & E-RAFT~\cite{gehrig2021eraft}
    & 69.48 & 53.99 & 47.92 & 40.25 & 16.08 & 4.06
    & 65.11 & 45.71 & 33.23 & 20.13 & 7.09 & 1.34
    & \textbf{19.44} & \textbf{7.96} & \underline{3.13} \\
        
    & MotionPriorCMax~\cite{hamann2024motion}
    & 87.87 & 78.54 & 72.07 & 60.35 & 26.36 & 3.35
    & 85.94 & 71.70 & 59.70 & 37.66 & 13.29 & 2.11
    & 42.71 & 18.86 & 4.94 \\
        
    \midrule 
        
    \multirow{2}{*}{\rotatebox[origin=c]{90}{MB}}
    & EINCM~\cite{karmokar2025secrets}
    & \textbf{37.46} & \underline{26.61} & \underline{20.38} & \underline{12.54} & \underline{4.97} & \underline{1.57}
    & \underline{47.20} & \underline{31.50} & \underline{23.80} & \underline{15.90} & \underline{6.45} & \underline{0.95}
    & 23.10 & 12.85 & 3.25 \\
        
    & MultiCM~\cite{shiba2022secrets}
    & \underline{39.79} & \textbf{25.98} & \textbf{18.83} & \textbf{11.68} & \textbf{4.88} & \textbf{0.54}
    & \textbf{45.87} & \textbf{30.02} & \textbf{22.44} & \textbf{14.06} & \textbf{5.28} & \textbf{0.51}
    & \underline{21.29} & \underline{10.73} & \textbf{2.12} \\
        
    \bottomrule
  \end{tabular}
  \end{adjustbox}
  \caption{The N-pixel error rates on the UEOF dataset (deep-water environment)
  across five scenes at $dt=1$. \textbf{Bold} indicates best,
  \underline{underline} indicates second-best.}
  \label{tab:varos_anpe}
\end{table*}

\section{Evaluation}
\label{sec:evaluation}
\subsection{Experimental Setup}
\label{subsec:experimental_setup}
We conducted experiments on all scenes of the UEOF dataset. The event-based
optical flow metrics include the average endpoint error (AEE) as well as
$\text{A}N\text{PE}$, which represents the percentage of pixels with an
endpoint error lower than $N$ pixels for $N \in \{1, 2, 3, 5, 10, 20\}$. To
ensure a comprehensive evaluation, we benchmarked the UEOF dataset against
representative state-of-the-art approaches categorized into two primary
paradigms: learning-based (LB) and model-based (MB) algorithms. The LB
techniques consist of supervised and unsupervised learning. Supervised
techniques, like E-RAFT~\cite{gehrig2021eraft}, leverage high-quality ground
truth for training.  Conversely, unsupervised learning algorithms, such as
MotionPriorCMax~\cite{hamann2024motion}, circumvent this by relying exclusively
on the raw event data itself. Conversely, the MB methods stray away from neural
networks and instead adopt traditional nonlinear optimization with contrast
maximization objectives.

To make the UEOF dataset more impactful, we first extended the existing
dataloaders of each evaluated technique. This involved adapting the algorithm's
input to correctly handle our specified data formats. In the evaluations of the
LB approaches, we utilized the checkpoints provided by each publication, which
were pretrained on the DSEC~\cite{gehrig2021dsec} dataset. The MB methods used
identical accumulated event window sizes for each respective scene. Otherwise,
we maintained the default configurations and parameters. The configuration
files and modified source code used for the evaluation are available on our
project website.

\subsection{Results}
\label{subsec:results}
The quantitative results for shallow-water and deep-water environments are
presented in \cref{tab:lofue_aee,tab:lofue_anpe} and
\cref{tab:varos_aee,tab:varos_anpe}, respectively. We note that the MB
approaches MultiCM~\cite{shiba2022secrets} and EINCM~\cite{karmokar2025secrets}
achieved the lowest error rates, with EINCM~\cite{karmokar2025secrets} reaching
a minimum AEE of 1.01\,px in \texttt{scene4} (shallow-water). Conversely, the
LB model E-RAFT~\cite{gehrig2021eraft} exhibited the highest errors, peaking at
16.32\,px in \texttt{scene5} (shallow-water). In the deep-water environment, we
observed a universal increase in error magnitudes across all techniques.

\newcommand{\imgscale}{0.2}
\newcommand{\rotboxscale}{0.2}
\newcommand{\spacer}{\phantom{ai}}
\begin{figure*}
  \centering
  \begin{adjustbox}{max width=0.9\textwidth}
  \begin{tabular}{@{}c@{\thinspace}c@{\thinspace}c@{\thinspace}c@{\thinspace}|@{\thinspace}c@{\thinspace}c@{}}
    \toprule
    \multicolumn{6}{c}{Shallow-Water Environment}\\
    \rotatebox[origin=l]{90}{\begin{adjustbox}{max width=\rotboxscale\textwidth}
    \spacer\texttt{scene1}
  \end{adjustbox}}
  & \includegraphics[width=\imgscale\textwidth,cfbox=gray 0.1pt 0pt]{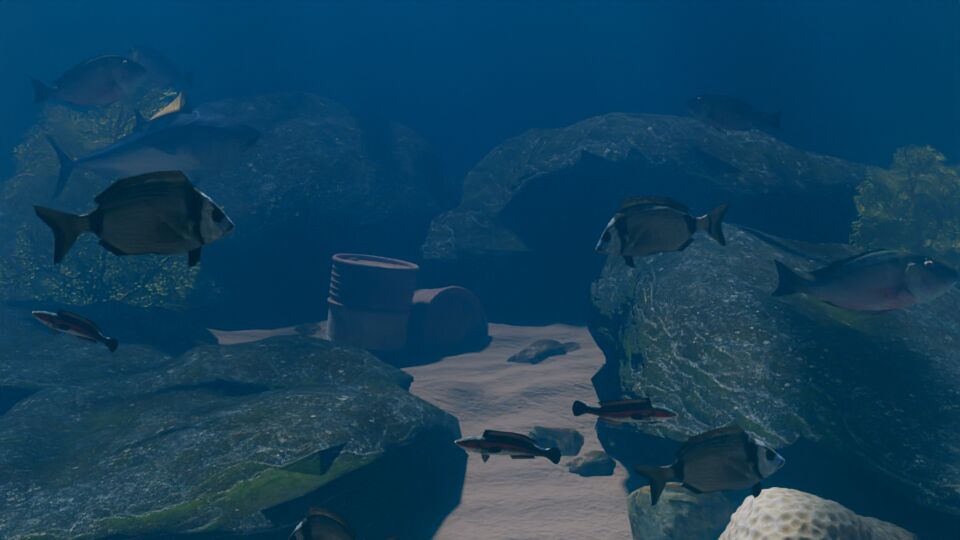}%
  & \includegraphics[width=\imgscale\textwidth,cfbox=gray 0.1pt 0pt]{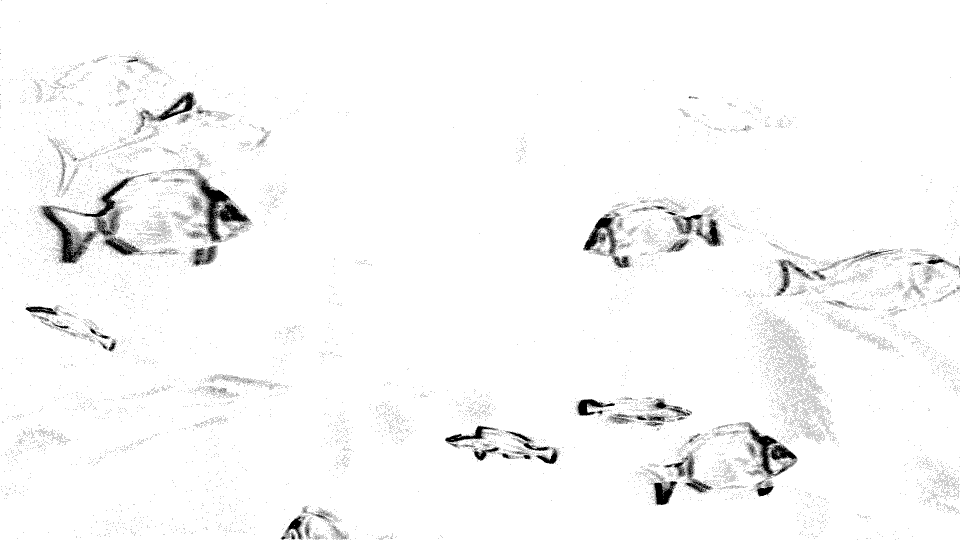}%
  & \includegraphics[width=\imgscale\textwidth,cfbox=gray 0.1pt 0pt]{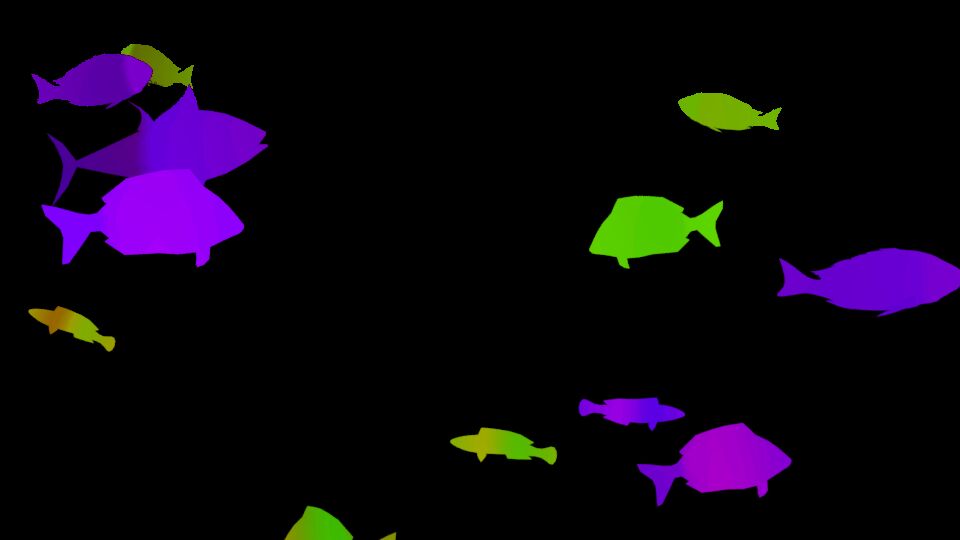}%
  & \includegraphics[width=\imgscale\textwidth,cfbox=gray 0.1pt 0pt]{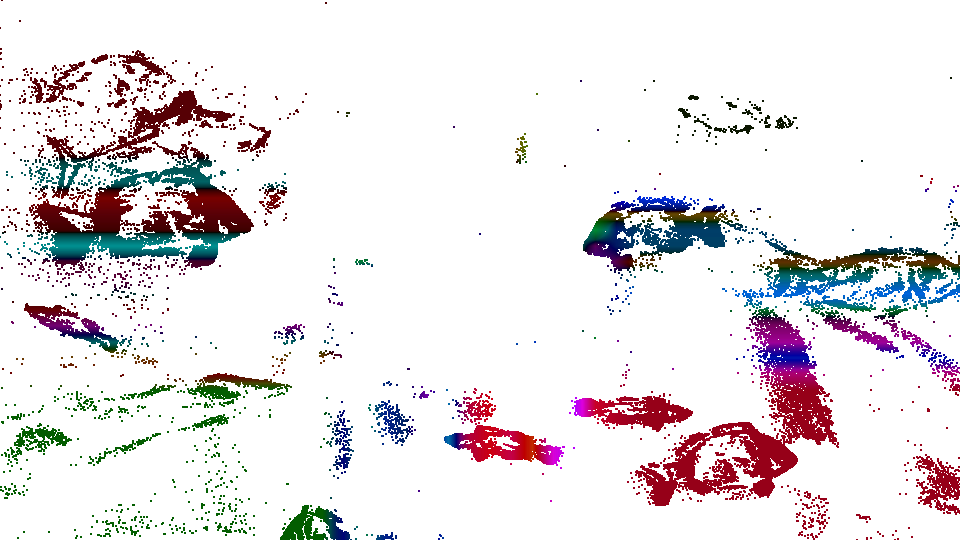}%
  & \includegraphics[width=\imgscale\textwidth,cfbox=gray 0.1pt 0pt]{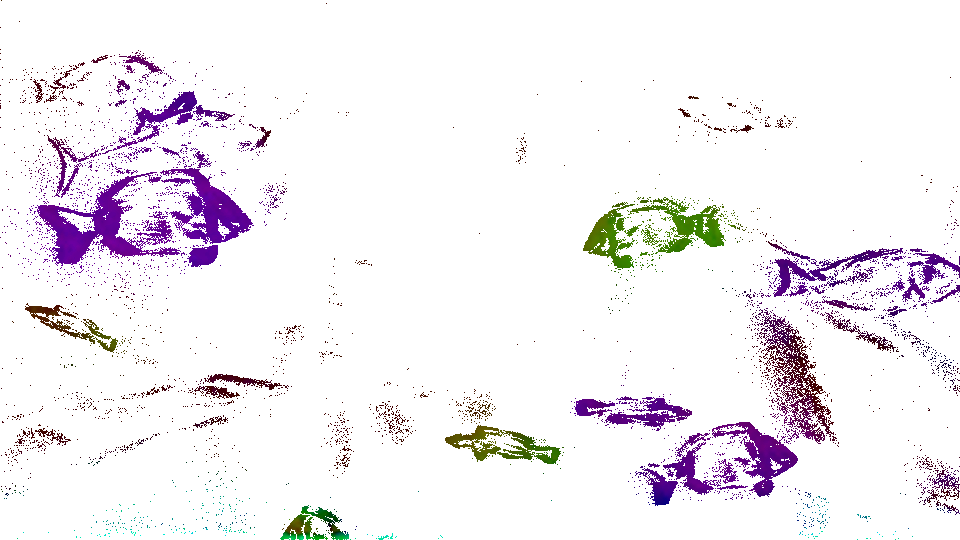}\\[-2pt]

  \rotatebox[origin=l]{90}{\begin{adjustbox}{max width=\rotboxscale\textwidth}
    \spacer\texttt{scene2}
  \end{adjustbox}}
  & \includegraphics[width=\imgscale\textwidth,cfbox=gray 0.1pt 0pt]{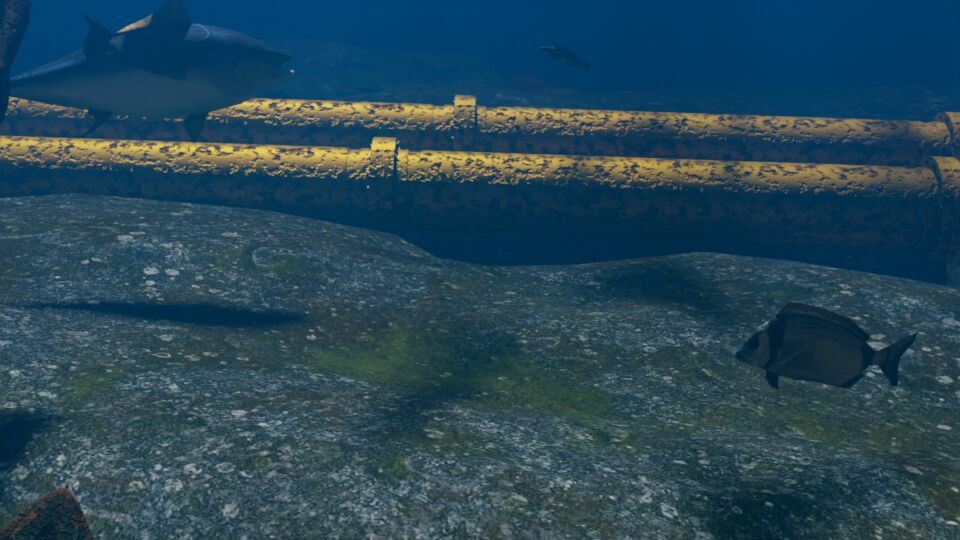}%
  & \includegraphics[width=\imgscale\textwidth,cfbox=gray 0.1pt 0pt]{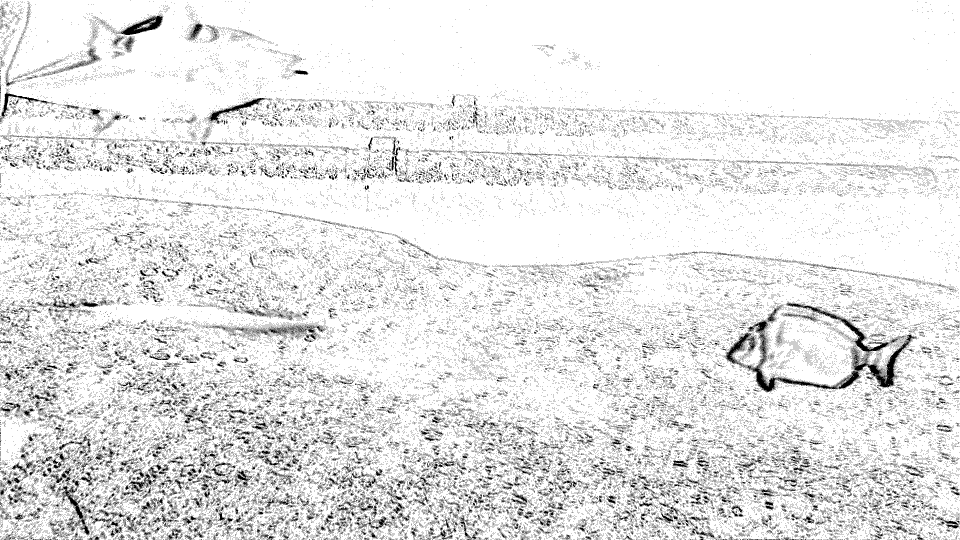}%
  & \includegraphics[width=\imgscale\textwidth,cfbox=gray 0.1pt 0pt]{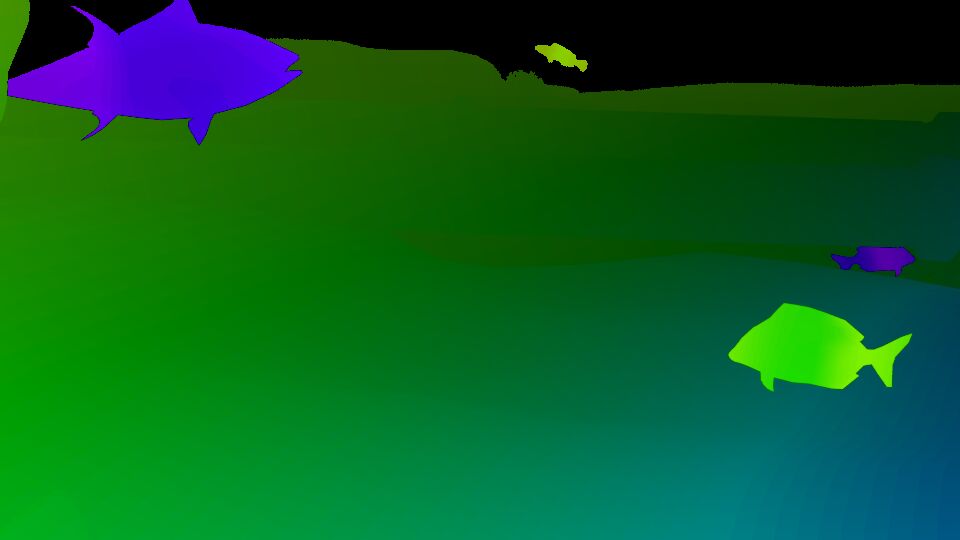}%
  & \includegraphics[width=\imgscale\textwidth,cfbox=gray 0.1pt 0pt]{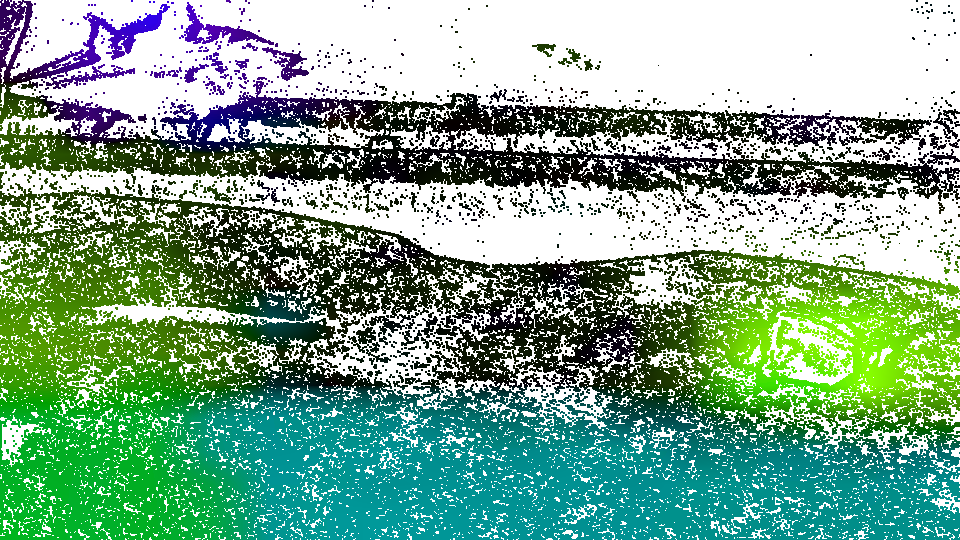}%
  & \includegraphics[width=\imgscale\textwidth,cfbox=gray 0.1pt 0pt]{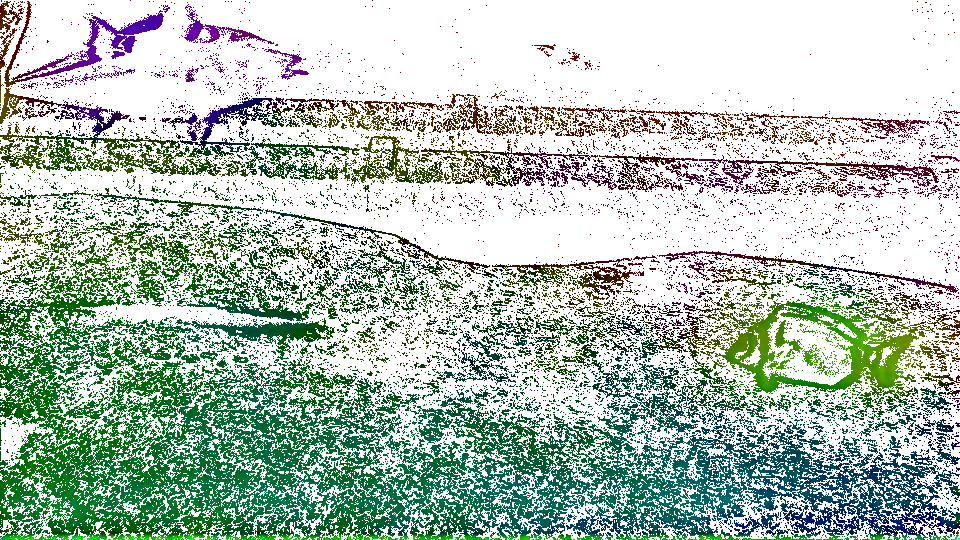}\\[-2pt]
      
  \rotatebox[origin=l]{90}{\begin{adjustbox}{max width=\rotboxscale\textwidth}
    \spacer\texttt{scene3}
  \end{adjustbox}}
  & \includegraphics[width=\imgscale\textwidth,cfbox=gray 0.1pt 0pt]{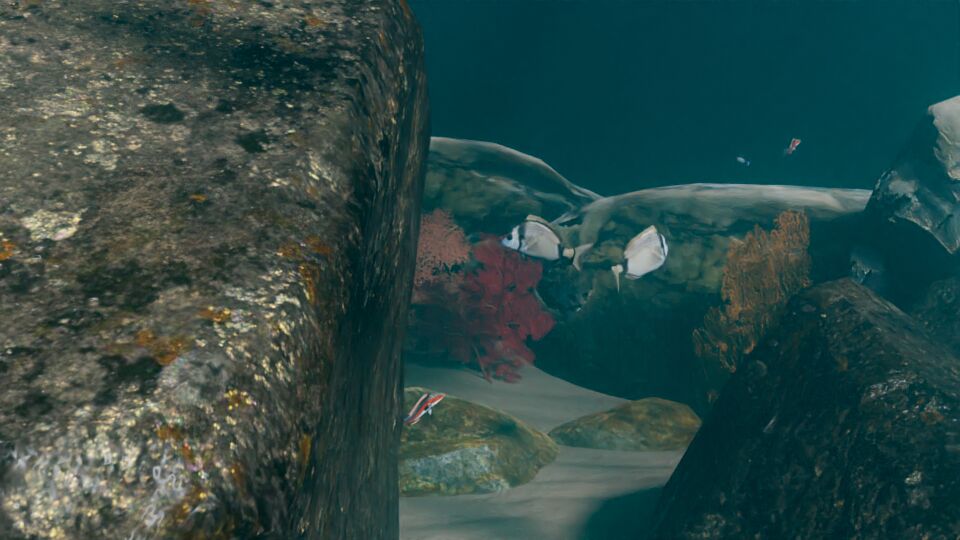}%
  & \includegraphics[width=\imgscale\textwidth,cfbox=gray 0.1pt 0pt]{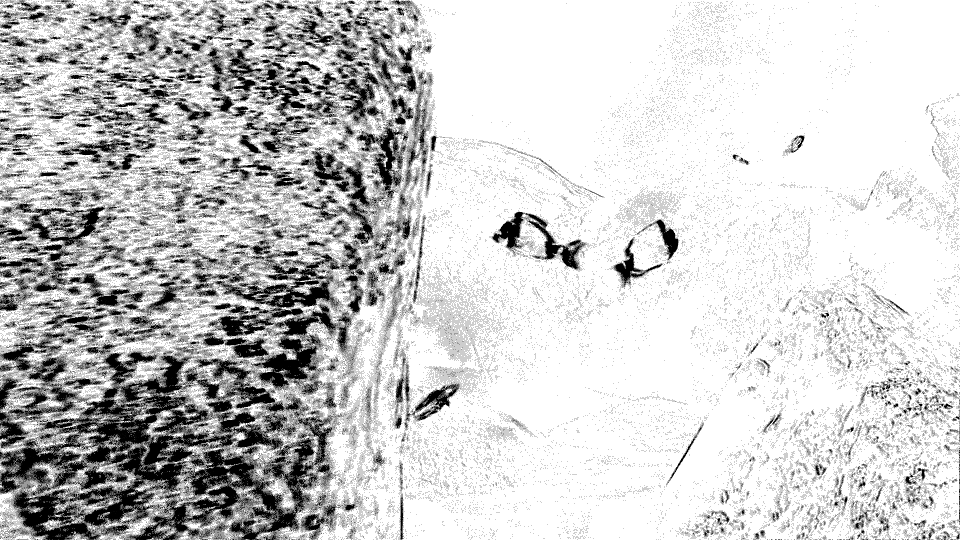}%
  & \includegraphics[width=\imgscale\textwidth,cfbox=gray 0.1pt 0pt]{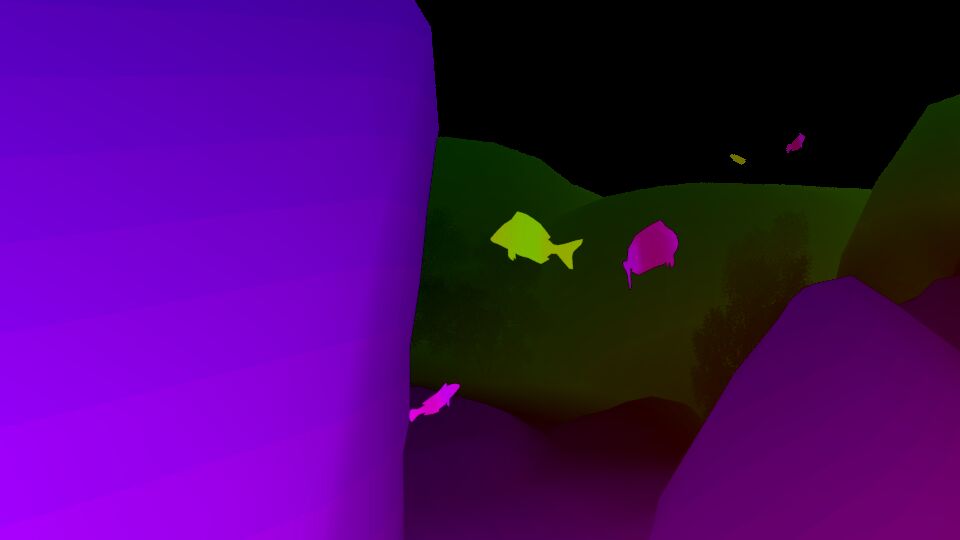}%
  & \includegraphics[width=\imgscale\textwidth,cfbox=gray 0.1pt 0pt]{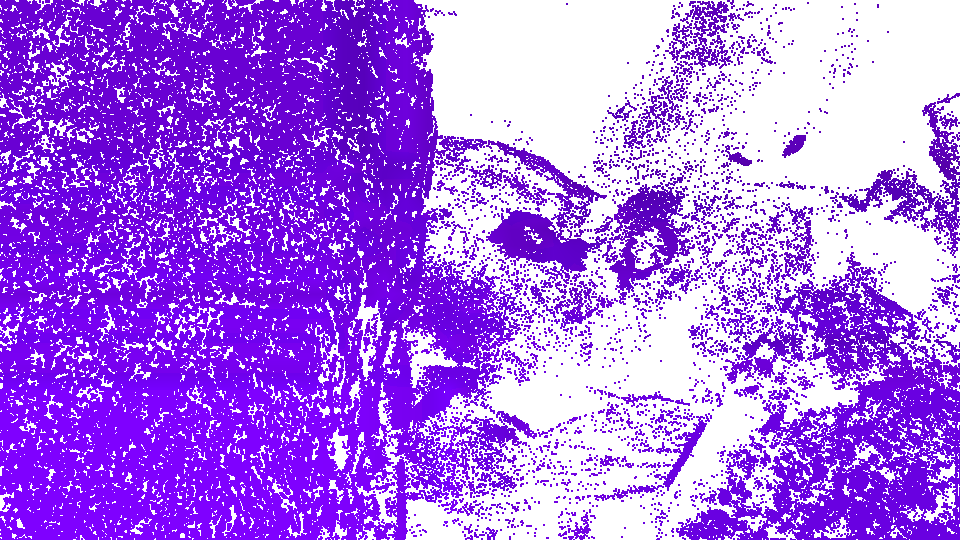}%
  & \includegraphics[width=\imgscale\textwidth,cfbox=gray 0.1pt 0pt]{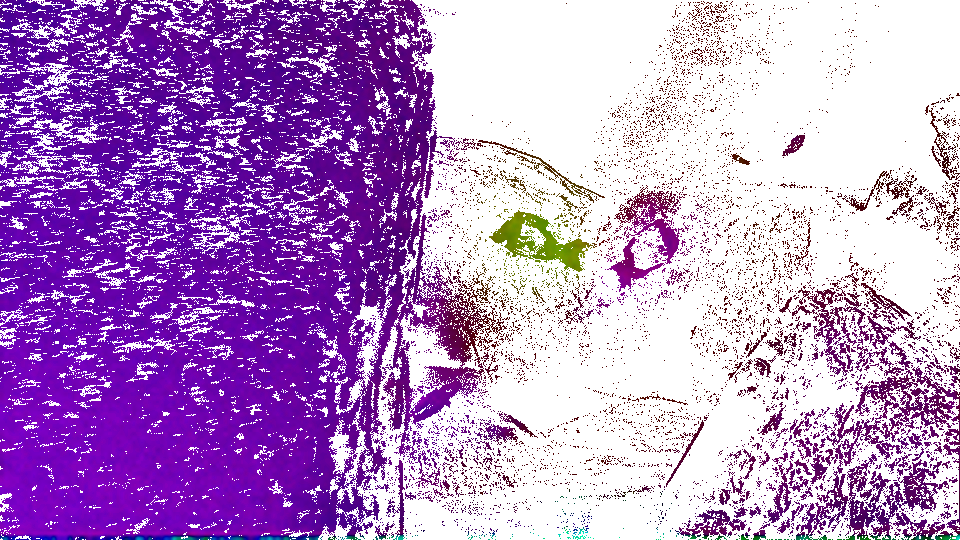}\\[-2pt]
        
  \rotatebox[origin=l]{90}{\begin{adjustbox}{max width=\rotboxscale\textwidth}
    \spacer\texttt{scene4}
  \end{adjustbox}}
  & \includegraphics[width=\imgscale\textwidth,cfbox=gray 0.1pt 0pt]{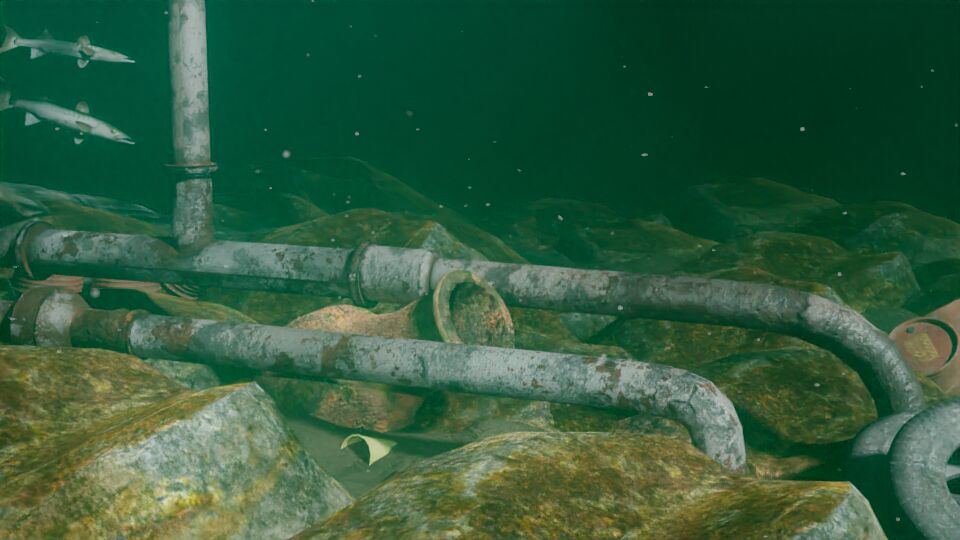}%
  & \includegraphics[width=\imgscale\textwidth,cfbox=gray 0.1pt 0pt]{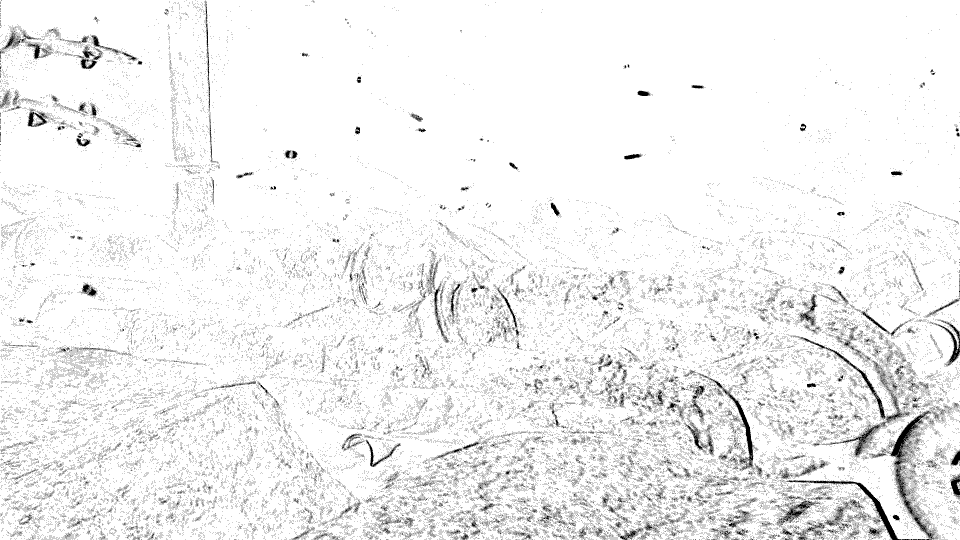}%
  & \includegraphics[width=\imgscale\textwidth,cfbox=gray 0.1pt 0pt]{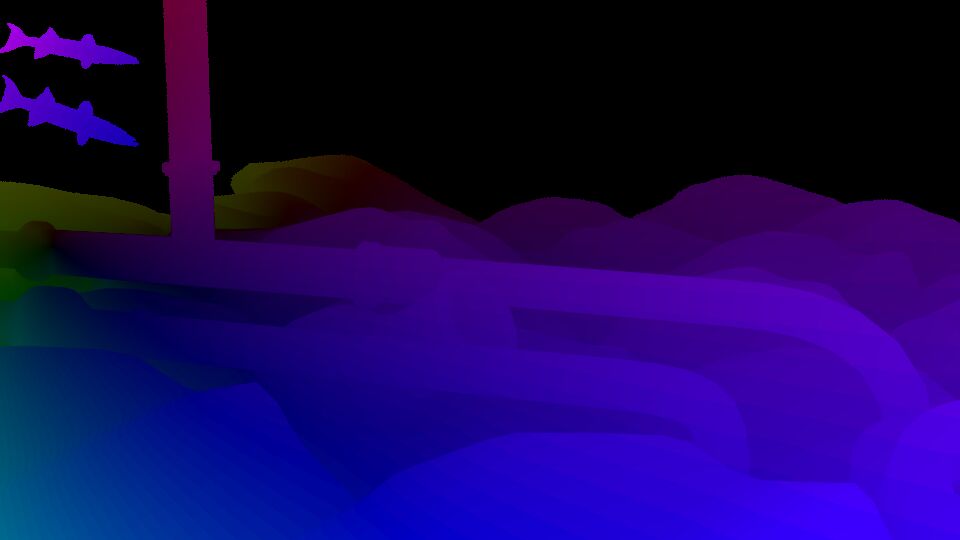}%
  & \includegraphics[width=\imgscale\textwidth,cfbox=gray 0.1pt 0pt]{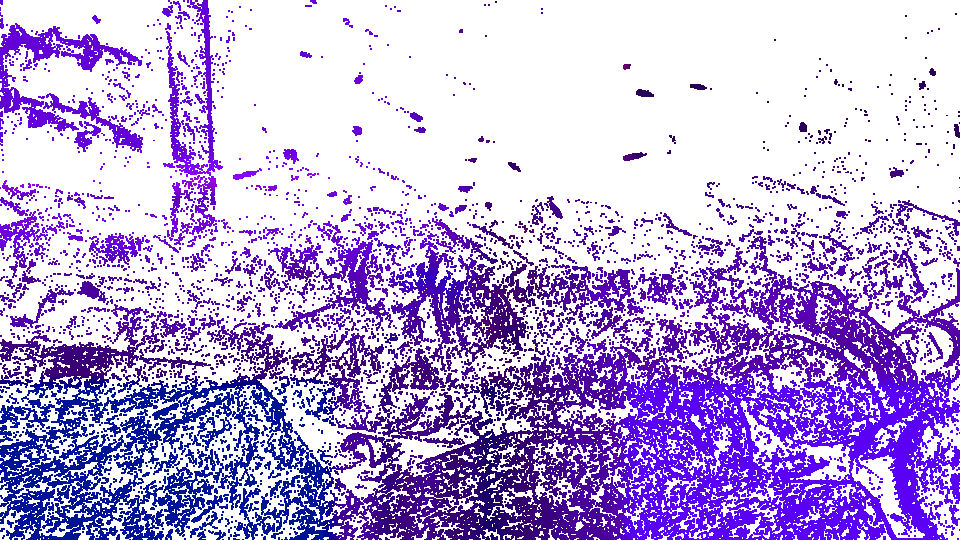}%
  & \includegraphics[width=\imgscale\textwidth,cfbox=gray 0.1pt 0pt]{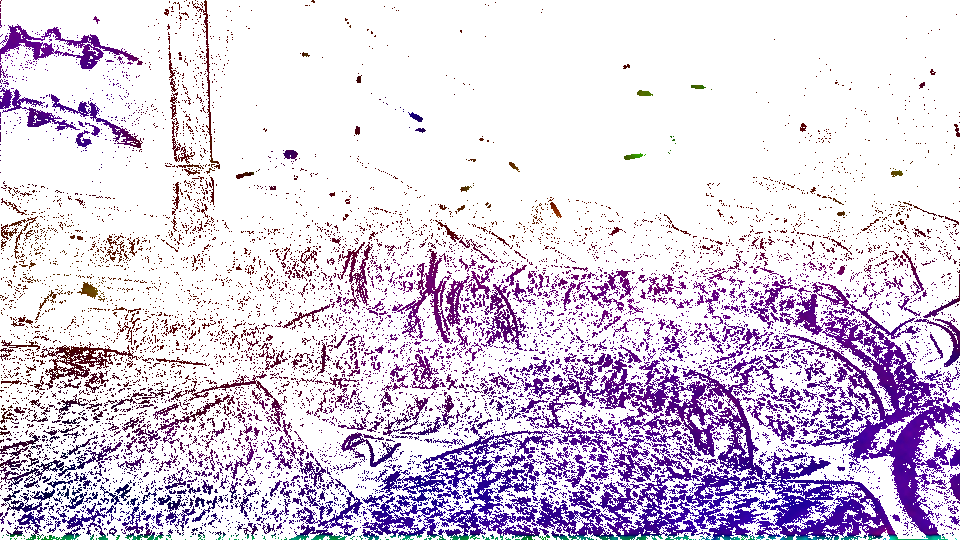}\\[-2pt]
        
  \rotatebox[origin=l]{90}{\begin{adjustbox}{max width=\rotboxscale\textwidth}
    \spacer\texttt{scene5}
  \end{adjustbox}}
  & \includegraphics[width=\imgscale\textwidth,cfbox=gray 0.1pt 0pt]{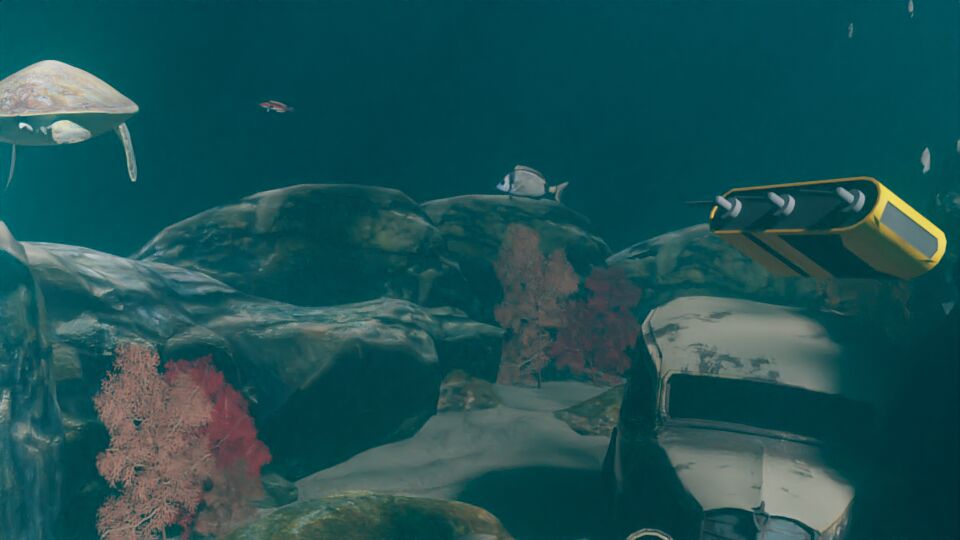}%
  & \includegraphics[width=\imgscale\textwidth,cfbox=gray 0.1pt 0pt]{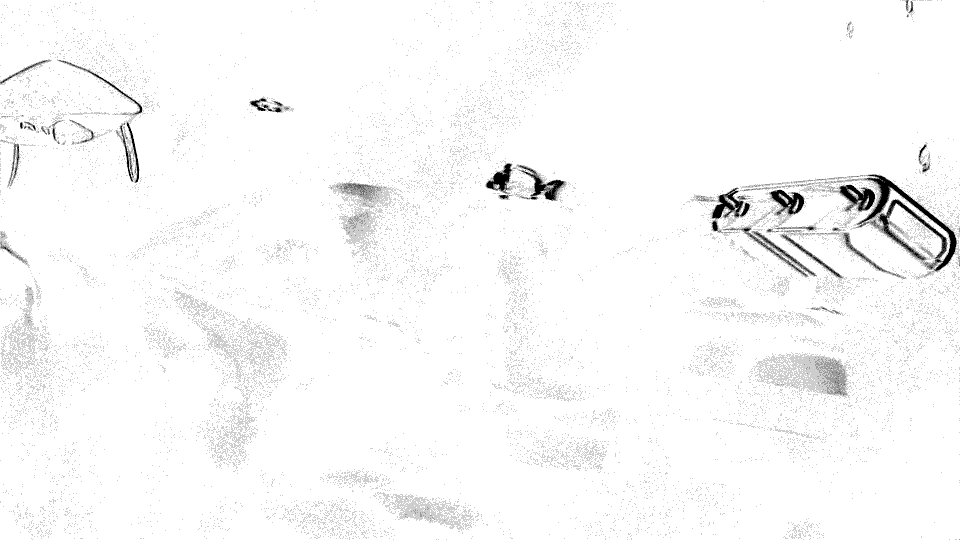}%
  & \includegraphics[width=\imgscale\textwidth,cfbox=gray 0.1pt 0pt]{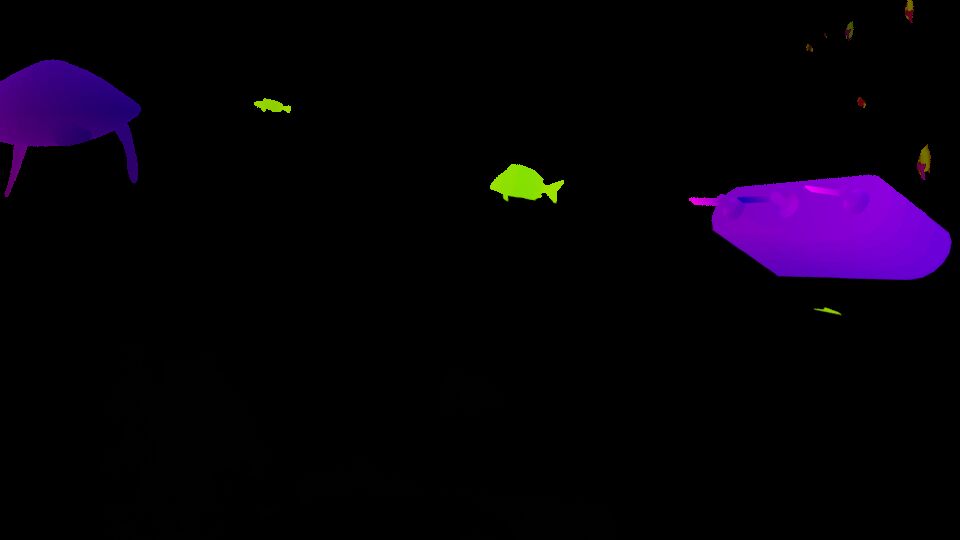}%
  & \includegraphics[width=\imgscale\textwidth,cfbox=gray 0.1pt 0pt]{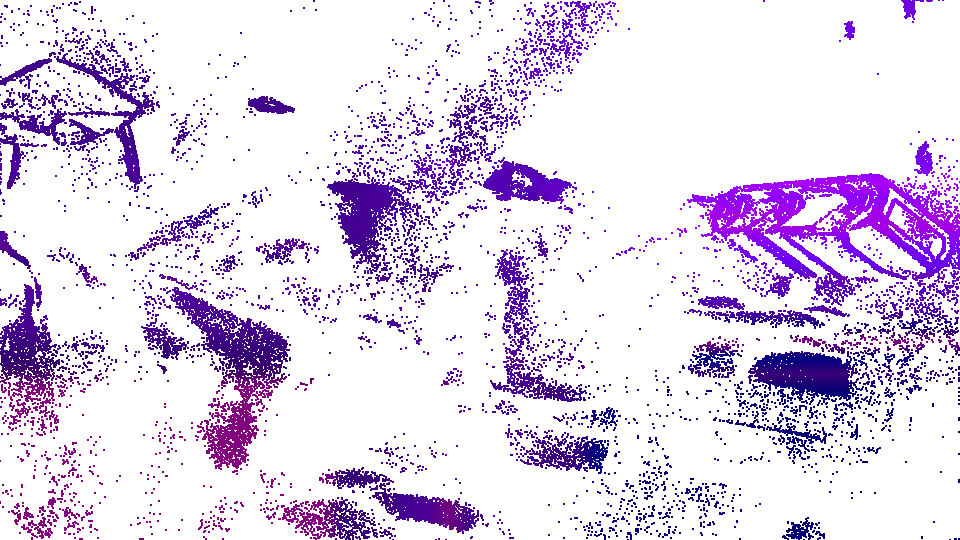}%
  & \includegraphics[width=\imgscale\textwidth,cfbox=gray 0.1pt 0pt]{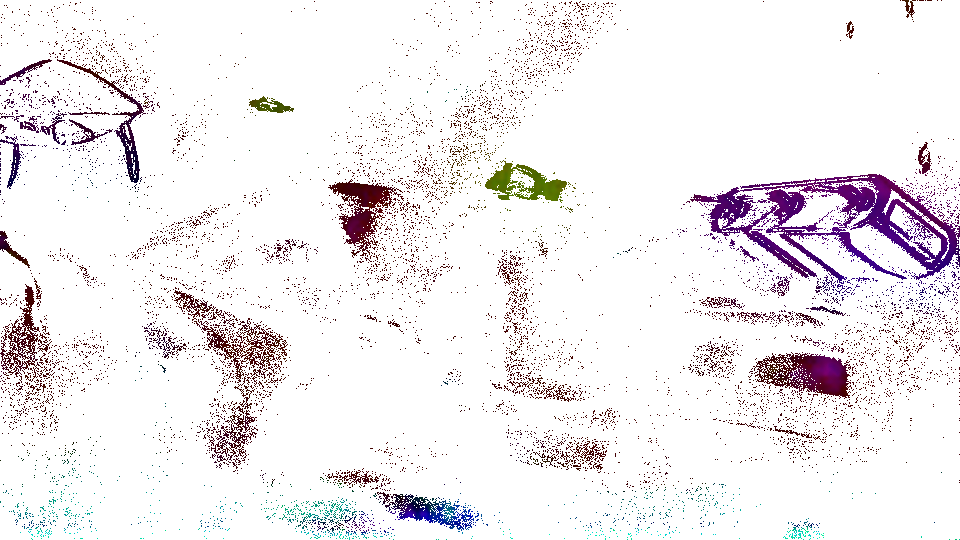}\\

  \toprule
  \multicolumn{6}{c}{Deep-Water Environment}\\
  \rotatebox[origin=l]{90}{\begin{adjustbox}{max width=\rotboxscale\textwidth}
    \spacer\texttt{scene1}
  \end{adjustbox}}
  & \includegraphics[width=\imgscale\textwidth,cfbox=gray 0.1pt 0pt]{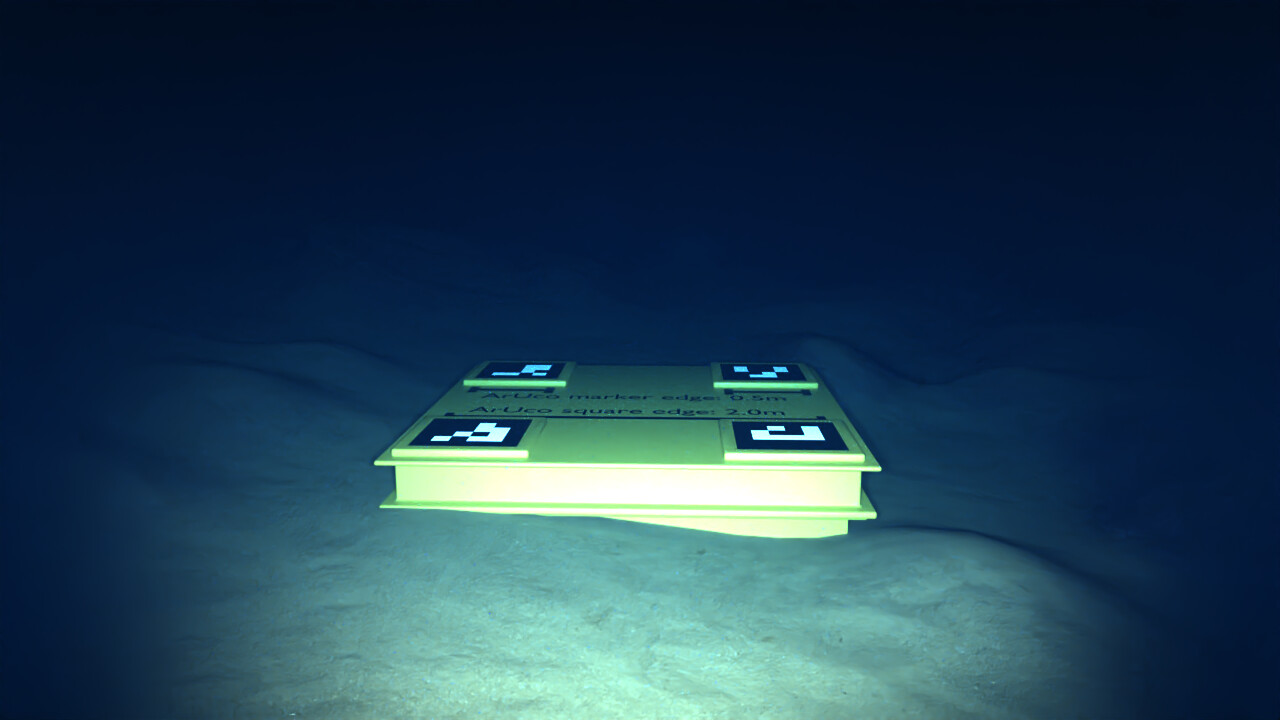}%
  & \includegraphics[width=\imgscale\textwidth,cfbox=gray 0.1pt 0pt]{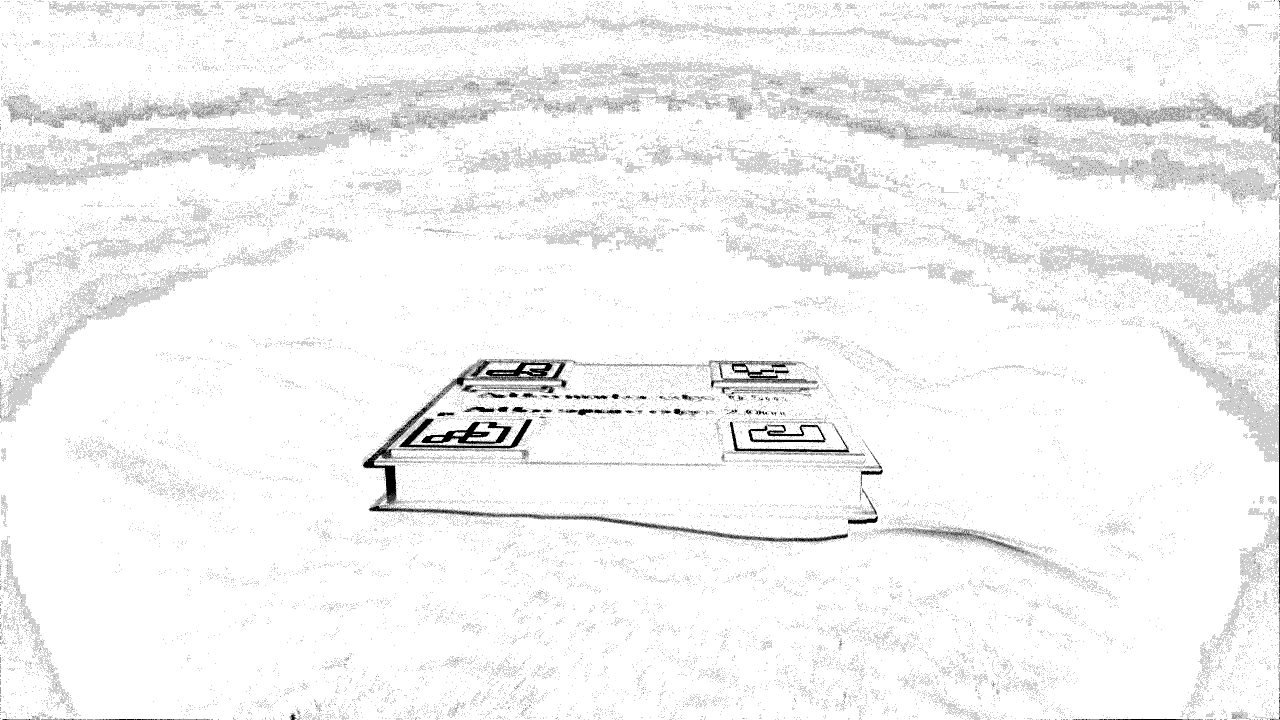}%
  & \includegraphics[width=\imgscale\textwidth,cfbox=gray 0.1pt 0pt]{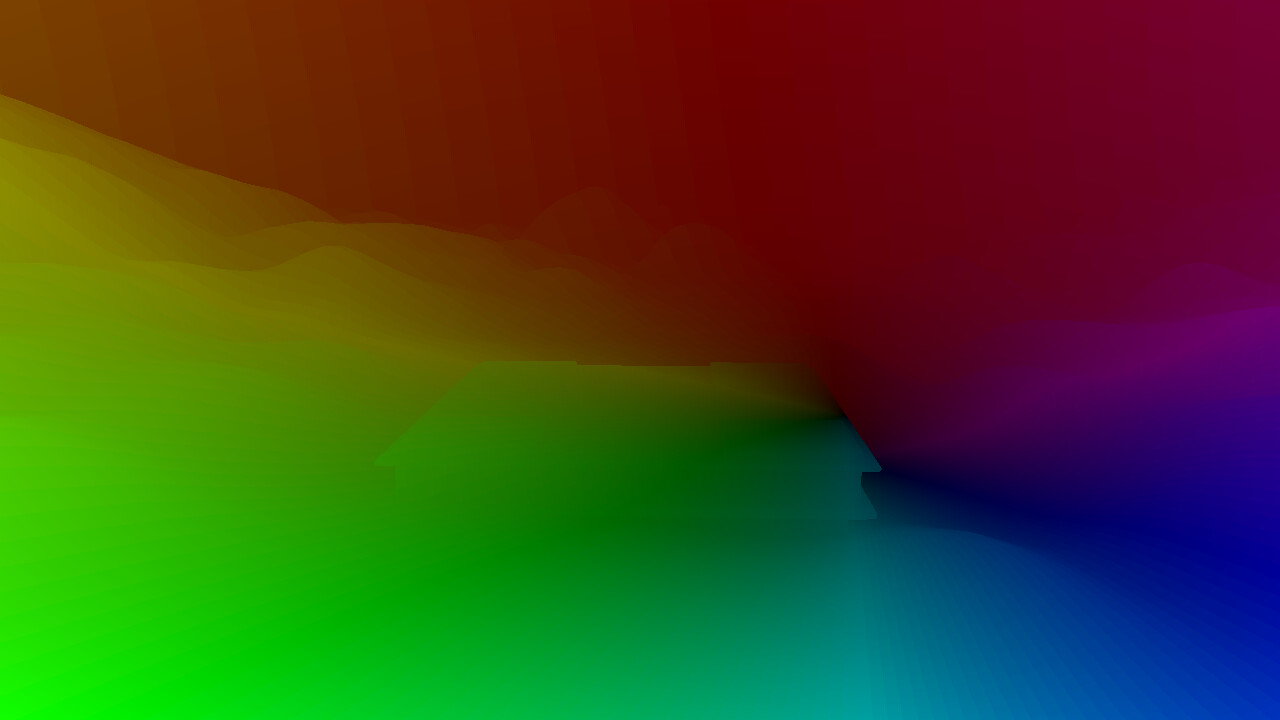}%
  & \includegraphics[width=\imgscale\textwidth,cfbox=gray 0.1pt 0pt]{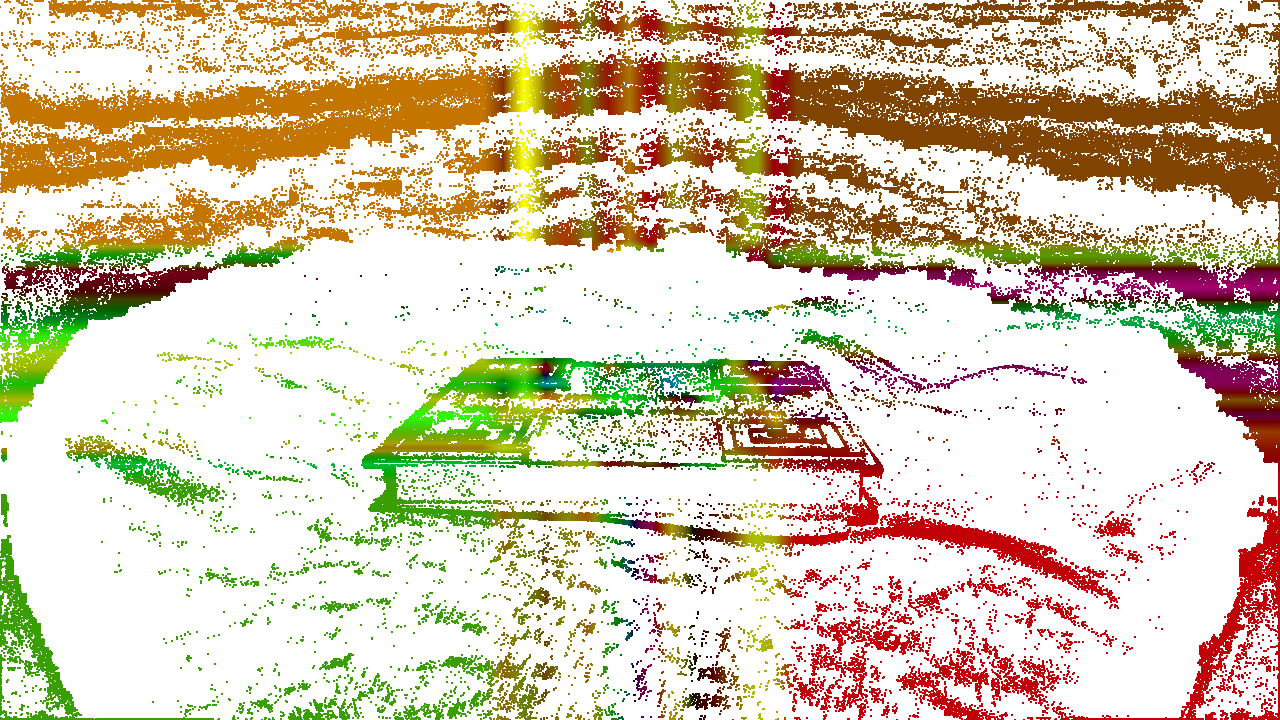}%
  & \includegraphics[width=\imgscale\textwidth,cfbox=gray 0.1pt 0pt]{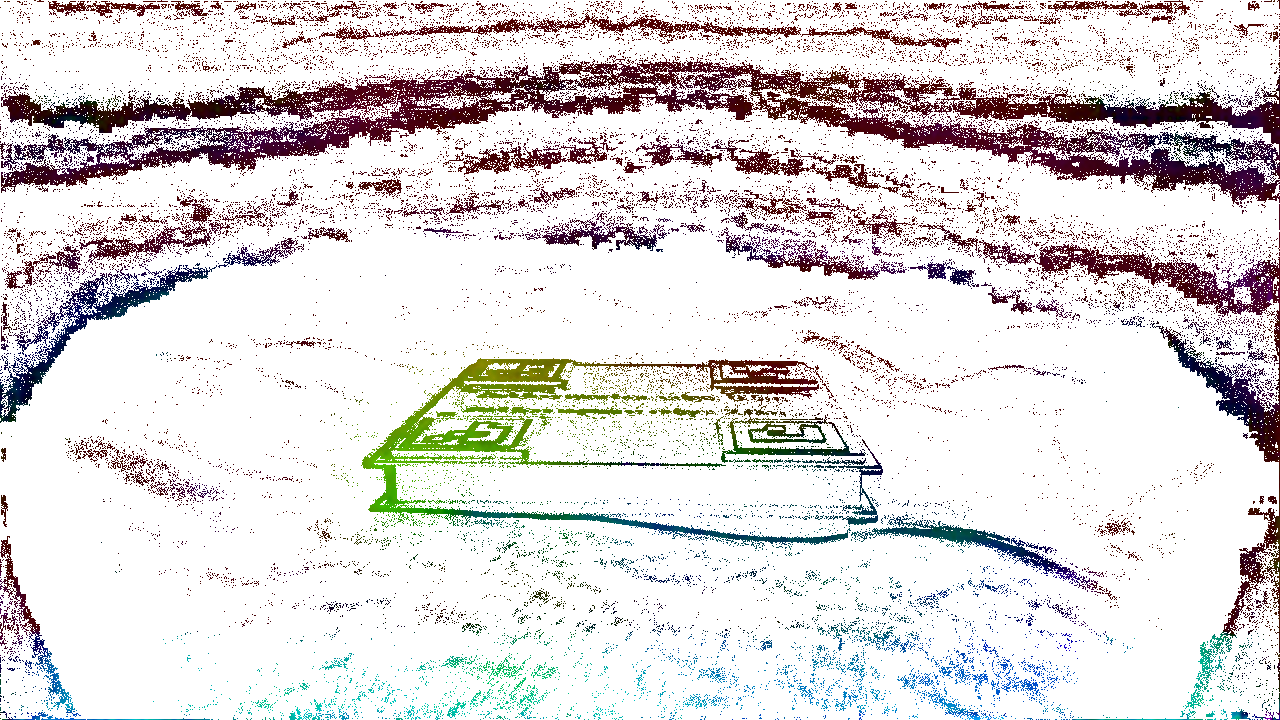}\\[-2pt]
        
  \rotatebox[origin=l]{90}{\begin{adjustbox}{max width=\rotboxscale\textwidth}
    \spacer\texttt{scene2}
  \end{adjustbox}}
  & \includegraphics[width=\imgscale\textwidth,cfbox=gray 0.1pt 0pt]{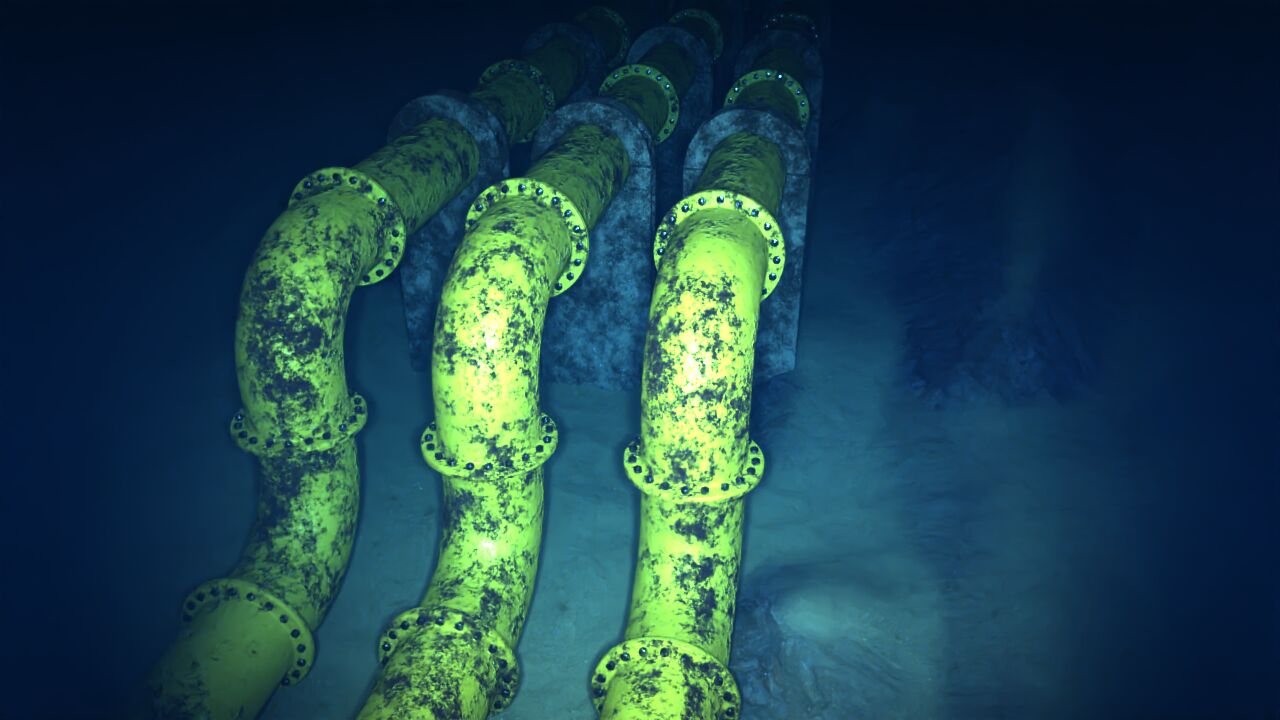}%
  & \includegraphics[width=\imgscale\textwidth,cfbox=gray 0.1pt 0pt]{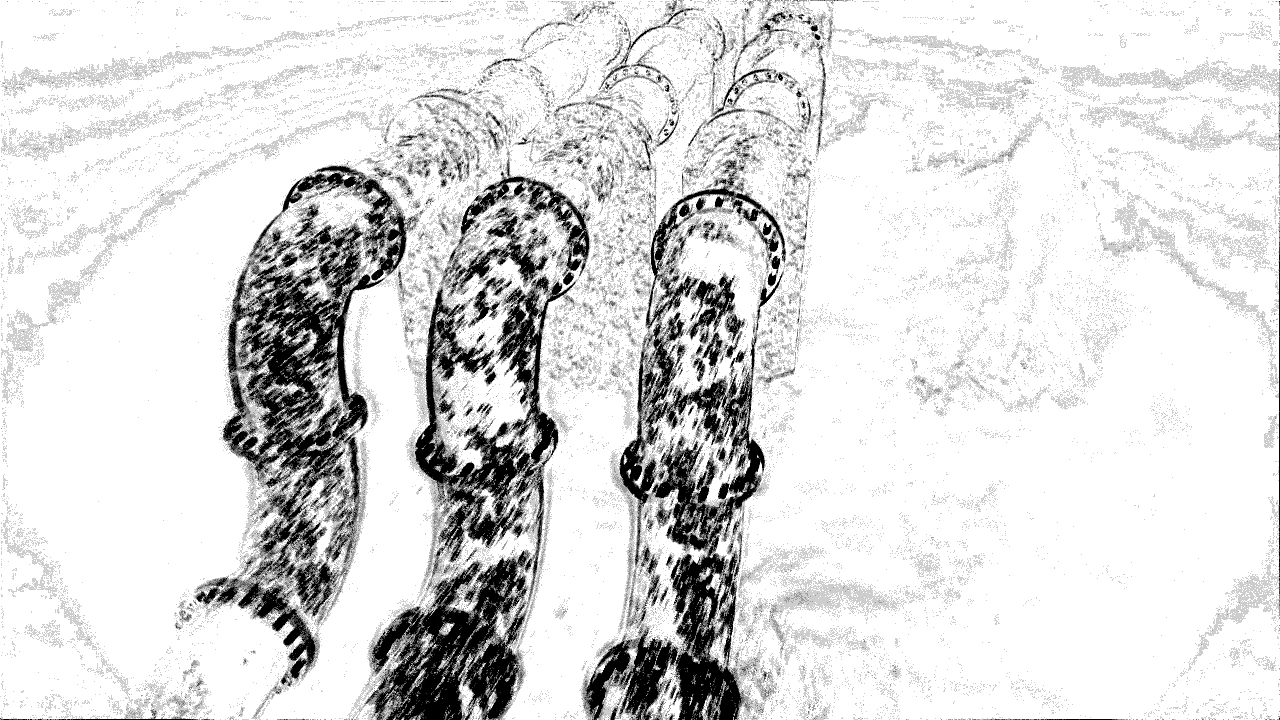}%
  & \includegraphics[width=\imgscale\textwidth,cfbox=gray 0.1pt 0pt]{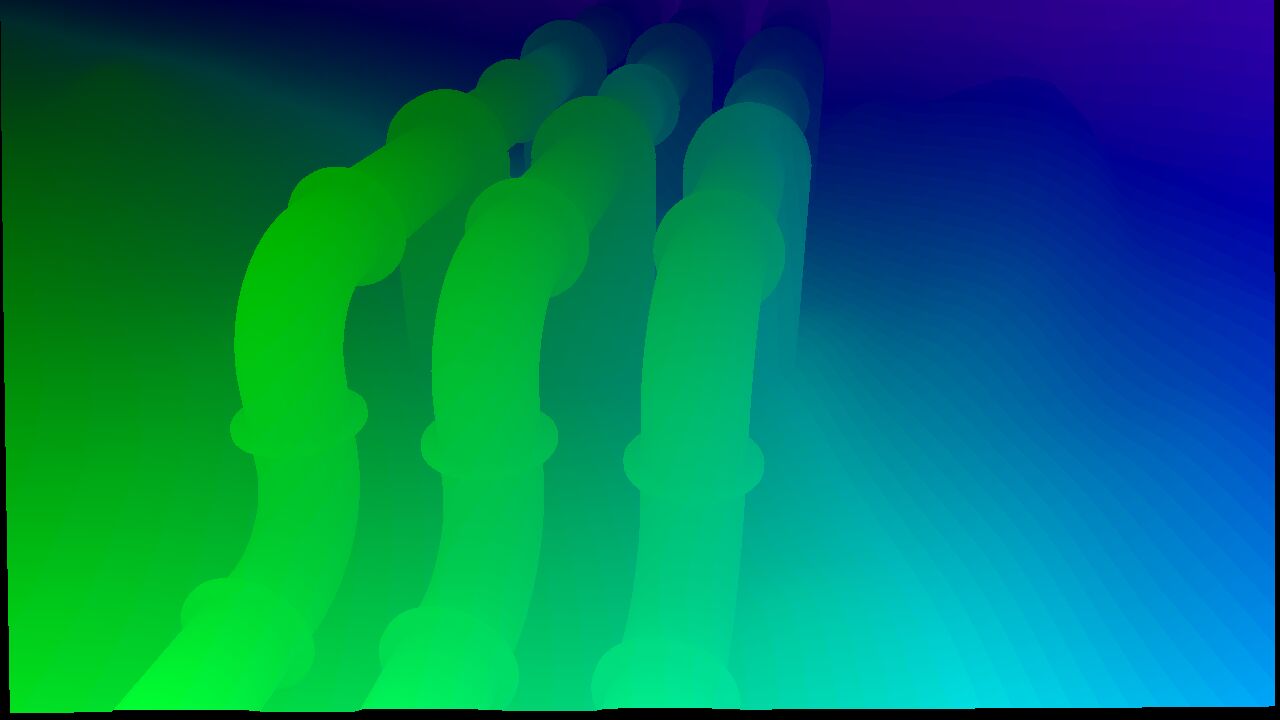}%
  & \includegraphics[width=\imgscale\textwidth,cfbox=gray 0.1pt 0pt]{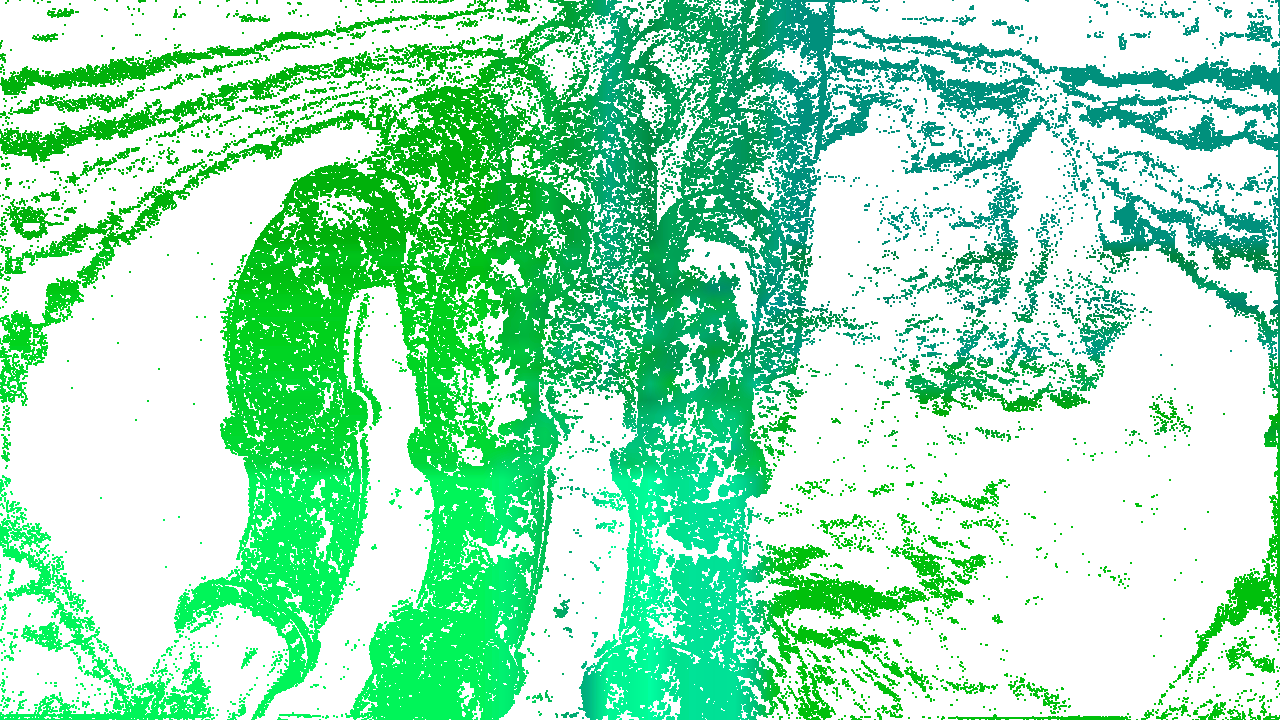}%
  & \includegraphics[width=\imgscale\textwidth,cfbox=gray 0.1pt 0pt]{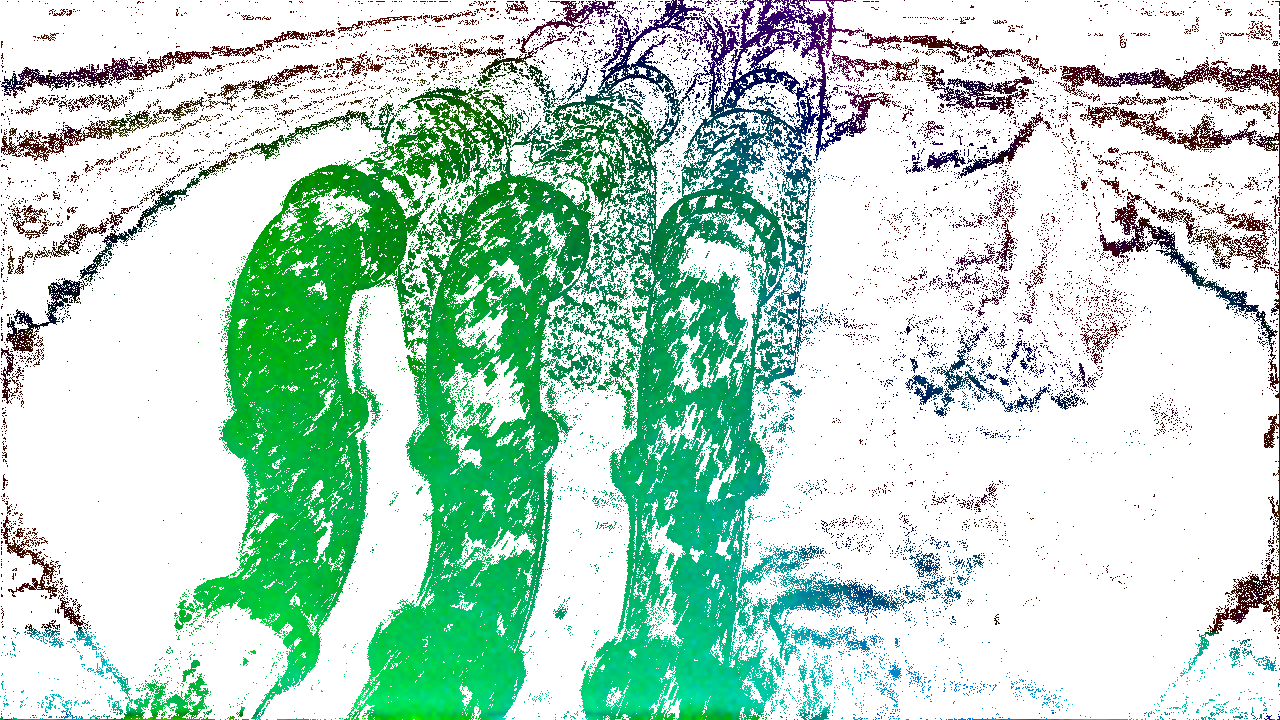}\\[-2pt]
        
  \rotatebox[origin=l]{90}{\begin{adjustbox}{max width=\rotboxscale\textwidth}
    \spacer\texttt{scene3}
  \end{adjustbox}}
  & \includegraphics[width=\imgscale\textwidth,cfbox=gray 0.1pt 0pt]{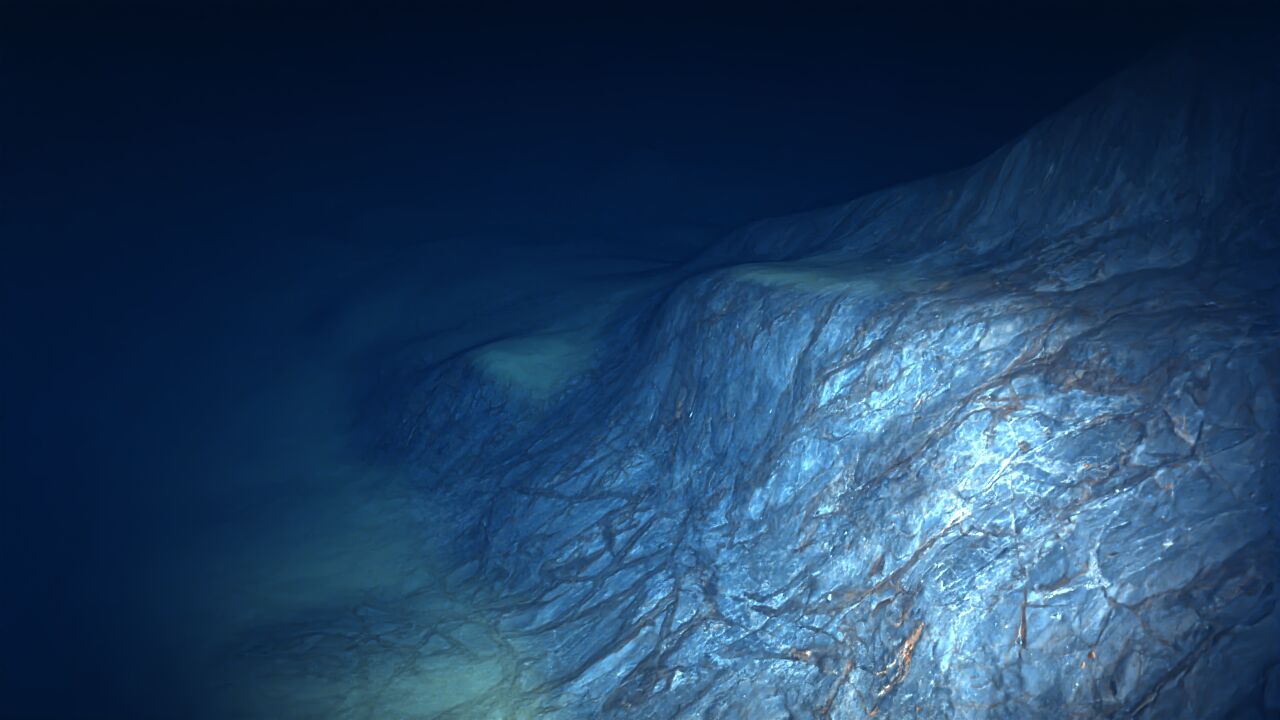}%
  & \includegraphics[width=\imgscale\textwidth,cfbox=gray 0.1pt 0pt]{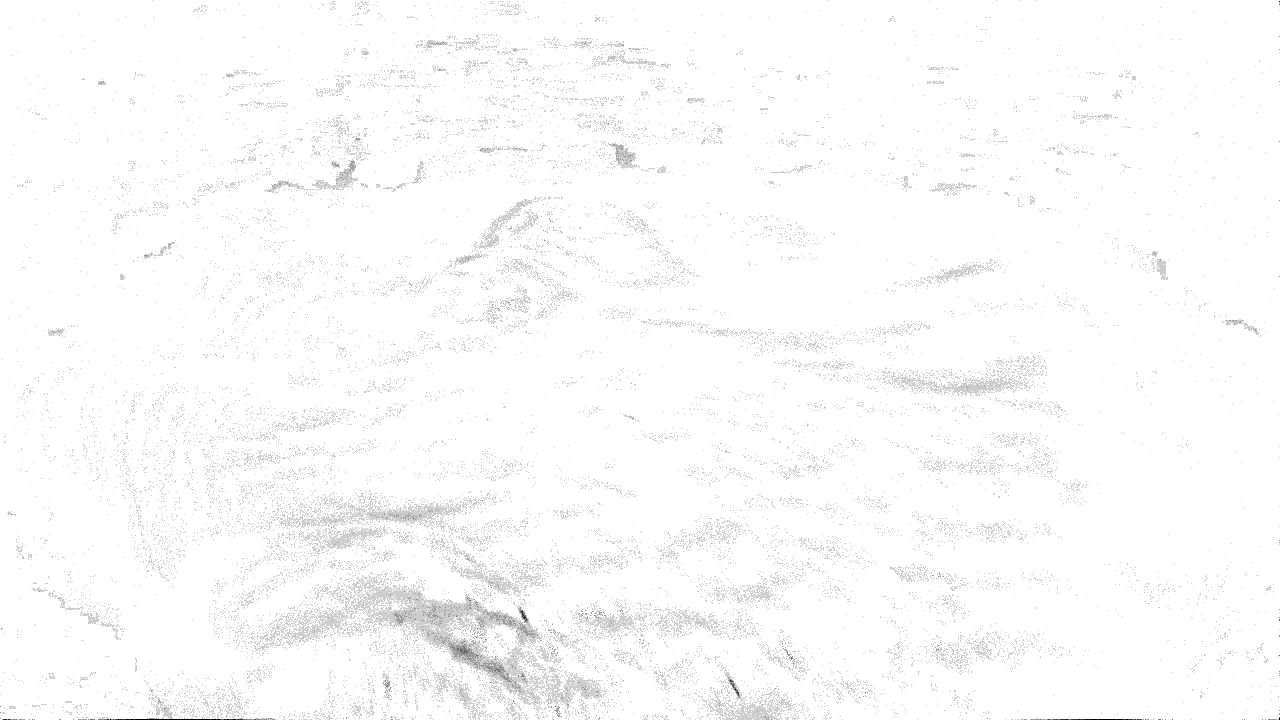}%
  & \includegraphics[width=\imgscale\textwidth,cfbox=gray 0.1pt 0pt]{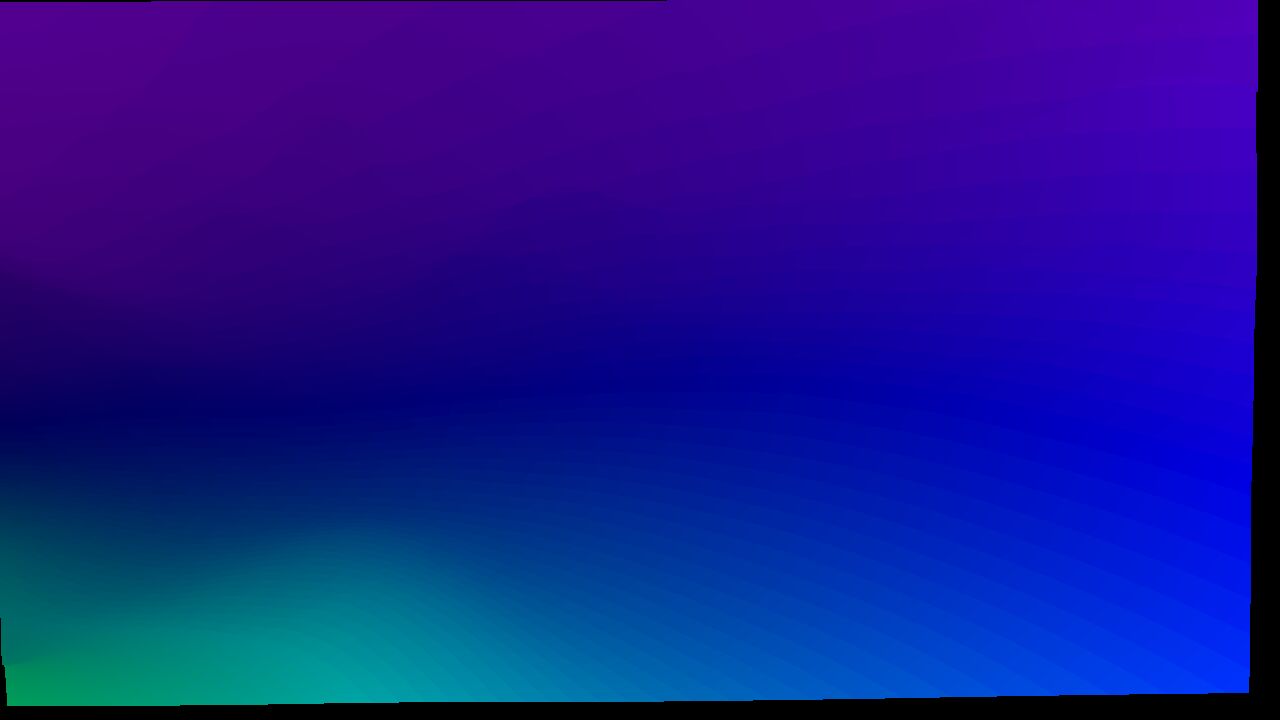}%
  & \includegraphics[width=\imgscale\textwidth,cfbox=gray 0.1pt 0pt]{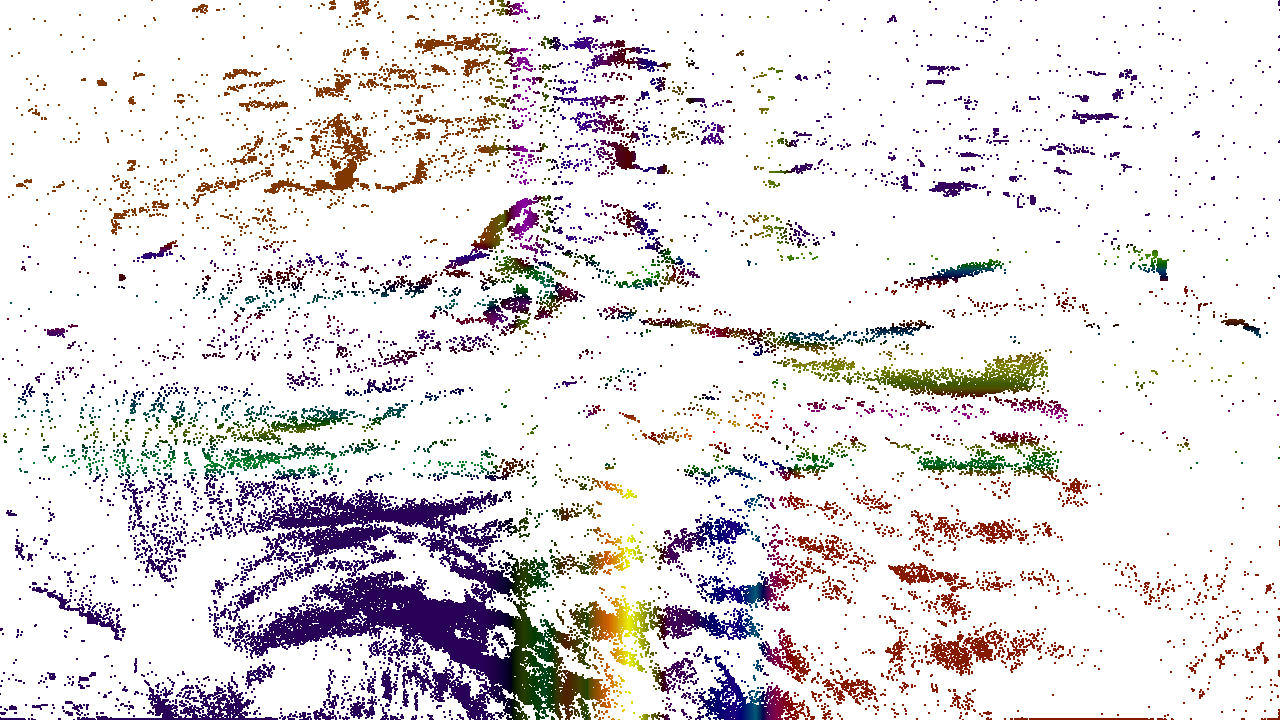}%
  & \includegraphics[width=\imgscale\textwidth,cfbox=gray 0.1pt 0pt]{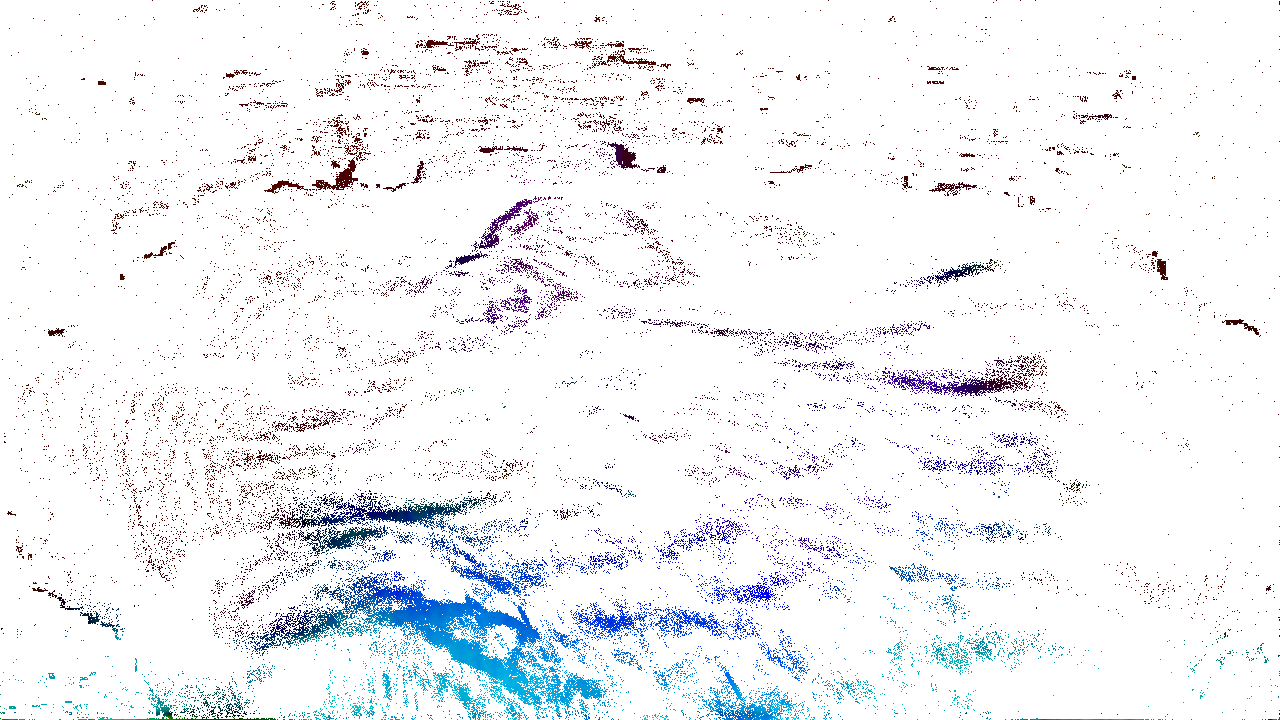}\\[-2pt]
        
  \rotatebox[origin=l]{90}{\begin{adjustbox}{max width=\rotboxscale\textwidth}
    \spacer\texttt{scene4}
  \end{adjustbox}}
  & \includegraphics[width=\imgscale\textwidth,cfbox=gray 0.1pt 0pt]{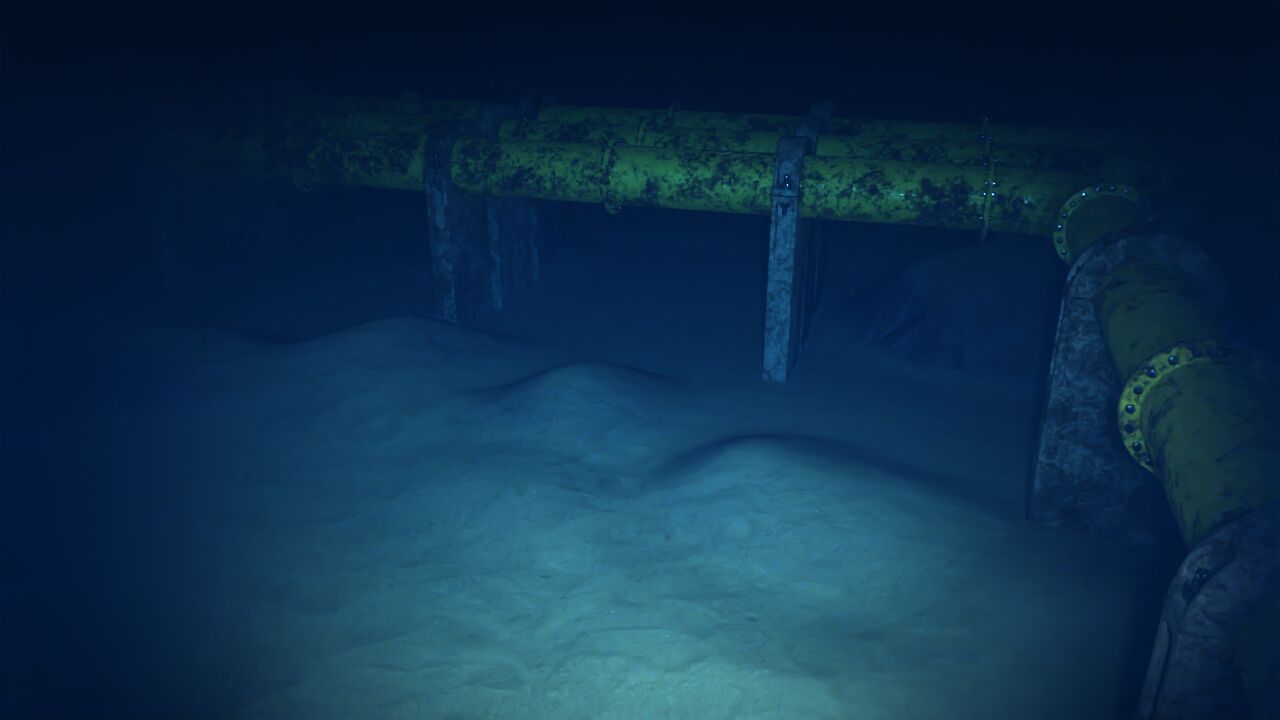}%
  & \includegraphics[width=\imgscale\textwidth,cfbox=gray 0.1pt 0pt]{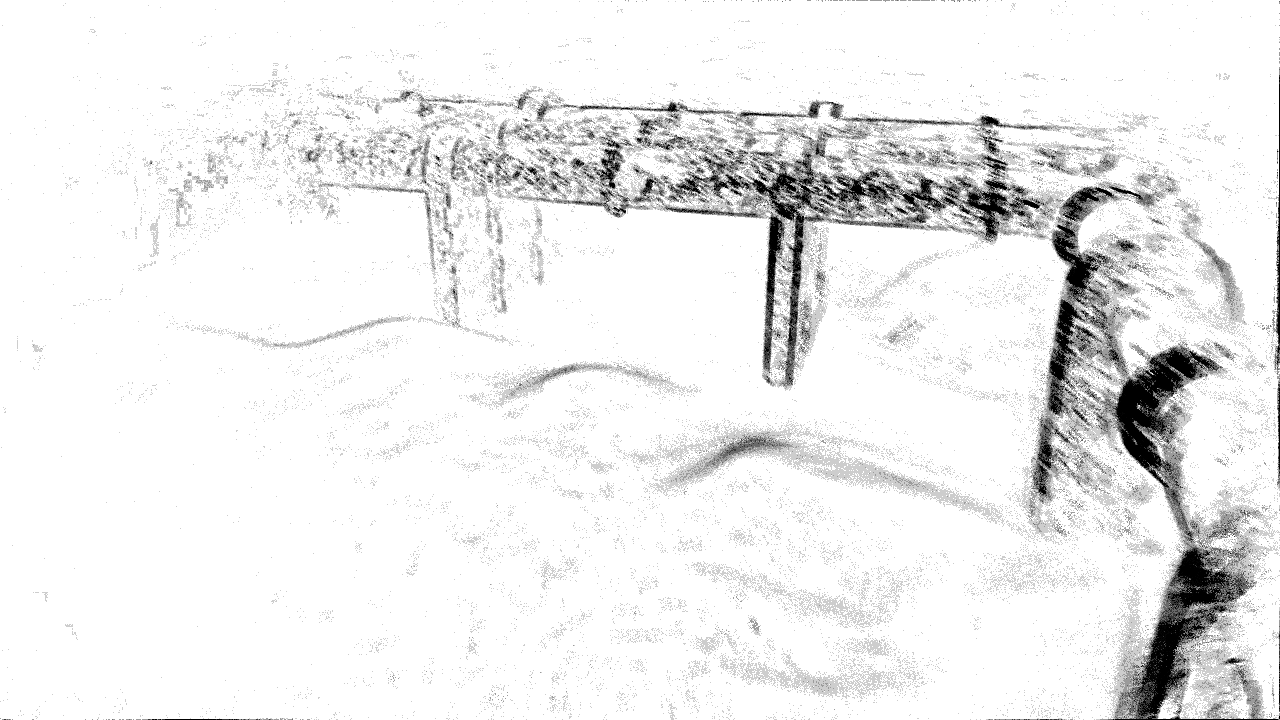}%
  & \includegraphics[width=\imgscale\textwidth,cfbox=gray 0.1pt 0pt]{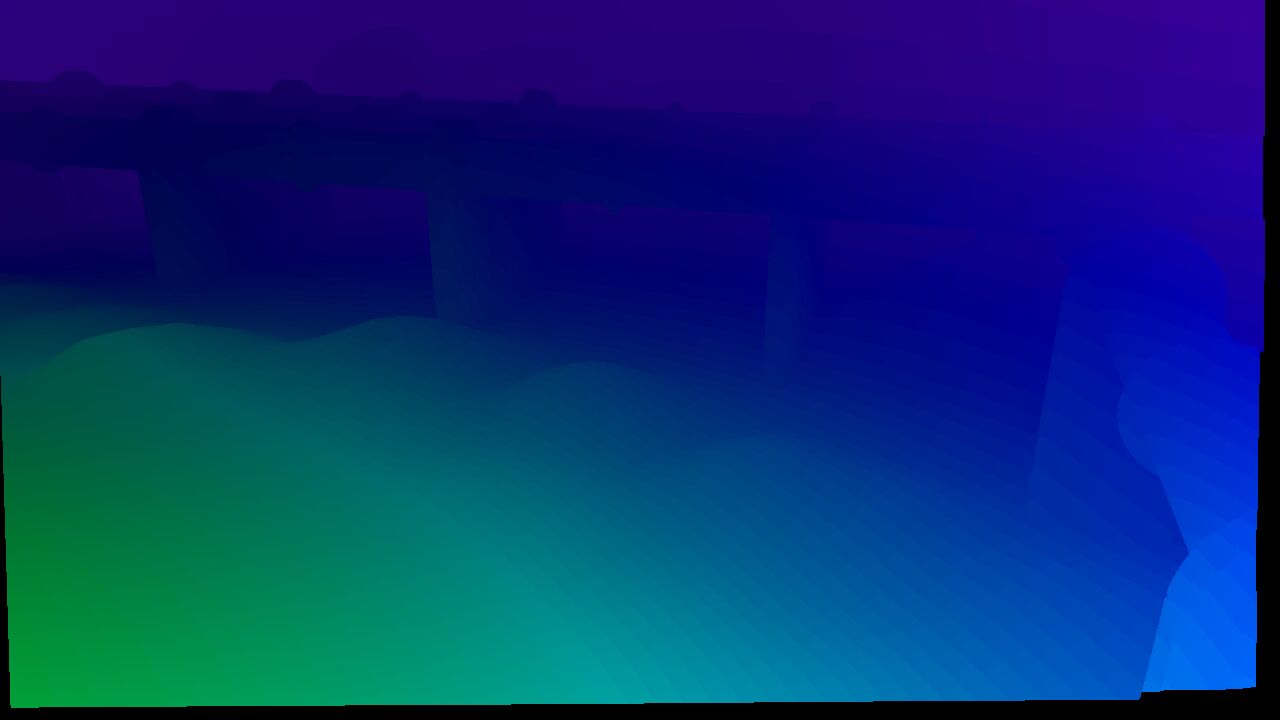}%
  & \includegraphics[width=\imgscale\textwidth,cfbox=gray 0.1pt 0pt]{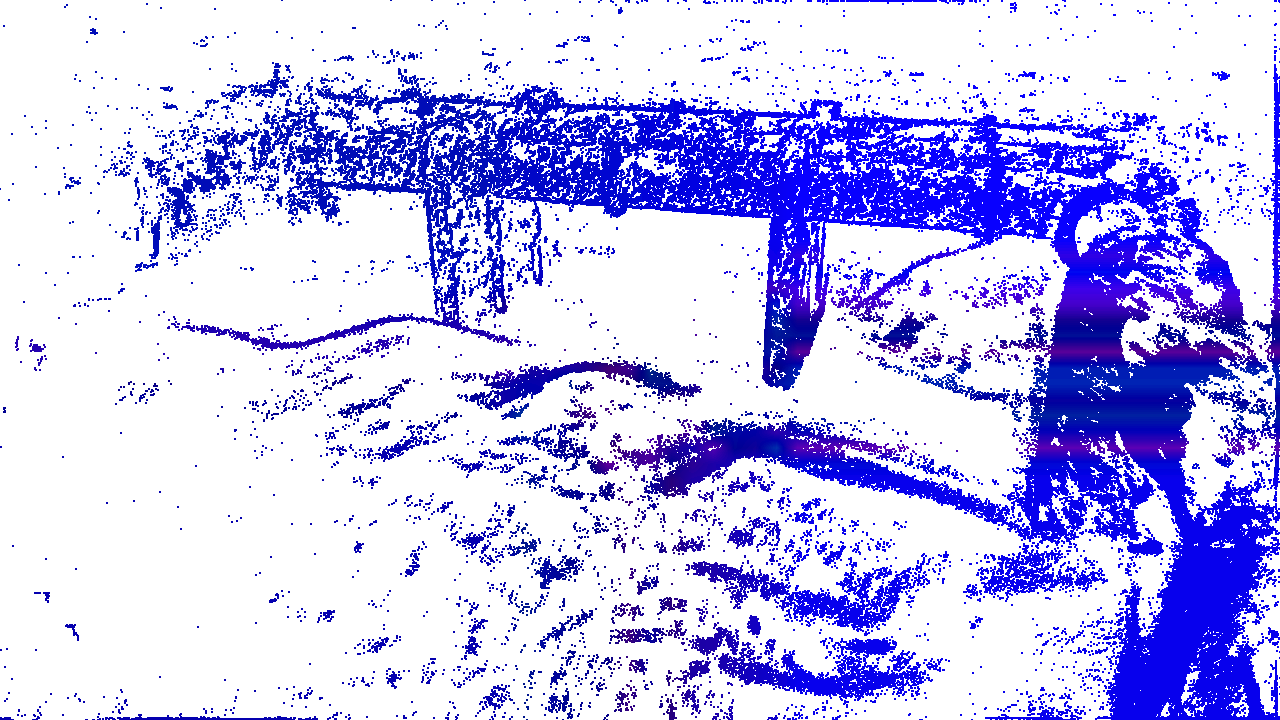}%
  & \includegraphics[width=\imgscale\textwidth,cfbox=gray 0.1pt 0pt]{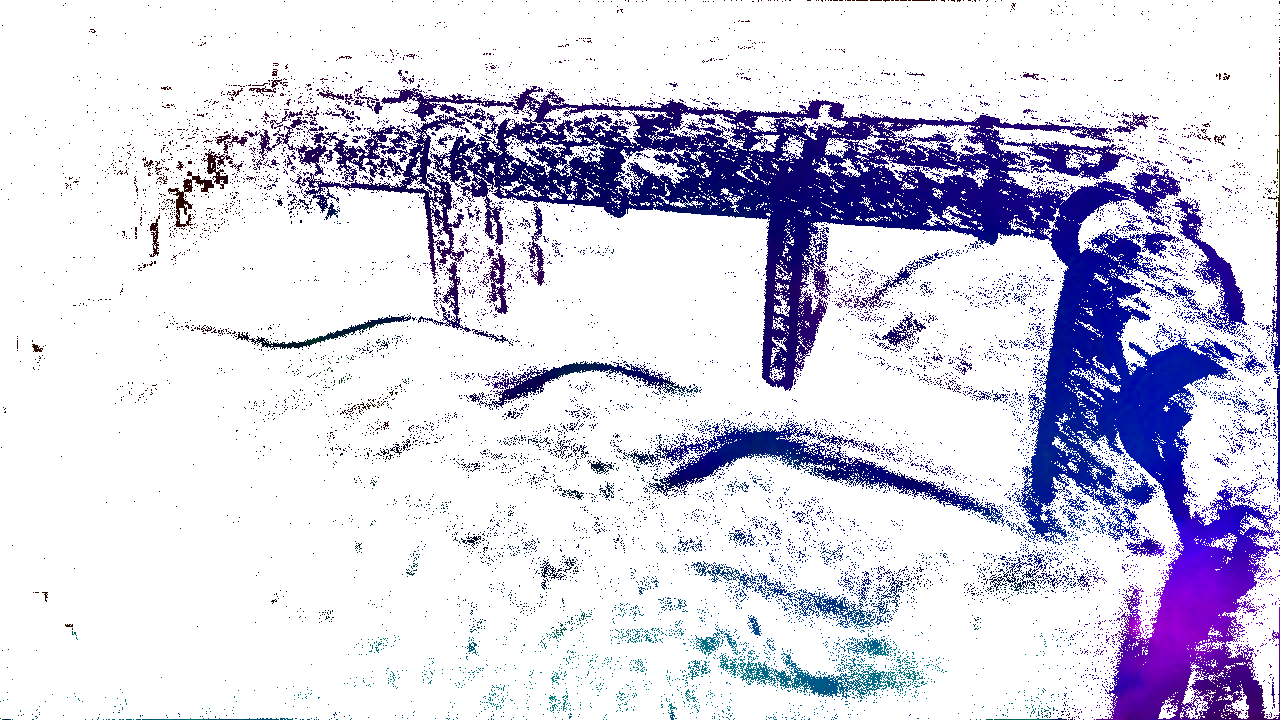}\\[-2pt]
        
  \rotatebox[origin=l]{90}{\begin{adjustbox}{max width=\rotboxscale\textwidth}
    \spacer\texttt{scene5}
  \end{adjustbox}}
  & \includegraphics[width=\imgscale\textwidth,cfbox=gray 0.1pt 0pt]{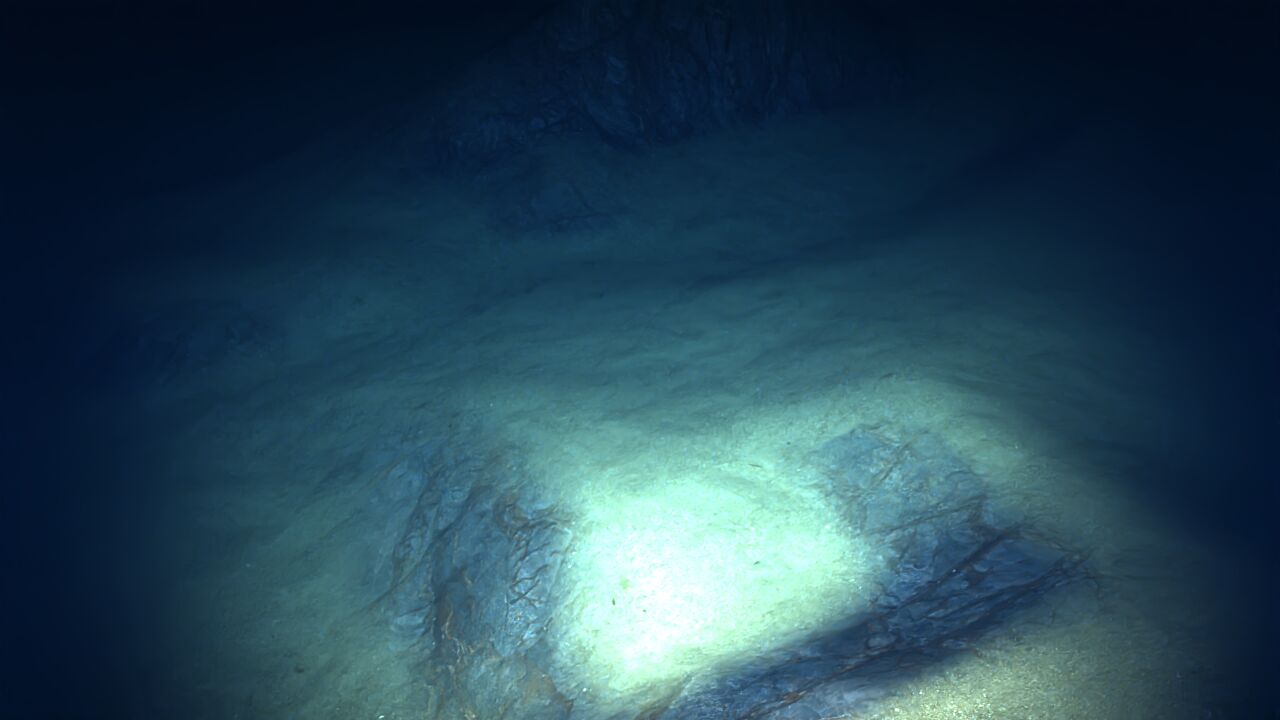}%
  & \includegraphics[width=\imgscale\textwidth,cfbox=gray 0.1pt 0pt]{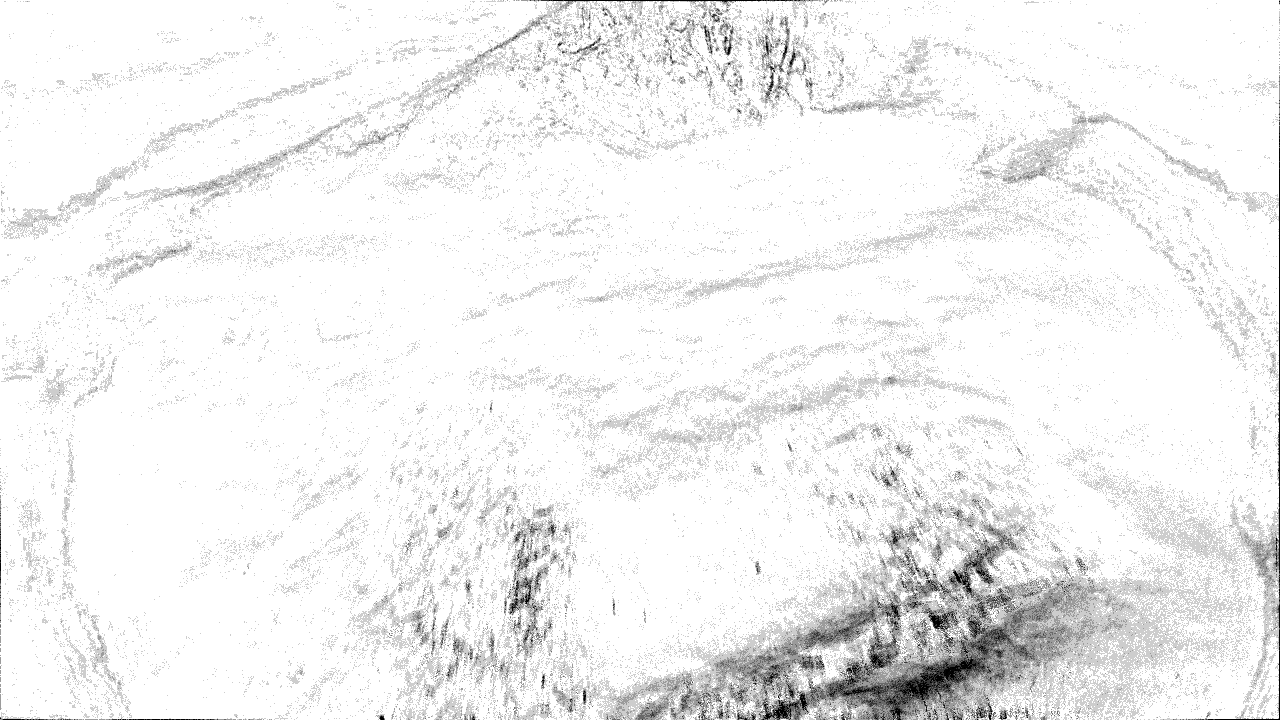}%
  & \includegraphics[width=\imgscale\textwidth,cfbox=gray 0.1pt 0pt]{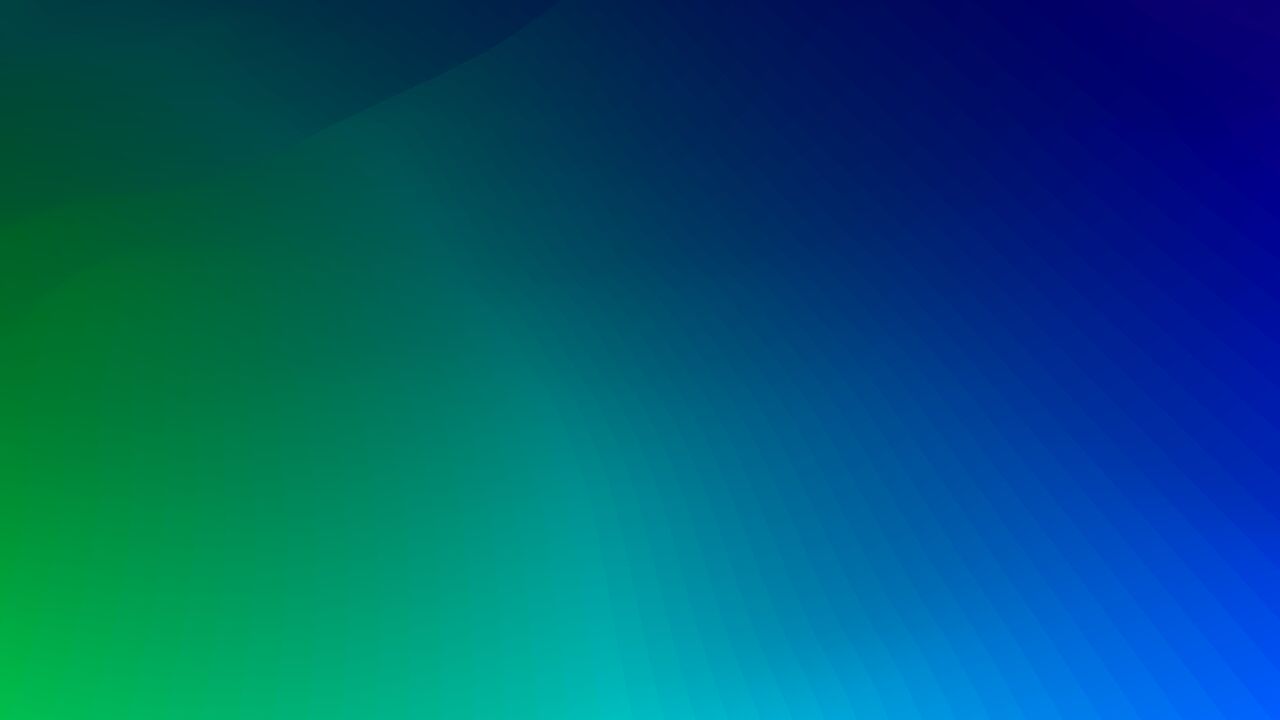}%
  & \includegraphics[width=\imgscale\textwidth,cfbox=gray 0.1pt 0pt]{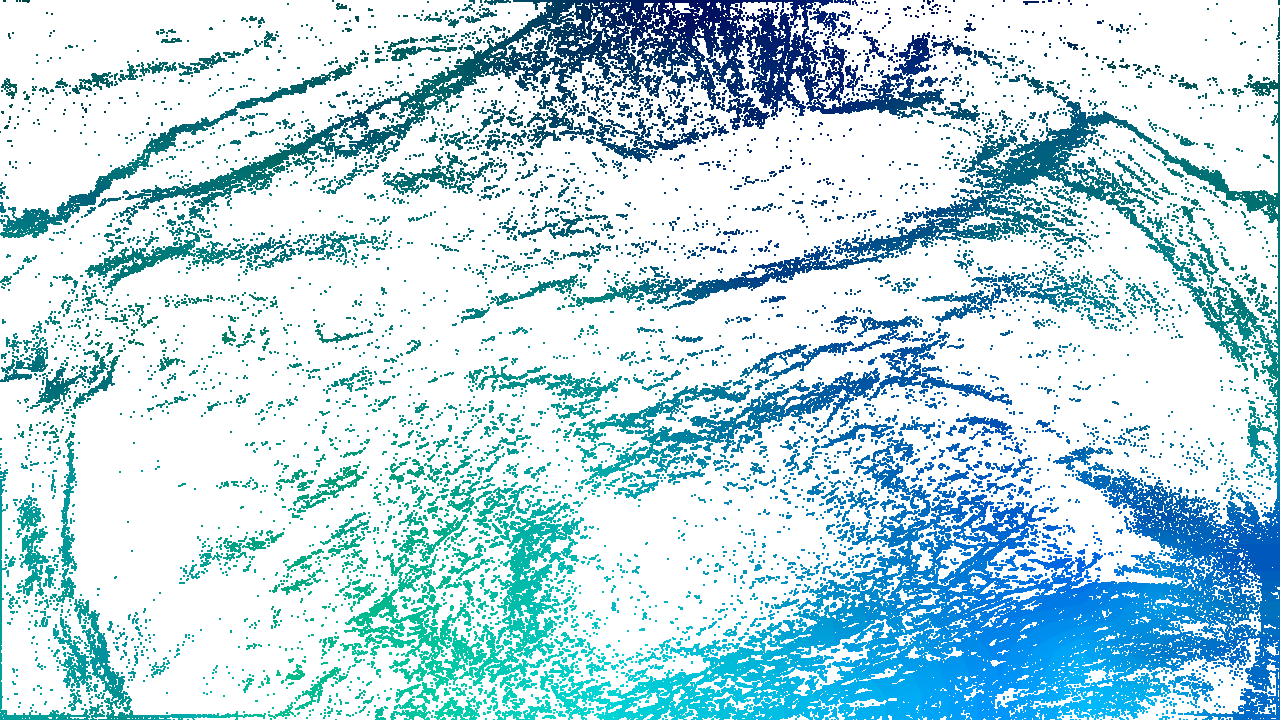}%
  & \includegraphics[width=\imgscale\textwidth,cfbox=gray 0.1pt 0pt]{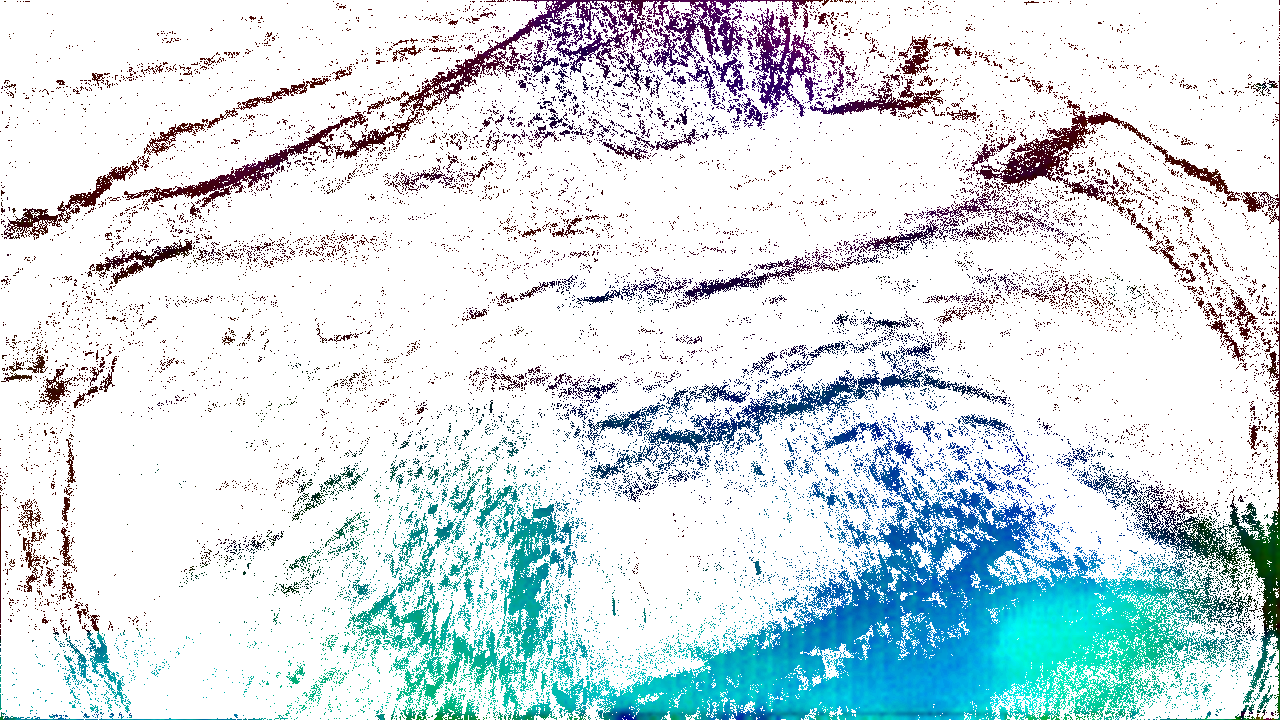}\\

        
  {}
  & (a) Frames 
  & (b) Events
  & (c) GT Flow
  & (d) MultiCM~\cite{shiba2022secrets}
  & (e) MotionPriorCM~\cite{hamann2024motion}\\
  \end{tabular}
  \end{adjustbox}
  \caption{Qualitative event-based optical flow estimation results using the
  UEOF dataset. Columns (a-e) show the frames, events, ground-truth (GT) optical
  flow, and the predicted event-based optical flow using a representative MB
  (MultiCM~\cite{shiba2022secrets}) and LB
  (MotionPriorCMax~\cite{hamann2024motion}) method. The top five and bottom
  five rows correspond to the shallow-water and deep-water environment
  sequences, respectively. The optical flow direction and magnitude are encoded
  according to the color legend shown below.}
  \smallskip
  \includegraphics[height=3.75em]{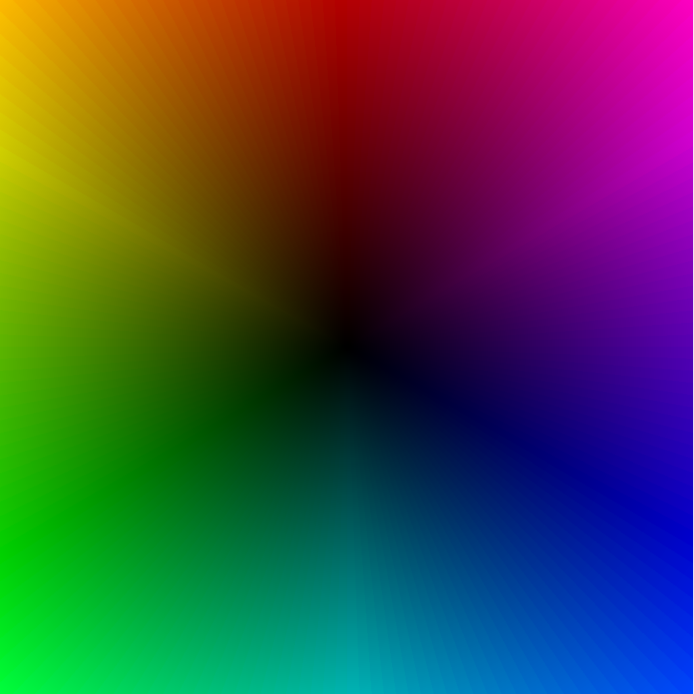}
  \label{fig:qualitative_comparisons}
\end{figure*}

\subsection{Analysis}
\label{subsec:analysis}
Our experiments reveal a significant performance degradation for
state-of-the-art event-based optical flow methods. Specifically, the results
indicate that the limitations are a compounding interaction wherein underwater
phenomena expose and heighten flaws in current event-based processing
algorithms. Supervised approaches such as E-RAFT~\cite{gehrig2021eraft}
exhibited high error rates when relying on pretrained weights from the DSEC
dataset. We hypothesize that this is due to the fact that the terrestrial
datasets are dominated by high-frequency rigid edges (\eg, buildings, vehicles,
\etc). Conversely, underwater scenes, particularly in turbid waters, are
defined by soft, low-frequency textures due to volumetric scattering. As a
result, the supervised networks suffered from severe feature distribution
shifts.

We also noticed that refractive caustics created high-contrast light patterns
that traversed the seafloor independent of physical geometry. Consequently,
contrast maximization approaches that rely on a brightness constancy assumption
revealed a limitation in estimating optical flow from events that are not
caused by motion. Since caustics generate dense, high-frequency event streams,
we observed that this can cause CM objectives to falsely align and produce
incorrect motion estimates. As shown in \cref{fig:qualitative_comparisons}
(shallow-water scenes 1, 2, 3, and 5), a performance gap appears in dynamic
sequences where independent objects (\eg, fish) traverse in different
directions. Methods such as MultiCM~\cite{shiba2022secrets} failed to resolve
these conflicting motions, while MotionPriorCMax~\cite{hamann2024motion}
maintained robustness. 

In addition, the error rates across all models evaluated on the deep-water
environment are notably higher when compared to the shallow-water environment.
We attribute this to the many adverse effects of the deep-water scenes, such as
haloing from UUV lighting, poor contrast due to attenuation, and a low-texture
seafloor. In comparison, the shallow-water scenes contain strong and consistent
sunlight, which can create textures on the seabed. This provides more, albeit
noisy, features for tracking. Finally, the evaluation results demonstrate that
multimodal approaches, specifically those integrating velocity data like
OPCM~\cite{karmokar2025inertia}, vary in improvement on dynamic scenes over
their non-multimodal counterparts.

\section{Conclusion and Future Work}
\label{sec:conclusion_and_future_work}
In this paper, we introduced UEOF, the first synthetic underwater dataset to
leverage PBRT simulation with temporally dense event streams, accurate
ground-truth optical flow, and ego-velocities. Our comprehensive benchmarking
results reveal that state-of-the-art approaches, whether LB or MB, struggle to
generalize from the terrestrial to the underwater domain. Thus, the UEOF
dataset establishes a necessary baseline for advanced research, fostering the
development of robust algorithms capable of handling the large domain gaps
present in underwater environments. We anticipate that this benchmark dataset
will accelerate progress towards reliable multimodal event-based perception for
the next generation of UUVs. Our future work includes extending the UEOF
dataset with additional data and ground-truth modalities, richer environmental
diversity, and physically-grounded event noise models. 

\section*{Acknowledgments}
This material is based upon work supported by the Office of Naval Research
under award number N000142512349.

{
\small
\bibliographystyle{ieeenat_fullname}
\bibliography{ueof_a_benchmark_dataset_for_underwater_event-based_optical_flow}

@PREAMBLE{"\def\authornoop#1{}"}

@inproceedings{alvarez2023mimiruw,
  title={MIMIR-UW: A multipurpose synthetic dataset for underwater navigation and inspection}, 
  author={{\'A}lvarez-Tu{\~n}{\'o}n, Olaya and Kanner, Hemanth and Marnet, Luiza Ribeiro and Pham, Huy Xuan and le Fevre Sejersen, Jonas and Brodskiy, Yury and Kayacan, Erdal},
  booktitle={Proceedings of the IEEE/RSJ International Conference on Intelligent Robots and Systems}, 
  pages={6141--6148},
  year={2023},
  doi={10.1109/IROS55552.2023.10341436}
}

@inproceedings{amer2023unavsim,
  title={UNav-Sim: A visually realistic underwater robotics simulator and synthetic data-generation framework}, 
  author={Amer, Abdelhakim and {\'A}lvarez-Tu{\~n}{\'o}n, Olaya and U{\u{g}}urlu, Halil {\.I}brahim and Sejersen, Jonas Le Fevre and Brodskiy, Yury and Kayacan, Erdal},
  booktitle={Proceedings of the International Conference on Advanced Robotics}, 
  pages={570--576},
  year={2023},
  organization={IEEE},
  doi={10.1109/ICAR58858.2023.10406819}
}

@article{barbosa2025physically,
  title={From physically based to generative models: A survey on underwater image synthesis techniques},
  author={Barbosa, Lucas Amparo and Apolinario Jr, Antonio Lopes},
  journal={Journal of Imaging},
  volume={11},
  number={5},
  pages={161},
  year={2025},
  publisher={MDPI},
  doi={10.3390/jimaging11050161}
}

@article{betancourt2020integrated,
  title={An integrated ROV solution for underwater net-cage inspection in fish farms using computer vision},
  author={Betancourt, J and Coral, W and Colorado, J},
  journal={SN Applied Sciences},
  volume={2},
  number={12},
  pages={1946},
  year={2020},
  publisher={Springer},
  doi={10.1007/s42452-020-03623-z}
}

@article{bi2022non,
  title={Non-uniform illumination underwater image enhancement via events and frame fusion},
  author={Bi, Xiuwen and Wang, Pengfei and Wu, Tao and Zha, Fusheng and Xu, Peng},
  journal={Applied Optics},
  volume={61},
  number={29},
  pages={8826--8832},
  year={2022},
  publisher={Optica Publishing Group},
  doi={10.1364/AO.463099},
}

@misc{blender50,
  title={Blender - A 3D modeling and rendering package},
  author={Blender Online Community},
  url={http://www.blender.org},
  version={5.0},
  year={2025}
}

@article{cartucho2020visionblender,
  title={VisionBlender: a tool to efficiently generate computer vision datasets for robotic surgery},
  author={Cartucho, Jo{\~a}o and Tukra, Samyakh and Li, Yunpeng and Elson, Daniel S. and Giannarou, Stamatia},
  journal={Computer Methods in Biomechanics and Biomedical Engineering: Imaging \& Visualization},
  volume={9},
  number={4},
  pages={331--338},
  year={2021},
  publisher={Taylor \& Francis},
  doi={10.1080/21681163.2020.1835546}
}

@inproceedings{cieslak2019stonefish,
  title={Stonefish: An advanced open-source simulation tool designed for marine robotics, with a ROS interface},
  author={Cie{\'s}lak, Patryk},
  booktitle={Proceedings of OCEANS},
  pages={1--6},
  year={2019},
  organization={IEEE},
  doi={10.1109/OCEANSE.2019.8867434}
}

@inproceedings{cowen1997underwater,
  title={Underwater docking of autonomous undersea vehicles using optical terminal guidance},
  author={Cowen, Steve and Briest, Susan and Dombrowski, James},
  booktitle={Proceedings of OCEANS},
  volume={2},
  pages={1143--1147},
  year={1997},
  organization={IEEE},
  doi={10.1109/OCEANS.1997.624153}
}

@inproceedings{dosovitskiy2017carla,
  title={CARLA: An open urban driving simulator},
  author={Alexey Dosovitskiy and German Ros and Felipe Codevilla and Antonio Lopez and Vladlen Koltun},
  booktitle={Proceedings of the Conference on Robot Learning},
  pages={1--16},
  year={2017},
  doi={10.48550/arXiv.1711.03938},
}

@article{fan2024underwater,
  title={Underwater robot self-localization method using tightly coupled events, images, inertial, and acoustic fusion}, 
  author={Fan, Junfeng and Liu, Xiaoyan and Ou, Yaming and Zhang, Pengju and Zhou, Chao and Hou, Zengguang},
  journal={IEEE Transactions on Industrial Electronics}, 
  volume={72},
  number={5},
  pages={5126--5135},
  year={2025},
  doi={10.1109/TIE.2024.3476931}
}

@incollection{ferone2023synthetic,
  title={A synthetic dataset for learning optical flow in underwater environment},
  author={Ferone, Alessio and Lazzaro, Marco and Scarrica, Vincenzo Mariano and Ciaramella, Angelo and Staiano, Antonino},
  booktitle={Applications of Artificial Intelligence and Neural Systems to Data Science},
  pages={147--156},
  year={2023},
  publisher={Springer},
  doi={10.1007/978-981-99-3592-5_14},
}

@article{gehrig2019video,
  title={Video to events: Bringing modern computer vision closer to event cameras},
  author={Gehrig, Daniel and Gehrig, Mathias and Hidalgo-Carri{\'o}, Javier and Scaramuzza, Davide},
  journal={ArXiv},
  year={2019},
  volume={abs/1912.03095},
  url={https://api.semanticscholar.org/CorpusID:208857551}
}

@article{gehrig2021dsec,
  title={Dsec: A stereo event camera dataset for driving scenarios},
  author={Gehrig, Mathias and Aarents, Willem and Gehrig, Daniel and Scaramuzza, Davide},
  journal={IEEE Robotics and Automation Letters},
  volume={6},
  number={3},
  pages={4947--4954},
  year={2021}
}

@inproceedings{gehrig2021eraft,
  title={E-RAFT: Dense optical flow from event cameras},
  author={Gehrig, Mathias and Millh{\"a}usler, Mario and Gehrig, Daniel and Scaramuzza, Davide},
  booktitle={Proceedings of the International Conference on 3D Vision},
  pages={197--206},
  year={2021},
  organization={IEEE},
  doi={10.1109/3DV53792.2021.00030}
}

@inproceedings{grimaldi2025stonefish,
  title={Stonefish: Supporting machine learning research in marine robotics},
  author={Grimaldi, Michele and Cie{\'s}lak, Patryk and Ochoa, Eduardo and Bharti, Vibhav and Rajani, Hayat and Carlucho, Ignacio and Koskinopoulou, Maria and Petillot, Yvan R and Gracias, Nuno},
  booktitle={Proceedings of the IEEE International Conference on Robotics and Automation},
  pages={1--7},
  year={2025},
  doi={10.1109/ICRA55743.2025.11127421},
}

@article{guo2025utnet,
  title={UTNet: Event-RGB multimodal fusion model for underwater transparent organism detection},
  author={Guo, Fengyue and Ren, Peng and Luo, Cai},
  journal={Intelligent Marine Technology and Systems},
  volume={3},
  number={1},
  pages={18},
  year={2025},
  publisher={Springer},
  doi={10.1007/s44295-025-00065-4},
}

@inproceedings{he2016deep,
  title={Deep residual learning for image recognition},
  author={He, Kaiming and Zhang, Xiangyu and Ren, Shaoqing and Sun, Jian},
  booktitle={Proceedings of the IEEE/CVF Conference on Computer Vision and Pattern Recognition},
  pages={770--778},
  year={2016}
}

@inproceedings{hu2021v2e,
  title={v2e: From video frames to realistic DVS events},
  author={Hu, Yuhuang and Liu, Shih-Chii and Delbruck, Tobi},
  booktitle={Proceedings of the IEEE/CVF Conference on Computer Vision and Pattern Recognition},
  pages={1312--1321},
  year={2021},
  doi={10.1109/CVPRW53098.2021.00144}
}

@incollection{islam2024computer,
  title={Computer vision applications in underwater robotics and oceanography},
  author={Islam, Md Jahidul and Li, Alberto Quattrini and Girdhar, Yogesh A and Rekleitis, Ioannis},
  booktitle={Computer Vision},
  pages={173--204},
  year={2024},
  publisher={Chapman and Hall/CRC},
  doi={10.1201/9781003328957-9},
}

@article{kaneko2024phiswid,
  title={PHISWID: Physics-inspired underwater image dataset synthesized from RGB-D images},
  author={Kaneko, Reina and Ueda, Takumi and Higashi, Hiroshi and Tanaka, Yuichi},
  journal={arXiv preprint arXiv:2404.03998},
  year={2024},
  doi={10.48550/arXiv.2404.03998},
}

@inproceedings{karmokar2025secrets,
  title={Secrets of edge-informed contrast maximization for event-based vision},
  author={Karmokar, Pritam P and Nguyen, Quan H and Beksi, William J},
  booktitle={Proceedings of the Winter Conference on Applications of Computer Vision},
  pages={630--639}, 
  year={2025},
  doi={10.1109/WACV61041.2025.00071}
}

@article{karmokar2025inertia,
  title={Inertia-informed orientation priors for event-based optical flow estimation},
  author={Karmokar, Pritam P and Beksi, William J},
  journal={arXiv preprint arXiv:2511.12961},
  year={2025}
}

@inproceedings{koenig2004gazebo,
  title={Design and use paradigms for gazebo, an open-source multi-robot simulator},
  author={Koenig, Nathan and Howard, Andrew},
  booktitle={Proceedings of the IEEE/RSJ International Conference on Intelligent Robots and Systems},
  pages={2149--2154},
  year={2004},
}

@inproceedings{kyatham2025erebus,
  title={EREBUS: End-to-end robust event based underwater simulation},
  author={Kyatham, Hitesh and Suresh, Arjun and Palnitkar, Aadi and Aloimonos, Yiannis},
  booktitle={Proceedings of the IEEE International Conference on Robotics and Automation AQUA2SIM Workshop},
  year={2025}
}

@article{liao2023gpu,
  title={GPU-accelerated Monte Carlo simulation for a single-photon underwater lidar},
  author={Liao, Yupeng and Shangguan, Mingjia and Yang, Zhifeng and Lin, Zaifa and Wang, Yuanlun and Li, Sihui},
  journal={Remote Sensing},
  volume={15},
  number={21},
  pages={5245},
  year={2023},
  publisher={MDPI},
  doi={10.3390/rs15215245}
}

@inproceedings{li2023blinkflow,
  title={Blinkflow: A dataset to push the limits of event-based optical flow estimation},
  author={Li, Yijin and Huang, Zhaoyang and Chen, Shuo and Shi, Xiaoyu and Li, Hongsheng and Bao, Hujun and Cui, Zhaopeng and Zhang, Guofeng},
  booktitle={Proceedings of the IEEE/RSJ International Conference on Intelligent Robots and Systems},
  pages={3881--3888},
  year={2023},
  doi={10.1109/IROS55552.2023.10341802}
}

@article{li2024breakthrough,
  title={Breakthrough underwater physical environment limitations on optical information representations: An overview and suggestions},
  author={Li, Shuangquan and Zhang, Zhichen and Zhang, Qixian and Yao, Haiyang and Li, Xudong and Mi, Jianjun and Wang, Haiyan},
  journal={Journal of Marine Science and Engineering},
  volume={12},
  number={7},
  pages={1055},
  year={2024},
  publisher={MDPI},
  doi={10.3390/jmse12071055},
}

@article{li2025realistic,
  title={Realistic simulation of underwater scene for image enhancement}, 
  author={Li, Songyang and Liu, Tingyu and Jiang, Qunyan and Li, Yuanqi and Guo, Jie and Jiao, Lei and Guo, Yanwen and Ni, Zhonghua},
  journal={IEEE Transactions on Geoscience and Remote Sensing}, 
  volume={63},
  pages={1--14},
  year={2025},
  doi={10.1109/TGRS.2025.3561927},
}

@inproceedings{liljeback2017eelume,
  title={Eelume: A flexible and subsea resident IMR vehicle},
  author={Liljeb{\"a}ck, P{\aa}l and Mills, Richard},
  booktitle={Proceedings of OCEANS},
  pages={1--4},
  year={2017},
  organization={IEEE},
  doi={10.1109/OCEANSE.2017.8084826}
}

@inproceedings{lin2022oystersim,
  title={OysterSim: Underwater simulation for enhancing oyster reef monitoring}, 
  author={Lin, Xiaomin and Jha, Nitesh and Joshi, Mayank and Karapetyan, Nare and Aloimonos, Yiannis and Yu, Miao},
  booktitle={Proceedings of OCEANS},
  pages={1--6},
  year={2022},
  organization={IEEE},
  doi={10.1109/OCEANS47191.2022.9977233}
}

@article{liu2018detection,
  title={Detection and pose estimation for short-range vision-based underwater docking},
  author={Liu, Shuang and Ozay, Mete and Okatani, Takayuki and Xu, Hongli and Sun, Kai and Lin, Yang},
  journal={IEEE Access},
  volume={7},
  pages={2720--2749},
  year={2018},
  doi={10.1109/ACCESS.2018.2885537}
}

@inproceedings{luo2023learning,
  title={Learning optical flow from event camera with rendered dataset},
  author={Luo, Xinglong and Luo, Kunming and Luo, Ao and Wang, Zhengning and Tan, Ping and Liu, Shuaicheng},
  booktitle={Proceedings of the IEEE/CVF International Conference on Computer Vision},
  pages={9847--9857},
  year={2023},
  doi={10.1109/ICCV51070.2023.00903}
}

@article{luo2023transcodnet,
  title={TransCODNet: Underwater transparently camouflaged object detection via RGB and event frames collaboration},
  author={Luo, Cai and Wu, Jihua and Sun, Shixin and Ren, Peng},
  journal={IEEE Robotics and Automation Letters},
  volume={9},
  number={2},
  pages={1444--1451},
  year={2023},
  doi={10.1109/LRA.2023.3346754}
}

@article{macaulay2025evtslowtv,
  title={EvtSlowTV - A large and diverse dataset for event-based depth estimation},
  author={Macaulay, Sadiq Layi and Kaygusuz, Nimet and Hadfield, Simon},
  journal={arXiv preprint arXiv:2511.02953},
  year={2025}
}

@article{makam2024oceanlens,
  title={OceanLens: An adaptive backscatter and edge correction using deep learning model for enhanced underwater imaging},
  author={Makam, Rajini and Shankari T M, Dhatri and Patil, Sharanya and Sundram, Suresh},
  journal={arXiv preprint arXiv:2411.13230},
  year={2024},
  doi={10.48550/arXiv.2411.13230}
}

@inproceedings{manhaes2016uuv,
  title={UUV Simulator: A Gazebo-based package for underwater intervention and multi-robot simulation}, 
  author={Manh{\~a}es, Musa Morena Marcusso and Scherer, Sebastian A and Voss, Martin and Douat, Luiz Ricardo and Rauschenbach, Thomas},
  booktitle={Proceedings of OCEANS},
  pages={1--8},
  year={2016},
  organization={IEEE},
  doi={10.1109/OCEANS.2016.7761080}
}

@article{mansour2024ecarla,
  title={eCARLA-scenes: A synthetically generated dataset for event-based optical flow prediction},
  author={Mansour, Jad and Rajani, Hayat and Garcia, Rafael and Gracias, Nuno},
  journal={arXiv preprint arXiv:2412.09209},
  year={2024},
  doi={10.48550/arXiv.2412.09209}
}

@article{mansour2025estonefish,
  title={eStonefish-scenes: A synthetically generated dataset for underwater event-based optical flow prediction tasks},
  author={Mansour, Jad and Realpe, Sebastian and Rajani, Hayat and Grimaldi, Michele and Garcia, Rafael and Gracias, Nuno},
  journal={arXiv preprint arXiv:2505.13309},
  year={2025}
}

@article{mohammed20216d,
  title={6D pose estimation for subsea intervention in turbid waters},
  author={Mohammed, Ahmed and Kvam, Johannes and T Thielemann, Jens and Haugholt, Karl H and Risholm, Petter},
  journal={Electronics},
  volume={10},
  number={19},
  pages={2369},
  year={2021},
  publisher={MDPI},
  doi={10.3390/electronics10192369}
}

@article{negahdaripour2007rov,
  title={An ROV stereovision system for ship-hull inspection},
  author={Negahdaripour, Shahriar and Firoozfam, Pezhman},
  journal={IEEE Journal of Oceanic Engineering},
  volume={31},
  number={3},
  pages={551--564},
  year={2007},
  doi={10.1109/JOE.2005.851391}
}

@inproceedings{nielsen2019evaluation,
  title={Evaluation of posenet for 6-dof underwater pose estimation},
  author={Nielsen, Mikkel Cornelius and Leonhardsen, Mari Hovem and Schj{\o}lberg, Ingrid},
  booktitle={Proceedings of OCEANS},
  pages={1--6},
  year={2019},
  organization={IEEE},
  doi={10.23919/OCEANS40490.2019.8962814}
}

@article{novo2024neuromorphic,
  title={Neuromorphic perception and navigation for mobile robots: A review},
  author={Novo, Alvaro and Lobon, Francisco and Garcia de Marina, Hector and Romero, Samuel and Barranco, Francisco},
  journal={ACM Computing Surveys},
  volume={56},
  number={10},
  year={2024},
  publisher={Association for Computing Machinery},
  doi={10.1145/3656469}
}

@article{peng2025aquaticvision,
  title={AquaticVision: Benchmarking visual SLAM in underwater environment with events and frames},
  author={Peng, Yifan and Hong, Yuze and Hong, Ziyang and Chui, Apple Pui-Yi and Wu, Junfeng},
  journal={arXiv preprint arXiv:2505.03448},
  year={2025}
}

@article{randall2023flsea,
  title={FLSea: Underwater visual-inertial and stereo-vision forward-looking datasets}, 
  author={Yelena Randall and Tali Treibitz},
  journal={arXiv preprint arXiv:2302.12772},
  year={2023},
  doi={10.48550/arXiv.2302.12772}
}

@inproceedings{rebecq2018esim,
  title={Esim: An open event camera simulator},
  author={Rebecq, Henri and Gehrig, Daniel and Scaramuzza, Davide},
  booktitle={Proceedings of the Conference on Robot Learning},
  pages={969--982},
  year={2018},
  organization={PMLR}
}

@inproceedings{saksvik2023oslomet,
  title={Towards an open-source benchmark for underwater object detection and pose estimation}, 
  author={Saksvik, Ivar Bj{\o}rgo and Weydahl, H{\aa}kon and Teigland, H{\aa}kon and Alcocer, Alex and Hassani, Vahid},
  booktitle={Proceedings of the IEEE Conference on Underwater Technology}, 
  pages={1--5},
  year={2023},
  doi={10.1109/UT49729.2023.10103392}
}

@article{schontag2025optical,
  title={Optical Ocean Recipes: Creating Realistic Datasets to Facilitate Underwater Vision Research},
  author={Sch{\"o}ntag, Patricia and Nakath, David and Fischer, Judith and R{\"o}ttgers, R{\"u}diger and K{\"o}ser, Kevin},
  journal={arXiv preprint arXiv:2509.20171},
  year={2025},
  doi={10.48550/arXiv.2509.20171},
}

@inproceedings{sedlazeck2011simulating,
  title={Simulating deep sea underwater images using physical models for light attenuation, scattering, and refraction},
  author={Sedlazeck, Anne and Koch, Reinhard},
  booktitle={Proceedings of Vision, Modeling, and Visualization},
  year={2011},
  publisher={The Eurographics Association},
  doi={doi.org/10.2312/PE/VMV/VMV11/049-056}
}

@inproceedings{hamann2024motion,
  title={Motion-prior contrast maximization for dense continuous-time motion estimation},
  author={Hamann, Friedhelm and Wang, Ziyun and Asmanis, Ioannis and Chaney, Kenneth and Gallego, Guillermo and Daniilidis, Kostas},
  booktitle={Proceedings of the European Conference on Computer Vision},
  pages={18--37},
  year={2024},
  organization={Springer},
  doi={10.1007/978-3-031-72646-0\_2}
}

@inproceedings{shah2017airsim,
  title={Airsim: High-fidelity visual and physical simulation for autonomous vehicles},
  author={Shah, Shital and Dey, Debadeepta and Lovett, Chris and Kapoor, Ashish},
  booktitle={Proceedings of the International Conference on Field and Service Robotics},
  pages={621--635},
  year={2017},
  organization={Springer},
  doi={10.1007/978-3-319-67361-5_40}
}

@article{sheikder2025marine,
  title={Marine-inspired multimodal sensor fusion and neuromorphic processing for autonomous navigation in unstructured subaquatic environments},
  author={Sheikder, Chandan and Zhang, Weimin and Chen, Xiaopeng and Li, Fangxing and Liu, Yichang and Zuo, Zhengqing and He, Xiaohai and Tan, Xinyan},
  journal={Sensors},
  volume={25},
  number={21},
  pages={6627},
  year={2025},
  publisher={MDPI},
  doi={10.3390/s25216627}
}

@inproceedings{shiba2022secrets,
  title={Secrets of event-based optical flow},
  author={Shiba, Shintaro and Aoki, Yoshimitsu and Gallego, Guillermo},
  booktitle={Proceedings of the European Conference on Computer Vision},
  pages={628--645},
  year={2022},
  organization={Springer},
  doi={10.1007/978-3-031-19797-0_36}
}

@article{steiner2013understanding,
  title={Understanding the basics of underwater lighting},
  author={Steiner, Aaron},
  journal={Ocean News \& Technology},
  volume={19},
  number={4},
  pages={10--12},
  year={2013},
  publisher={Technology Systems Corporation}
}

@inproceedings{teigland2020operator,
  title={Operator focused automation of ROV operations},
  author={Teigland, H{\aa}kon and Hassani, Vahid and M{\o}ller, Ments Tore},
  booktitle={Proceedings of the IEEE/OES Autonomous Underwater Vehicles Symposium},
  pages={1--7},
  year={2020},
  doi={10.1109/AUV50043.2020.9267917}
}

@misc{unrealengine57,
  title={Unreal Engine},
  author={Epic Games},
  url={https://www.unrealengine.com},
  version={5.7},
  year={2025}
}

@article{wang2025eum,
  title={Eum-slam: An enhancing underwater monocular visual slam with deep learning-based optical flow estimation},
  author={Wang, Xiaotian and Fan, Xinnan and Liu, Yueyue and Xin, Yuanxue and Shi, Pengfei},
  journal={IEEE Transactions on Instrumentation and Measurement},
  year={2025},
  doi={10.1109/TIM.2025.3541785}
}

@inproceedings{yang2025seasplat,
  title={SeaSplat: Representing underwater scenes with 3D gaussian splatting and a physically grounded image formation model}, 
  author={Yang, Daniel and Leonard, John J and Girdhar, Yogesh},
  booktitle={Proceedings of the IEEE International Conference on Robotics and Automation}, 
  pages={7632--7638},
  year={2025},
  doi={10.1109/ICRA55743.2025.11128502}
}

@article{yazdani2020survey,
  title={A survey of underwater docking guidance systems},
  author={Yazdani, Amir Mehdi and Sammut, Karl and Yakimenko, Oleg and Lammas, Andrew},
  journal={Robotics and Autonomous Systems},
  volume={124},
  pages={103382},
  year={2020},
  publisher={Elsevier},
  doi={10.1016/j.robot.2019.103382}
}

@article{yue2025monte,
  title={Monte-Carlo based non-line-of-sight underwater wireless optical communication channel modeling and system performance analysis under turbulence},
  author={Yue, Peng and Wang, XiangRu and Xu, Shan and Li, YunLong},
  journal={arXiv preprint arXiv:2501.12859},
  year={2025},
  doi={10.48550/arXiv.2501.12859}
}

@inproceedings{zhang2022dave,
  title={DAVE aquatic virtual environment: Toward a general underwater robotics simulator}, 
  author={Zhang, Mabel M and Choi, Woen-Sug and Herman, Jessica and Davis, Duane and Vogt, Carson and McCarrin, Michael and Vijay, Yadunund and Dutia, Dharini and Lew, William and Peters, Steven and Bingham, Brian},
  booktitle={Proceedings of the IEEE/OES Autonomous Underwater Vehicles Symposium}, 
  year={2022},
  pages={1--8},
  doi={10.1109/AUV53081.2022.9965808}
}

@inproceedings{zwilgmeyer2021varos,
  title={The VAROS synthetic underwater data set: Towards realistic multi-sensor underwater data with ground truth},
  author={Zwilgmeyer, Peder Georg Olofsson and Yip, Mauhing and Teigen, Andreas Langeland and Mester, Rudolf and Stahl, Annette},
  booktitle={Proceedings of the IEEE/CVF International Conference on Computer Vision},
  pages={3722--3730},
  year={2021},
  doi={10.1109/ICCVW54120.2021.00415}
}
}

\end{document}